\documentclass{article}

\usepackage{arxiv}

\usepackage[utf8]{inputenc} 
\usepackage[T1]{fontenc}    
\usepackage{hyperref}       
\usepackage{url}            
\usepackage{booktabs}       
\usepackage{amsfonts}       
\usepackage{nicefrac}       
\usepackage{microtype}      
\usepackage{lipsum}		
\usepackage{graphicx}
\usepackage{natbib}
\usepackage{doi}
\usepackage{lmodern}  
\usepackage{amsmath, amssymb}
\usepackage{graphicx}
\usepackage{booktabs}     
\usepackage{multirow}
\usepackage{float}
\usepackage{url}   
\usepackage{caption}
\usepackage{subcaption}
\usepackage{xspace}
\usepackage{siunitx}      
\usepackage{algpseudocode}
\usepackage{algorithm}
\usepackage{rotating}
\usepackage{array}
\usepackage{afterpage}
\hypersetup{
    colorlinks=false,
    pdfborder={0 0 0}
}
\usepackage{enumitem}
\newtheorem{thm}{Theorem}
\newtheorem{df}{Definition}
\newtheorem{pf}{Proof}

\title{Bandwidth Selectors on Semiparametric Bayesian Networks}

\author{
\href{https://orcid.org/0009-0000-2396-2446}{\includegraphics[scale=0.06]{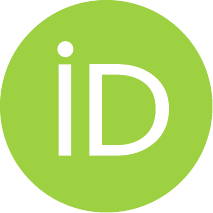}\hspace{1mm}Victor Alejandre} \\
Computational Intelligence Group \\
Departamento de Inteligencia Artificial \\
Universidad Politécnica de Madrid \\
Campus de Montegancedo, Boadilla del Monte \\
28660 Madrid, Spain \\
\texttt{victor.alejandre@upm.es} \\
\And
\href{https://orcid.org/0000-0001-7109-2668}{\includegraphics[scale=0.06]{orcid.pdf}\hspace{1mm}Concha Bielza} \\
Computational Intelligence Group \\
Departamento de Inteligencia Artificial \\
Universidad Politécnica de Madrid \\
Campus de Montegancedo, Boadilla del Monte \\
28660 Madrid, Spain \\
\texttt{mcbielza@fi.upm.es} \\
\And
\href{https://orcid.org/0000-0003-0652-9872}{\includegraphics[scale=0.06]{orcid.pdf}\hspace{1mm}Pedro Larrañaga} \\
Computational Intelligence Group \\
Departamento de Inteligencia Artificial \\
Universidad Politécnica de Madrid \\
Campus de Montegancedo, Boadilla del Monte \\
28660 Madrid, Spain \\
\texttt{pedro.larranaga@fi.upm.es} \\
}

\renewcommand{\shorttitle}{Bandwidth Selectors on Semiparametric Bayesian Networks}

\hypersetup{
pdftitle={Bandwidth Selectors on Semiparametric Bayesian Networks},
pdfsubject={CS, AI},
pdfauthor={Victor Alejandre, Concha Bielza, Pedro Larrañaga},
pdfkeywords={Semiparametric Bayesian networks, Multivariate kernel density estimation, Bandwidth selection, Cross-validation, Plug-in},
}

\begin{document}
\maketitle

\begin{abstract}
Semiparametric Bayesian networks (SPBNs) integrate parametric and non-parametric probabilistic models, offering flexibility in learning complex data distributions from samples. In particular, kernel density estimators (KDEs) are employed for the non-parametric component. Under the assumption of data normality, the normal rule is used to learn the bandwidth matrix for the KDEs in SPBNs. This matrix is the key hyperparameter that controls the trade-off between bias and variance. However, real-world data often deviates from normality, potentially leading to suboptimal density estimation and reduced predictive performance. This paper first establishes the theoretical framework for the application of state-of-the-art bandwidth selectors and subsequently evaluates their impact on SPBN performance. We explore the approaches of cross-validation and plug-in selectors, assessing their effectiveness in enhancing the learning capability and applicability of SPBNs. To support this investigation, we have extended the open-source package \texttt{PyBNesian} for SPBNs with the additional bandwidth selection techniques and conducted extensive experimental analyses.  Our results demonstrate that the proposed bandwidth selectors leverage increasing information more effectively than the normal rule, which, despite its robustness, stagnates with more data. In particular, unbiased cross validation generally outperforms the normal rule, highlighting its advantage in high sample size scenarios. 
\end{abstract}

\keywords{Semiparametric Bayesian networks\and Multivariate kernel density estimation\and Bandwidth selection\and Cross-validation\and Plug-in}

\section{Introduction}
\label{sec:intro}
Statistics is a broad discipline encompassing the collection, description, analysis, and interpretation of data to inform decision-making. One of its pivotal tasks, within the subfield of statistical inference, is the modeling and learning of probability distribution functions (PDFs) from data samples. These functions serve fundamental roles in problems such as exploratory data analysis, regression, or classification (\cite{bishop2006pattern}, \cite{scott2015multivariate}, \cite{silverman2018density}, \cite{Gelman_Hill_Vehtari_2020}). 
There are two primary approaches for learning PDFs: parametric and non-parametric. Parametric models assume data follows a specific distribution, such as Gaussian, and use estimators like maximum likelihood to approximate its parameters from samples. While parametric models converge quickly to the true distribution, they can produce significant errors if the data deviates from the assumed distribution. On the other hand, non-parametric models do not presume data to follow any particular type of distribution. Instead, they are designed to adapt to any probability density function. Hence, these type of models have more flexibility than their counterparts, avoiding those large errors from family distribution assumptions. Nonetheless, this comes with a trade-off concerning convergence rates. The recently introduced semiparametric Bayesian networks (SPBNs) (\cite{atienza2022semiparametric}) integrate methods from both parametric and non-parametric approaches to leverage their respective strengths.

A Bayesian network (BN) is a probabilistic graphical model that encodes the PDF of a set of random variables using the conditional independences among them (\cite{koller2009probabilistic}). It factorizes the joint probability distribution through a directed acyclic graph and conditional probability distributions (CPDs). Nodes represent random variables, and arcs encode dependencies, with CPDs defining the PDF of each variable conditioned on its parents in the graph.

One of the key advantages of BNs is their ability to efficiently represent complex probability distributions while leveraging independence assumptions to reduce computational complexity. They provide a structured and interpretable framework for reasoning under uncertainty, making them particularly useful for probabilistic inference, decision-making, and learning from data. Their modular structure also allows for flexible model specification and easy incorporation of domain knowledge.

SPBNs leverage both parametric and non-parametric models to model CPDs optimally. Specifically, CPDs are modeled parametrically using Gaussian distributions when appropriate, while kernel density estimators (KDEs) offer a non-parametric alternative when needed. KDEs rely on a key hyperparameter, the bandwidth, which controls the trade-off between smoothness and sensitivity in the estimated density function. The accuracy of these estimators in modeling PDFs heavily depends on the proper selection of this hyperparameter, as it plays a crucial role in approximating the true distribution (\cite{scott2015multivariate}). Consequently, a vast body of research focuses on optimizing bandwidth selection, particularly based on data-driven approaches.

Currently, SPBNs employ KDEs with a Gaussian kernel and the normal rule (NR) for bandwidth selection (\cite{atienza2022semiparametric}). However, the inherent normality assumption of the normal rule often leads to bandwidth choices that oversmooth the KDE estimates, reducing their ability to capture fine distributional details and resulting in underfitting. Consequently, SPBNs could benefit from incorporating state-of-the-art bandwidth selectors, enhancing their learning capabilities in both CPD's parameter learning and SPBN structure learning.

This paper establishes a theoretical framework to guarantee the performance of SPBNs in density estimation when incorporating state-of-the-art bandwidth selectors, specifically the plug-in and cross-validation methods. Additionally, the study investigates the effects of alternative bandwidth selection techniques through extensive experimental research on both synthetic and real-world datasets. We demonstrate the enhanced performance of SPBNs compared to the conventional normal rule across several scenarios, ultimately recommending the use of unbiased cross-validation, particularly in high-sample-size scenarios.

The rest of this paper is organized as follows. Section \ref{sec:Background} provides the background on BNs, KDEs and bandwidth selection. Section \ref{sec:3} presents bandwidth selection in SPBNs. Section \ref{sec:experiments} presents experiments analyzing the impact of these selectors on SPBN structure and CPD learning. Finally, Section \ref{sec:conclusions} concludes the study and outlines future research directions. 

\section{Background}
\label{sec:Background}
\subsection{Bayesian networks}
\label{sec:bayesian_networks}
\begin{figure}[h]
\centering 
\includegraphics[scale = 0.6]{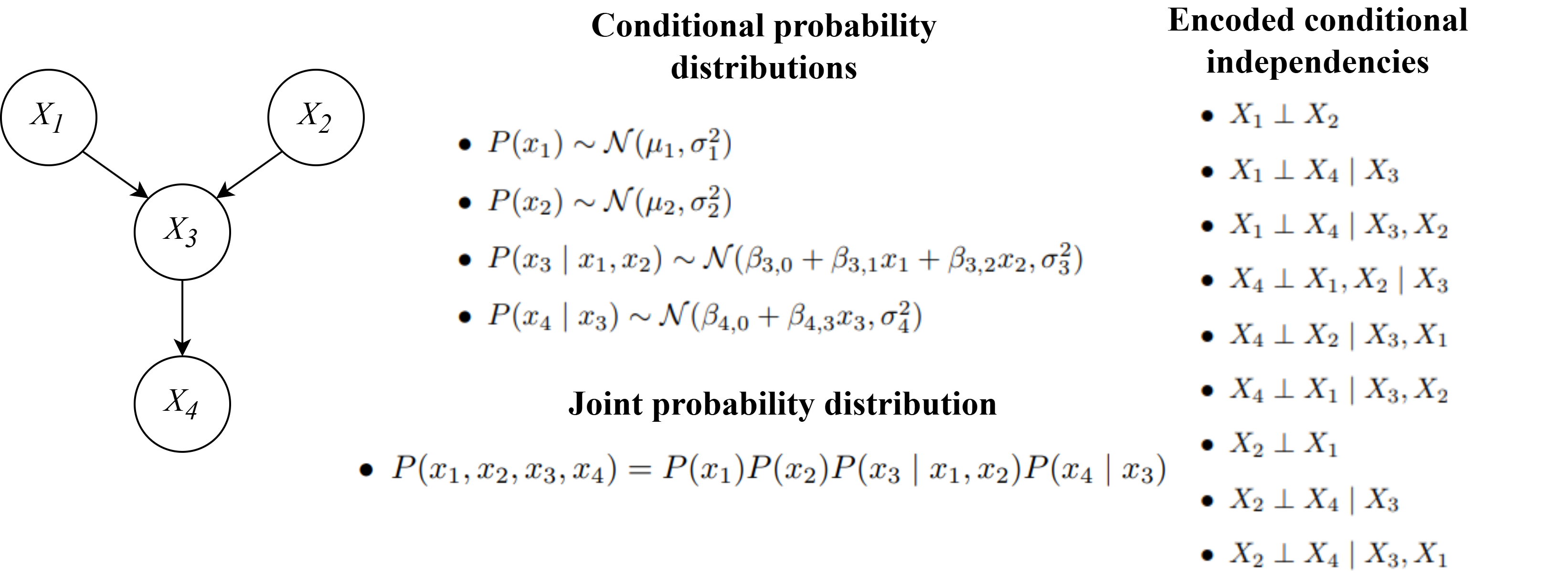}
\caption{Example of a Gaussian Bayesian network modeling the multivariate Gaussian distribution of $(X_1, X_2, X_3, X_4)$, including the graph, CPDs, and the conditional independencies between variables encoded in the model.
} 
\label{fig:bn_example}
\end{figure}

A BN (\cite{pearl1988probabilistic}, \cite{koller2009probabilistic}) 
is a type of probabilistic graphical model. It represents in a compact way a joint probability distribution $P(X_{1},...,X_{d})$ using conditional independences between triplets of variables (see Figure \ref{fig:bn_example}). Formal definitions follow.

\begin{df}
Given a set of random variables $\mathbf{Z}$ and two random variables $X,Y$ then $X$ is conditional independent of $Y$ given $\mathbf{Z}$, $X\bot Y\mid \mathbf{Z}$, if and only if 
\[P(X,Y\mid \mathbf{Z})=P(X\mid \mathbf{Z})P(Y\mid \mathbf{Z})\]
\label{definition2:2}
\end{df}

\begin{df}
Given a random vector $\mathbf{X}=(X_{1},...,X_{d})$ following a joint probability distribution $P(X_{1},...,X_{d})$, a BN is a tuple $\mathcal{B}=(\mathcal{G},\boldsymbol{\theta})$  where $\mathcal{G}=(\mathbf{V},\mathbf{A})$ is a \textit{directed acyclic graph} (DAG), with $\mathbf{V}=\{X_{1},...,X_{d}\}$ the set of vertices and $\mathbf{A}\subset \mathbf{V}\times \mathbf{V}$ the set of arcs. $\boldsymbol{\theta}=\{P(X_{j}\mid \mathbf{X}_{Pa(j)}), j=1,...,d\}$ represents the set of CPDs for each random variable  $X_j$, given its parents $\mathbf{X}_{Pa(j)}$ in $\mathcal{G}$, where $Pa(j)$ indicates the indices of the parent nodes.
\label{definition2:3}
\end{df}

The graph encodes a set of conditional independences between triplets of variables. This results in a factorization of $P(X_{1},...,X_{d})$:
\[P(X_{1},...,X_{d})=\prod_{j=1}^dP(X_{j}\mid \mathbf{X}_{Pa(j)})\]

The key advantage of BNs lies in their ability to factorize complex distributions into smaller, more manageable estimation problems. This decomposition simplifies the overall task by breaking it down into a series of CPDs, making it more tractable and computationally efficient. The learning of CPDs from data is dependent on the chosen modeling approach. Additionally, the network structure can also be induced from data, with various methodologies available to capture conditional (in)dependencies present in the data.

Bayesian Networks (BNs) can employ either parametric or non-parametric conditional probability distributions (CPDs). Common parametric models include: (i) Discrete BNs, where CPDs are represented as categorical distributions through conditional probability tables (CPTs); each configuration of parent variables corresponds to a categorical distribution over the child variable (\cite{koller2009probabilistic}); (ii) Gaussian BNs, where the joint distribution \( P(X_1, \dots, X_d) \) is assumed to be multivariate Gaussian, and dependencies are modeled using linear Gaussian (LG) CPDs (\cite{gaussiannetworks}); (iii) Tweedie BNs, where CPDs belong to the Tweedie family, a specialized subclass of the exponential family (\cite{masmoudi2019new}); and (iv) more flexible models based on mixtures, such as mixtures of truncated exponentials (\cite{moral2001mixtures}), mixtures of polynomials (\cite{shenoy2011inference}), and mixtures of truncated basis functions (\cite{langseth2012mixtures}).

For non-parametric CPDs, the modeling is done by means of Gaussian processes (\cite{Friedman2000GaussianPN}), infinite mixtures (\cite{ickstadt2010nonparametric}) and KDEs (\cite{hofmann1995discovering}). 

Structure learning, i.e., the inference of the graph $\mathcal{G}$, can be categorized into three main approaches: (i) constraint-based methods, which analyze conditional independencies among variables using statistical tests and construct a structure that best captures the identified independencies; (ii) score-based methods, which search for the optimal graph structure by maximizing a goodness-of-fit function of the structure to the data. Common scoring criteria include penalized log-likelihood measures, such as the Bayesian information criterion, and Bayesian scores; and (iii) hybrid methods, which integrate elements of both constraint-based and score-based approaches.

For score-based methods, a space of structures and a search method for that space are required in addition to the score. Although there are three potential search spaces—DAGs, equivalent classes of DAGs, and variable orderings—the literature typically focuses on the DAGs space. The most representative algorithms in this context are hill-climbing (HC) (\cite{cooper1992bayesian}) and its variants. Some works, such as those by \cite{chickering1996learning} and \cite{larranaga1996learning}, explore the equivalence classes of DAGs and variable orderings, respectively.

\subsection{Kernel density estimation}
\subsubsection{Kernel density estimators}
\label{sec:2.1}
A kernel density estimator is a non-parametric technique used to estimate the density $f$ of a random variable $X$  or a random vector $\mathbf{X}=(X_1,....,X_d)$. The KDE for the PDF of a multivariate random vector $\mathbf{X}$ is in Definition \ref{definition2:1} (\cite{scott2015multivariate}).

\begin{df}
Given a non-singular matrix $\mathbf{H}_N\in \mathbb{M}^{dxd}$, a set of random vectors i.i.d. $\mathbf{X}^1,...,\mathbf{X}^N\sim \mathbf{X}$ with density $f(\mathbf{x})$ and a kernel function $K:\mathbb{R}^d \rightarrow \mathbb{R}$ which satisfies:
\begin{itemize}
    \item $\displaystyle \int_{\mathbb{R}^d} K(\mathbf{w}) d\mathbf{w} = 1$
    \item $\displaystyle \int_{\mathbb{R}^d} \mathbf{w} K(\mathbf{w}) d\mathbf{w} = \mathbf{0}_d \in \mathbb{R}^d$
    \item $\displaystyle \int_{\mathbb{R}^d} \mathbf{w} \mathbf{w}^T K(\mathbf{w}) d\mathbf{w} = \mathbf{I}_d \in \mathbb{M}^{d\times d}$
\end{itemize}

a multivariate KDE of PDF $f$ is defined as:
\[\hat{f}(\mathbf{x};\mathbf{H}_N)=\frac{1}{N}\sum_{i=1}^{N}K_{\mathbf{H}_N}(\mathbf{x}-\mathbf{X}^i)=\frac{1}{N |\mathbf{H}_N|}\sum_{i=1}^{N}K(\mathbf{H}_N^{-1}(\mathbf{x}-\mathbf{X}^i))\]
where $K_{\mathbf{H}_N}$ is the scaled kernel function and $|\mathbf{H}_N|$ is the determinant of matrix $\mathbf{H}_N$.
\label{definition2:1}
\end{df}

As any estimator, the KDE is a random variable depending on the distribution of $\mathbf{X}$ through the i.i.d. $\mathbf{X}^1,...,\mathbf{X}^N\sim \mathbf{X}$. Therefore, given a sample dataset $\mathcal{D} = \left\{ \mathbf{x}^i \right\}_{i=1}^{N}$ extracted from the i.i.d. $\mathbf{X}^1,...,\mathbf{X}^N\sim \mathbf{X}$, an actual estimation of $f(\mathbf{x})$ is:
\[\hat{f}(\mathbf{x};\mathbf{H}_N,\mathcal{D})=\frac{1}{N}\sum_{i=1}^{N}K_{\mathbf{H}_N}(\mathbf{x}-\mathbf{x}^i)=\frac{1}{N |\mathbf{H}_N|}\sum_{i=1}^{N}K(\mathbf{H}_N^{-1}(\mathbf{x}-\mathbf{x}^i))\]

$\mathbf{H}_N$ is the bandwidth parameter, which is a linear transformation of the original space. For ease of notation, we will denote $\mathbf{H} = \mathbf{H}_N$, where the dependence of the bandwidth parameter on the sample size $N$ is implicitly understood. As mentioned by \cite{scott2015multivariate}, the bandwidth can be included in the definition of the kernel without loss of generality. For example, using either $K=\mathcal{N}(\mathbf{0}_d, \mathbf{\Sigma})$ with $\mathbf{H} = \mathbf{I}_d$ or $K=\mathcal{N}(\mathbf{0}_d, \mathbf{I}_d)$ with $\mathbf{H} = \mathbf{\Sigma}^{-\frac{1}{2}}$ would be equivalent. In this case, we are dealing with the Gaussian KDE, which is defined as:

\begin{equation}
\begin{aligned}
\widehat{f}(\mathbf{x}; \mathbf{H}) &= \frac{1}{N |\mathbf{H}|^{1/2}} \sum_{i=1}^{N} K\left(\mathbf{H}^{-1/2}(\mathbf{x} - \mathbf{X}^i)\right), \\
\text{with } K(\mathbf{x}) &= \frac{1}{(2\pi)^{d/2}} \exp\left(-\frac{1}{2} \mathbf{x}^T \mathbf{x} \right)
\end{aligned}
\label{eq:gaussiankde}
\end{equation}

where the bandwidth $\mathbf{H}$ is a symmetric, positive definite matrix, i.e. it belongs to the matrix class $\mathcal{F}$\footnote{$\mathcal{F}$ is the set of symmetric, positive definite, unconstrained matrices of dimension $d\times d$}. This type of KDE is employed in SPBNs, although other types, such as the Epanechnikov, Cauchy, or ASH triangle kernels, also exist (\cite{scott2015multivariate}).

As with any estimator, properties such as bias, consistency, and convergence rates can be studied for KDEs. Bias refers to the difference between the expected value of an estimator and the true value, which in this case is $f(\mathbf{x})$. In other words, bias measures the systematic error introduced by the estimator. Consistency states that an estimator converges in probability to the true value, guaranteeing the performance of the estimator:
\[
\hat{f}(\mathbf{x}; \mathbf{H}) \xrightarrow{P} f(\mathbf{x}) \text{ as } N \to \infty \quad \Longleftrightarrow \quad  \forall \epsilon > 0,\ \lim_{N \to \infty} P\left( \left| \hat{f}(\mathbf{x};\mathbf{H}) - f(\mathbf{x}) \right| \geq \epsilon \right) = 0.
\]

The convergence rates are typically studied in terms of the mean integrated squared error (MISE) (see Equation \ref{eq:MISE}), which serves as a global error measure of the KDE. It is defined as the mean of the random variable known as the integrated squared error (ISE) (see Equation \ref{eq:ISE}), or equivalently, as the integral of the mean squared error (MSE) (see Equation \ref{eq:MSE}), which serves as a local error measure at a given point $\mathbf{x}$. A faster convergence of $\text{MISE} \to 0$ indicates a better estimator.

\begin{equation}
\begin{split}
    \text{MISE}\{\hat{f}(\mathbf{x};\mathbf{H})\} &= \mathbb{E}\left[\int_{\mathbb{R}^d} (f(\mathbf{x})-\hat{f}(\mathbf{x};\mathbf{H}))^2d\mathbf{x} \right] = \int_{\mathbb{R}^d}\mathbb{E}\left[(f(\mathbf{x})-\hat{f}(\mathbf{x};\mathbf{H}))^2\right]d\mathbf{x}  \\
    &= \int_{\mathbb{R}^d}\text{MSE}\{\hat{f}(\mathbf{x};\mathbf{H})\}d\mathbf{x}= \text{IV}\{\hat{f}(\mathbf{x};\mathbf{H})\} + \text{ISB}\{\hat{f}(\mathbf{x};\mathbf{H})\}
\end{split}
\label{eq:MISE}
\end{equation}

\begin{equation}
    \text{ISE}\{\hat{f}(\mathbf{x};\mathbf{H})\} =  \int_{\mathbb{R}^d}(f(\mathbf{x})-\hat{f}(\mathbf{x};\mathbf{H}))^2 d\mathbf{x} 
    \label{eq:ISE}
\end{equation}

\begin{equation}
    \text{MSE}\{\hat{f}(\mathbf{x};\mathbf{H})\} = \mathbb{E}\left[(f(\mathbf{x})-\hat{f}(\mathbf{x};\mathbf{H}))^2\right]= \text{Var}\{\hat{f}(\mathbf{x};\mathbf{H})\}+\text{Bias}^2\{\hat{f}(\mathbf{x};\mathbf{H})\}
    \label{eq:MSE}
\end{equation}

Notice that MSE decomposes into the sum of the squared bias and variance of the KDE. Thus, MISE decomposes into the integrated variance (IV) and the integrated squared bias (ISB). Moreover, MISE is the expected $L_2$ distance between $\hat{f}(\mathbf{x};\mathbf{H})$ and $f(\mathbf{x})$. $L_2$ is the usual study metric for discrepancy between functions due to its mathematical properties. Other $L_p$ measures can be used with special focus on $L_1$ (\cite{chacon2018multivariate}). The convergence of MISE, i.e. $\text{MISE}\rightarrow 0$ as $N \to \infty$, guarantees the global performance of the KDE as it implies consistency of the estimator and asymptotically unbiasedness.

The study of the necessary and sufficient conditions on $f(\mathbf{x})$, $\hat{f}(\mathbf{x};\mathbf{H})$, and $\mathbf{H}$ for establishing the different properties and convergence behaviors of a KDE is a well-developed research topic. Specifically, under minimal assumptions, the KDE $\hat{f}(\mathbf{x};\mathbf{H})$ is a consistent estimator in terms of the MISE (\cite{bosq1987th}). However, rather than focusing on the minimal conditions for MISE convergence, we concentrate on the conditions outlined in \cite{chacon2018multivariate}:

\begin{itemize}[label=\textbf{(\arabic*)}, ref=\textbf{(\arabic*)}]

    \item [(A1)] \label{item:conditionA1} The density function $f(\mathbf{x})$ is square integrable and twice differentiable, with all of its second order partial derivatives bounded, continuous and square integrable.
    \item  [(A2)] \label{item:conditionA2} The kernel $K$ is square integrable, spherically symmetric and with a finite second order moment; this means that $\int_{\mathbb{R}^d}\mathbf{x}K(\mathbf{x})d\mathbf{x}=\mathbf{0}_d$ and  $\int_{\mathbb{R}^d}\mathbf{x}\mathbf{x}^TK(\mathbf{x})d\mathbf{x}=m_2(K)\mathbf{I}_d$ with $m_2(K)=\int_{\mathbb{R}^d}x_j^2K(\mathbf{x})d\mathbf{x}$ for all $j=1,...,d$.
    \item  [(A3)] \label{item:conditionA3} The bandwidth matrices $\mathbf{H} = \mathbf{H}_N$ form a sequence of positive definite, symmetric matrices such that $vec\mathbf{H}\rightarrow \mathbf{0}_{d^2}$ and $N^{-1}|\mathbf{H}|^{-1/2}\rightarrow 0$ as $N\rightarrow\infty$.
\end{itemize}

Here, $vec\mathbf{H}$ is the vectorization operator (\cite{henderson1979vec}). Notice that in condition A2 we assume that the second order moment is a constant and not just the identity matrix as in Definition \ref{definition2:1}. Kernel functions must integrate to one; other conditions imposed on the kernel definition vary through the literature, mainly depending on the task at hand. 

Under this theoretical framework it can also be proved that the KDE is consistent in terms of MISE (\cite{chacon2018multivariate}) leading to a consistent ($\hat{f}(\mathbf{x};\mathbf{H})\xrightarrow{P}f(\mathbf{x})\text{ when }N\rightarrow \infty$) and asymptotically unbiased estimator.

The origin of these conditions stems from data-driven bandwidth selection methods. Given the significance and consequences of MISE convergence, much of the literature on bandwidth selection has focused on minimizing this measure. However, working directly with MISE can be challenging, which is why many authors have turned to the asymptotically mean integrated squared error (AMISE) (see Equation \ref{eq:AMISE}). 

\begin{equation}
\begin{aligned}
\text{AMISE}\{\hat{f}(\mathbf{x};\mathbf{H})\} = N^{-1}|\mathbf{H}|^{-1/2}R(K) + \frac{1}{4}m_2(K)^2\int_{\mathbb{R}^d} \text{tr}^2\{\mathbf{H} Hf(\mathbf{x})\} d\mathbf{x},
\end{aligned}
\label{eq:AMISE}
\end{equation}
\[
R(K) = \int_{\mathbb{R}^d} K(\mathbf{x})^2 d\mathbf{x}, 
Hf(\mathbf{x}) \text{ is the Hessian matrix of } f(\mathbf{x}), \text{ and }\text{tr}\{ \mathbf{A}\} \text{ is the trace operator of } \mathbf{A}.\]

Under conditions (A1), (A2) and (A3), AMISE simplifies MISE by applying a Taylor expansion to $f(\mathbf{x})$, making it more tractable in mathematical terms. The first summation term corresponds to the approximation of the IV, whereas the second to the approximation of the ISB. Also, since AMISE is an asymptotic approximation of MISE, convergence in AMISE carries the same implications.

Moreover, we can define the optimal (also called oracle) bandwidth matrices (Equations \ref{eq:oracle_bandwidth_mise} and \ref{eq:oracle_bandwidth_amise}), which are the ones that minimize the MISE and AMISE, respectively, given $\hat{f}(\mathbf{x};\mathbf{H})$: 

\begin{equation}
\mathbf{H}_{\text{MISE}} = \arg\min_{\mathbf{H} \in \mathcal{F}} \text{MISE}(\hat{f}(\mathbf{x}; \mathbf{H}))
\label{eq:oracle_bandwidth_mise}
\end{equation}

\begin{equation}
\mathbf{H}_{\text{AMISE}} = \arg\min_{\mathbf{H} \in \mathcal{F}} \text{AMISE}(\hat{f}(\mathbf{x}; \mathbf{H}))
\label{eq:oracle_bandwidth_amise}
\end{equation}

Under conditions (A1)-(A3), it can be proven that these matrices are $O(N^{-2/(d+4)})$. Therefore, the minimal MISE rate is bounded and converges to zero at a rate of $O(N^{-4/(d+4)})$ (\cite{chacon2018multivariate}). Consequently, using $\mathbf{H}_{\text{MISE}}$ or $\mathbf{H}_{\text{AMISE}}$ as the smoothing parameter is optimal since the MISE convergence guarantees the properties of the KDE. However, oracle bandwidth matrices depend on the unknown density $f$, so data-driven bandwidth selectors address this issue by estimating and minimizing MISE or AMISE.

The minimal MISE rate given the oracle bandwidths is a counterpart for KDEs, as, compared to the typical convergence rate of the parametric case, $O(N^{-1})$, it is slower. As noted by \cite{atienza2022semiparametric}, the MISE convergence rate depends on the dimension $d$. This motivates the use of Bayesian networks, which reduce dimensionality through factorization and conditional independencies.

\subsubsection{Bandwidth selection}
\label{sec:bandwidth_selection}
The bandwidth in KDEs serves as a crucial smoothing parameter that controls the level of approximation in density estimation. It determines how much influence each data point has on the overall estimate, directly affecting the balance between bias and variance. A small bandwidth leads to a highly localized model, capturing fine details but risking overfitting by adapting too closely to noise, a phenomenon known as undersmoothing. Conversely, a large bandwidth produces a smoother estimate, potentially underfitting by oversimplifying the underlying structure, which is referred to as oversmoothing. 

We have seen in Section \ref{sec:2.1} that oracle bandwidth matrices would be ideal as bandwidth parameters. However, they are not computable in practice. Consequently, a vast body of research has focused on bandwidth selection, particularly on automatic data-driven approaches that optimize a criterion based on the data sample. While this paper focuses on methods that minimize MISE or its asymptotic approximations, alternative approaches exist. For instance, one could select $\mathbf{H}$ by maximizing the likelihood. However, this is not directly feasible (\cite{atienza2022semiparametric}) because the training data is incorporated into the KDE model, causing the bandwidth matrix $\mathbf{H}$ to shrink, $|\mathbf{H}| \to 0$, leading to an infinite likelihood value. Nonetheless, some works circumvent this issue using leave-one-out estimators (\cite{leiva2012algorithms}). Additionally, Bayesian estimation has also been explored for bandwidth selection (\cite{zhang2006bayesian}). 
 
\begin{figure}[h]
\centering 
\includegraphics[scale = 0.49]{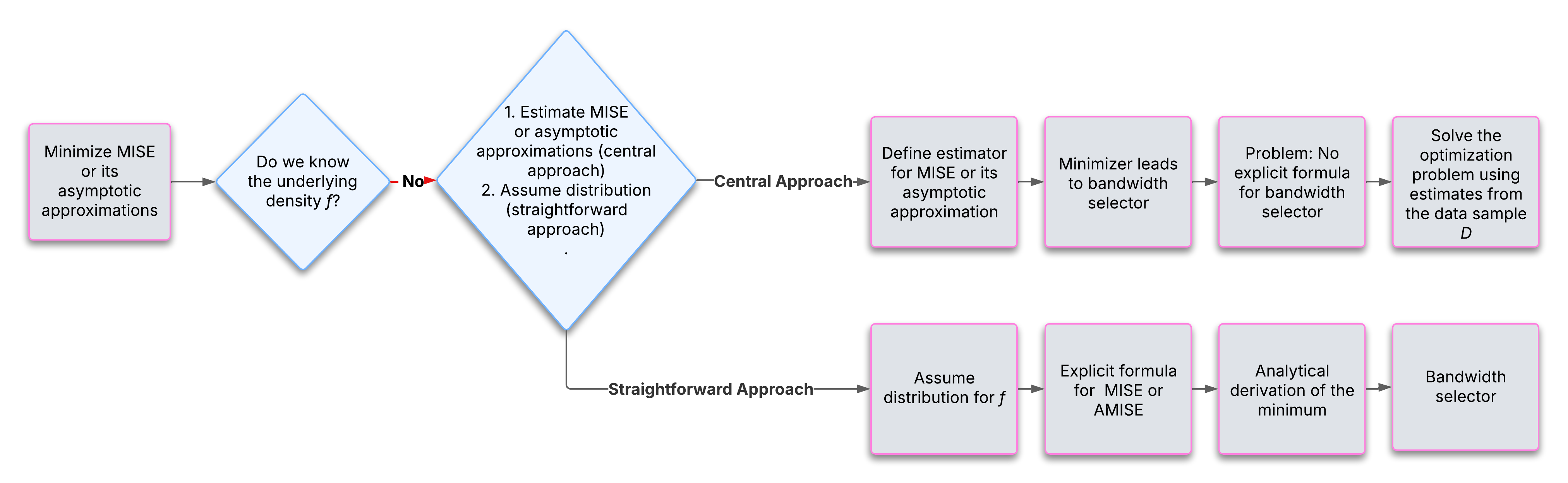}
\caption{General scheme of bandwidth selection aimed at minimizing MISE or its asymptotic approximations.
} 
\label{fig:BS_process}
\end{figure}

We focus on bandwidth selectors based on MISE or its asymptotic approximations, as they are the most relevant, given the desirable properties imposed by convergence in MISE. Figure \ref{fig:BS_process} displays the general scheme of bandwidth selection targeting MISE or its asymptotic approximations.

The central approach involves defining estimators for the MISE or its asymptotic approximations. Research has therefore focused on developing such estimators, whose theoretical minima yield the desired bandwidth selectors denoted by $\widehat{\mathbf{H}}$. These selectors estimate the oracle bandwidth matrices, but explicit formulas for them are rarely available. As a result, bandwidth selection becomes a practical minimization problem of the estimated MISE (or its approximation) based on a data sample $\mathcal{D}$. This must be considered when choosing and implementing a bandwidth selection method.

\begin{figure}[htbp]
    \centering
    \begin{subfigure}[t]{0.48\linewidth}
        \centering
        \includegraphics[width=\linewidth]{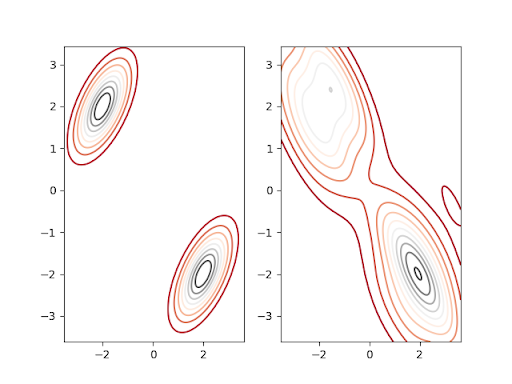}
        \caption{Normal reference rule (oversmoothed)}
        \label{fig:NR_vs_PI(NR)}
    \end{subfigure}
    \hfill
    \begin{subfigure}[t]{0.48\linewidth}
        \centering
        \includegraphics[width=\linewidth]{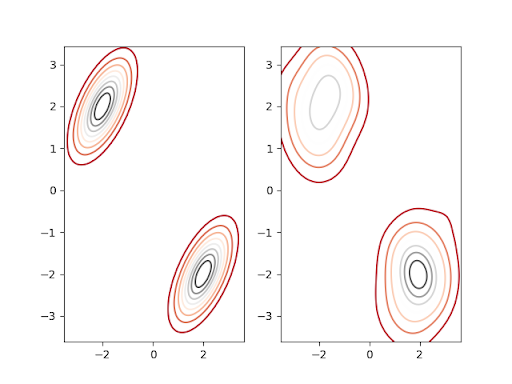}
        \caption{Plug-in (better detail preservation)}
        \label{fig:NR_vs_PI(PI)}
    \end{subfigure}
    \caption{Comparison of bandwidth selectors in a bivariate KDE. Each plot shows the estimated density using either the normal reference rule or the plug-in method compared to the true underlying density, overlaid with contour levels.}
    \label{fig:nr_vs_pi}
\end{figure}

A more straightforward approach involves directly assuming a specific distribution for the unknown density, such as a Gaussian independent distribution, and computing the MISE or AMISE. The normal reference rule bandwidth selector is an example of this approach. However, this approach introduces bias when the underlying distribution assumption does not hold. This is exemplified by the normal reference rule, which tends to oversmooth the density estimate (Figure \ref{fig:nr_vs_pi}).

Plug-in and cross-validation bandwidth selectors, based on the central approach, are crucial for KDE bandwidth selection. Initially, for univariate KDE (\cite{heidenreich2013bandwidth}), these methods were extended to multivariate KDE with constrained bandwidth matrices (\cite{wand1994multivariate}, \cite{sain1994cross}). Recent research adapts these methods for a broader class of symmetric, positive definite, unconstrained matrices in $\mathcal{F}$, which offer more flexibility for KDE. This section focuses on the latter.

Each method defines an estimator for the MISE or its asymptotic approximations based on different approaches. This gives each method different properties in terms of bias, variance, and convergence. Additionally, none of these methods have an explicit formula for the theoretical minima, $\widehat{\mathbf{H}}$. Therefore, for each method we approximate this minimum by solving an optimization problem where the optimization criterion is the particular estimation of MISE or its approximations given a sample dataset $\mathcal{D}$. This may affect the theoretical properties of the bandwidth selectors in some scenarios. 

The methods are reviewed below, along with their theoretical and practical advantages and disadvantages. For the explicit formulation of the estimators, refer to the original works. Only the key properties of the methods that impact their application on SPBN  are discussed here.

\paragraph{Unbiased Cross-Validation}\mbox{}\\
Unbiased cross-validation (UCV) (\cite{duong2005cross}), also known as least-squares cross-validation, aims to minimize the MISE. The method is based on the ISE (Equation \ref{eq:ISE}), where a leave-one-out density estimator is used to approximate this term, leading to the UCV estimator.

As an estimator, UCV is unbiased for $\text{MISE}\{\widehat{f}(\mathbf{x};\mathbf{H})\}$. Also, UCV is known for its high variance and tendency to produce undersmoothed kernel density estimates. There is no closed-form solution for the oracle bandwidth matrix estimator $\widehat{\mathbf{H}}_{\text{UCV}}$, making it necessary to solve an optimization problem for the estimation of UCV given a sample dataset $\mathcal{D}$. Consequently, UCV often exhibits multiple local optima. Despite these challenges, UCV remains a computationally efficient and competitive method.

\paragraph{Smooth Cross-Validation}\mbox{}\\
\label{sec:SCV}
Smooth cross-validation (SCV) (\cite{chacon2011unconstrained}) aims to reduce the variability of UCV by targeting an asymptotic approximation of MISE, similar to AMISE, referred to as MISE2 (\cite{chacon2011unconstrained}). SCV estimates this asymptotic approximation by introducing an independent KDE estimator with bandwidth matrix $\mathbf{G}$ for $f(\mathbf{x})$. Minimization of SCV leads to the estimator $\widehat{\mathbf{H}}_{\text{SCV}}$.

Furthermore, the bandwidth matrix $\mathbf{G}$ of the independent KDE has a significant influence on the quality of the SCV estimate. Thus, $\mathbf{G}$ must be carefully chosen to achieve optimal approximations, leading to an $s$-stage algorithm that involves a cascade of optimization procedures. At each stage, a parameter is optimized and subsequently used in the next stage until $\mathbf{G}$ is determined. This matrix is then employed in the optimization problem for minimizing SCV given the dataset $\mathcal{D}$. Specifically, the original authors recommend a two-stage procedure. For details on how the stages relate to each other and the optimization problems involved, we refer the reader to the original work of \cite{chacon2011unconstrained}.

The SCV criterion serves its purpose by reducing the variance of UCV, making it, in theory, one of the most effective methods for bandwidth selection. However, the choice of the optimization method at each stage can influence the theoretical properties of SCV and significantly increase the computational cost, which must be carefully considered in practical applications. Nonetheless, experimental evidence supports the effectiveness of the method (\cite{chacon2011unconstrained}, \cite{chacon2013comparison}, \cite{chacon2018multivariate}).

\paragraph{Plug-in} \mbox{}\\
Bias cross-validation (BCV) (\cite{scott1987biased, duong2005cross}) reduces UCV's variability as a result of introducing bias by targeting AMISE. It estimates AMISE by replacing the unknown density $f(\mathbf{x})$ with the KDE $\hat{f}(\mathbf{x};\mathbf{H})$ in the integral of the squared trace of $f$'s Hessian (see Equation \ref{eq:AMISE}).  

However, the trade-off between improved performance and higher computational cost has limited BCV to theoretical use. Alternative estimators, such as using a separate KDE $\hat{f}(\mathbf{x};\mathbf{G})$, have led to plug-in methods.

The plug-in method (PI) (\cite{chacon2010multivariate}) minimizes AMISE by using an independent KDE estimator for the unknown density $f(\mathbf{x})$. PI serves as an estimator of AMISE, and its minimization yields the bandwidth selector $\widehat{\mathbf{H}}_{\text{PI}}$.

Similar to SCV, $\mathbf{G}$ must be chosen for optimal approximations, leading to a two-stage procedure. Actually, the approach for selecting $\mathbf{G}$ in SCV is inspired by the plug-in method, and as a result, both SCV and plug-in estimators exhibit similar behavior despite targeting different discrepancy measures (\cite{chacon2011unconstrained}). Both methods have low variance and strong empirical performance (\cite{chacon2018multivariate}). 

As with SCV, there is no closed-form solution for $\widehat{\mathbf{H}}_{\text{PI}}$ and the rest of the parameters. Therefore, at each stage, a practical optimization problem is solved based on the data sample $\mathcal{D}$, making the plug-in method computationally expensive and dependent on the optimization algorithm.

\paragraph{Asymptotic theory} \mbox{}\\
\label{sec:asymptotic_theory}
As random variables, the bias and variance of the proposed bandwidth selectors have been addressed. However, their convergence properties are also critical. Under minimal conditions, all these MISE and approximations estimators and bandwidth selectors converge. This ensures convergence in MISE for $\hat{f}(\mathbf{x}, \widehat{\mathbf{H}})$ with the different bandwidth selectors, guaranteeing the KDE performance.

Convergence in MISE for the proposed bandwidth selectors follows from the convergence in MISE for the oracle bandwidths. The consistency in terms of MISE of the KDE estimator $\widehat{f}(\mathbf{x}; \mathbf{H})$ is well established in the literature for the oracle bandwidth $\mathbf{H}_{\text{AMISE}}$ under assumptions (A1)–(A3) (see Section \ref{item:conditionA1}). As $\mathbf{H}_{\text{AMISE}}$ is asymptotically equivalent to $\mathbf{H}_{\text{MISE}}$, MISE convergence for this oracle bandwidth matrix is also guaranteed.

Given that MISE is a smooth integral operator, if $\text{MISE}\{\hat{f}(\cdot; \mathbf{H}_{\text{(A)MISE}})\}$ converges to zero and $\widehat{\mathbf{H}}$ approaches $\mathbf{H}_{\text{(A)MISE}}$, then $\text{MISE}\{\hat{f}(\cdot; \widehat{\mathbf{H}})\}$ also converges, ensuring that $\hat{f}(\cdot; \widehat{\mathbf{H}})$ is a consistent estimator (\cite{chacon2018multivariate}). This applies to all proposed bandwidth selectors, ensuring confidence in their application. For a detailed theoretical study of the convergence rates, we refer the reader to Section 3.9 of \cite{chacon2018multivariate}.

\section{Bandwidth selection in semiparametric Bayesian networks}
\label{sec:3}

SPBNs combine the strengths of both parametric and non-parametric models, offering greater flexibility, as introduced in \cite{atienza2022semiparametric}. Like any BN, SPBNs decompose complex probability distributions into smaller, more manageable estimation problems using the conditional independencies encoded in the DAG $\mathcal{G}$ (see Section \ref{sec:bayesian_networks}). The main advantage of SPBNs is the ability to model CPDs either parametrically or non-parametrically, achieving better estimation results for each CPD and, consequently, for the joint probability distribution.

For the parametric approach, a linear Gaussian relationship between random variable $X_j$ and its parents $\mathbf{X}_{Pa(j)}$ is assumed:
\[X_j=\beta_{j,0}+\sum_{k\in Pa(j)}\beta_{j,k}X_{k}+\epsilon_j \hspace{5mm}\text{with }\epsilon_j\sim \mathcal{N}(0, \sigma_j^2)\]

This leads to the known LG CPD (\cite{gaussiannetworks}), the main ingredient in Gaussian BNs. For the non-parametric approach, a KDE is used. Specifically, SPBNs utilize conditional kernel density estimators (CKDEs) to model CPDs. The different ways to define a CKDE of $X_j$ conditioned to its parents $\mathbf{X}_{Pa(j)}$ are based on the next relationship:
\[f(x_j\mid \mathbf{x}_{Pa(j)})=\frac{f(x_j,\mathbf{x}_{Pa(j)})}{f(\mathbf{x}_{Pa(j)})}.\]

The differences lie on how the KDEs are defined to approximate $f(x_j,\mathbf{x}_{Pa(j)})$ and $f(\mathbf{x}_{Pa(j)})$ (\cite{rosenblatt1969conditional}, \cite{hyndman1996estimating}, \cite{fan1996estimation}, \cite{gooijer2003conditional}). Specifically, the CKDE used in SPBNs is given by Definition \ref{definition3:1}. 

\begin{df}
Let $X_j$ be a random variable, and $\mathbf{X}_{Pa(j)}$ be its parents random vector. Then, given a set of random vectors i.i.d. $(X_j^1,\mathbf{X}^1_{Pa(j)}),...,(X_j^N,\mathbf{X}^N_{Pa(j)})\sim (X_j,\mathbf{X}_{Pa(j)})$, the CKDE estimator of the conditional density of $X_j$ given $\mathbf{X}_{Pa(j)}$ is defined as:
\begin{equation}
\hat{f}(x_j\mid \mathbf{x}_{Pa(j)};\mathbf{H})=\frac{\hat{f}(x_j, \mathbf{x}_{Pa(j)}; \mathbf{H}(X_j,\mathbf{X}_{Pa(j)}))}{\hat{f}( \mathbf{x}_{Pa(j)}; \mathbf{H}(\mathbf{X}_{Pa(j)}))}=\frac{\frac{1}{N}\sum_{i=1}^NK_{\mathbf{H}(X_j,\mathbf{X}_{Pa(j)})}((x_j,\mathbf{x}_{Pa(j)})-(X_j^i,\mathbf{X}_{Pa(j)}^i))}{\frac{1}{N}\sum_{i=1}^NK_{\mathbf{H}(\mathbf{X}_{Pa(j)})}(\mathbf{x}_{Pa(j)}-\mathbf{X}_{Pa(j)}^i)}
\label{eq:CKDE}
\end{equation}

$\hat{f}(x_j, \mathbf{x}_{Pa(j)}; \mathbf{H}(X_j,\mathbf{X}_{Pa(j)}))$ and $\hat{f}(\mathbf{x}_{Pa(j)}; \mathbf{H}(\mathbf{X}_{Pa(j)}))$ are Gaussian KDEs (Equation \ref{eq:gaussiankde}), and the bandwidth matrices $\mathbf{H}(X_j,\mathbf{X}_{Pa(j)})$ and $\mathbf{H}(\mathbf{X}_{Pa(j)})$ verify:
\begin{equation}
\begin{aligned}
\mathbf{H}(X_j, \mathbf{X}_{Pa(j)}) &= \begin{bmatrix}
\alpha & \mathbf{v}^t \\
\mathbf{v} & \mathbf{M}
\end{bmatrix}, \text{where} \quad \alpha\in\mathbb{R}, \hspace{1mm}\mathbf{v}\in \mathbb{R}^{d-1},\text{ and }\mathbf{M}\in\mathbb{M}^{(d-1)\times (d-1)} \text{ is a principal submatrix.}\\
\mathbf{H}(\mathbf{X}_{Pa(j)}) &= \mathbf{M}
\end{aligned}
\label{eq:bandwidth_relationship}
\end{equation}
\label{definition3:1}
\end{df}

Given this definition, it can be proved (\cite{atienza2022semiparametric}) that:
\[\hat{f}(\mathbf{x}_{Pa(j)})=\int \hat{f}(x_j,\mathbf{x}_{Pa(j)})dx_j\hspace{5mm}\forall (x_j,\mathbf{x}_{Pa(j)}),\]
which is a sufficient condition to guarantee that $\hat{f}(x_j\mid \mathbf{x}_{Pa(j)})$ is a density function. 

CPDs are learned from data. For each CPD, the parametric approach uses the maximum likelihood estimator for linear Gaussian CPDs, with parameters $\boldsymbol{\theta}_j = \left\{\beta_{j,0}, (\beta_{j,k})_{k \in \text{Pa}(j)}, \sigma_j \right\}$, as in Gaussian BNs (\cite{geiger1994learning}).

For the CKDE CPD's bandwidth parameter, only the bandwidth matrix $\mathbf{H}(X_j, \mathbf{X}_{Pa(j)})$ must be derived given the relationship between $\mathbf{H}(X_j, \mathbf{X}_{Pa(j)})$ and $\mathbf{H}( \mathbf{X}_{Pa(j)})$ (Equation \ref{eq:bandwidth_relationship}). This is equivalent to the problem of learning the bandwidth matrix for the KDE $\hat{f}(x_j, \mathbf{x}_{Pa(j)}; \mathbf{H}(X_j,\mathbf{X}_{Pa(j)}))$.

In the original work of SPBNs the normal rule  (\cite{wand1992error}, \cite{scott2015multivariate}) is proposed:
\begin{equation}
\widehat{\mathbf{H}}_{\text{NR}}(X_j,\mathbf{X}_{Pa(j)}) = \left(\frac{4}{d+2}\right)^{\frac{1}{d+4}} N^{\frac{-2}{|Pa(j)| + 5}} \widehat{\Sigma}_{X_j,\mathbf{X}_{Pa(j)}}.
\label{eq:bandwidth_matrix_NR}
\end{equation}
where $\hat{\Sigma}_{X_j,\mathbf{X}_{Pa(j)}}$ is the estimator for the covariance matrix of $(X_j,\mathbf{X}_{Pa(j)})$. The normal rule bandwidth selector in Equation \ref{eq:bandwidth_matrix_NR} is a data-driven approach derived by minimizing AMISE (Equation \ref{eq:AMISE}). This derivation assumes that the true underlying density of $(X_j, \mathbf{X}_{Pa(j)})$ follows an independent normal distribution—an assumption that rarely holds in real-world data. However, under this assumption, AMISE can be explicitly computed, allowing for analytical derivation of the bandwidth selector $\widehat{\mathbf{H}}_{\text{NR}}$. Although fast and simple, the normal rule may lead to an oversmoothing kernel, i.e, a kernel that gives overestimated density values. In the context of machine learning, this leads to underfitting, which is undesired.

Therefore, SPBNs can benefit from the introduced bandwidth selectors, which provide a finer methodology for establishing bandwidth parameters, though they come with their drawbacks. Better estimation of the CPDs not only improves the estimation of the joint probability distribution given a fixed SPBN structure $\mathcal{G}$, but can also influence the structure learning process of SPBNs.

The structure of an SPBN is learned using adapted versions of the HC algorithm (see Section \ref{sec:bayesian_networks}) and the PC algorithm (\cite{spirtes2000causation}). Both adaptations rely on CPD learning to determine the final SPBN structure, particularly HC. The latter is used in the synthetic and real experiments of Section \ref{sec:experiments} to study the impact of bandwidth selectors on SPBN's structure learning. 

\begin{figure}[h]
\centering 
\includegraphics[scale = 0.43]{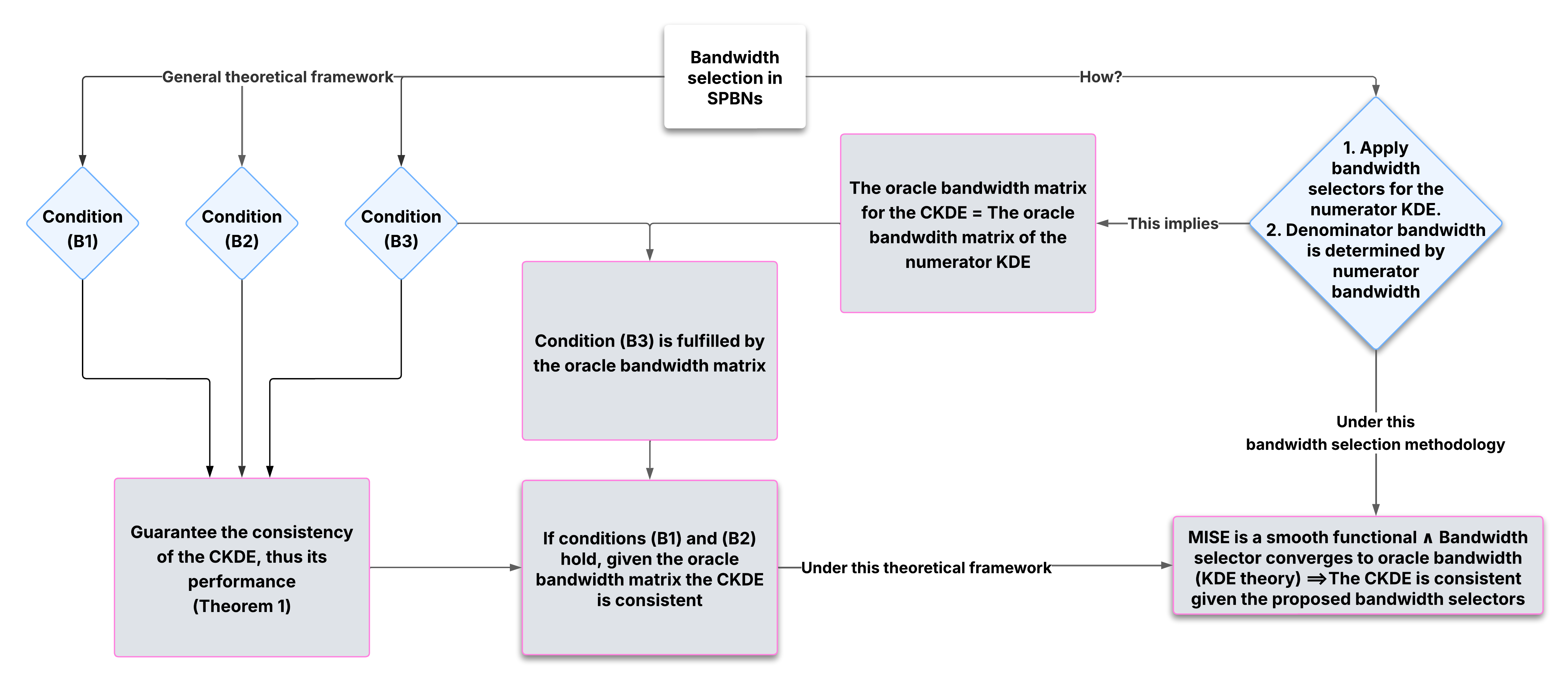}
\caption{Summary of the proposed theoretical framework for guaranteeing the performance of CKDEs with the proposed bandwidth selectors
} 
\label{fig:theoretical_framework_summary}
\end{figure}

As established previously, bandwidth selection is required only for $\mathbf{H}(X_j, \mathbf{X}_{Pa(j)})$, as $\mathbf{H}(\mathbf{X}_{Pa(j)})$ is automatically determined by this matrix. This simplifies the application of the bandwidth selectors, as they only need to be employed for the numerator KDE. Nonetheless, there is still a need to establish certain conditions to guarantee the performance of the CKDE, as the matrix $\mathbf{H}(X_j, \mathbf{X}_{Pa(j)})$ directly influences $\mathbf{H}(\mathbf{X}_{Pa(j)})$. Below, we establish the theoretical framework (see Figure \ref{fig:theoretical_framework_summary}) for developing bandwidth selectors in SPBN's CKDE, ensuring their consistency and performance when applied.

\subsection{Bandwidth selectors in SPBNs: theoretical framework}

Given the relationship between bandwidth matrices $\mathbf{H}(X_j, \mathbf{X}_{Pa(j)})$ and $\mathbf{H}(\mathbf{X}_{Pa(j)})$, we will denote the principal submatrix $\mathbf{H}(\mathbf{X}_{Pa(j)}) = \mathbf{M}$ by $\mathbf{M}(\mathbf{H}(X_j, \mathbf{X}_{Pa(j)}))$. In the following, we establish the conditions necessary for applying the bandwidth selectors and ensuring the performance of the CKDE:

\begin{itemize}[label=\textbf{(\arabic*)}, ref=\textbf{(\arabic*)}]
    \item [(B1)] \label{item:conditionB1} The density function $f(x_j,\mathbf{x}_{Pa(j)})$ is square integrable and twice differentiable, with all of its second order partial derivatives bounded, continuous and square integrable.
    \item [(B2)] \label{item:conditionB2} The density function $f(\mathbf{x}_{Pa(j)})$ is square integrable.
    \item [(B3)] \label{item:conditionB3} The bandwidth matrices $\mathbf{H}(X_j, \mathbf{X}_{Pa(j)}) = \mathbf{H}(X_j, \mathbf{X}_{Pa(j)})_N$  form a sequence of positive definite, symmetric matrices such that $$vec\mathbf{H}(X_j, \mathbf{X}_{Pa(j)})\rightarrow 0 \text{  and  }  
 N^{-1}|\mathbf{H}(X_j, \mathbf{X}_{Pa(j)})|^{-1/2}\rightarrow 0 \text{   when } N\rightarrow \infty.$$
\end{itemize}

Given these conditions, the MISE of the CKDE's numerator,  
\(\text{MISE}\{\hat{f}(x_j,\mathbf{x}_{Pa(j)};\mathbf{H}(X_j, \mathbf{X}_{Pa(j)}))\}\), can be decomposed into AMISE and MISE2. This decomposition allows for the application of bandwidth selection techniques to optimize the estimator's performance. Furthermore, the CKDE in the SPBN can be shown to be a consistent estimator, ensuring its reliability in density estimation.

\begin{thm}
\label{theorem1}
Given a SPBN's CKDE, if conditions (B1)-(B3) hold, then the CKDE (Definition \ref{definition3:1}) is a consistent estimator of $f(x_j\mid \mathbf{x}_{Pa(j)})$:
\[\hat{f}(x_j\mid \mathbf{x}_{Pa(j)};\mathbf{H})\xrightarrow{P}f(x_j\mid \mathbf{x}_{Pa(j)})\]
\end{thm}

The proof of Theorem \ref{theorem1} can be found in Appendix \ref{apd:first}. As discussed in Section \ref{sec:2.1}, these conditions can be relaxed. Nonetheless, we assume these conditions, as they are necessary for applying bandwidth selection methods in CKDEs. 

Furthermore, within this framework, we refer to the oracle bandwidth matrices of the CKDE,  
\(\mathbf{H}_{\text{MISE}}\) and \(\mathbf{H}_{\text{AMISE}}\), as the oracle bandwidth matrices of  
\(\mathbf{H}(X_j, \mathbf{X}_{Pa(j)})\), since \(\mathbf{H}(\mathbf{X}_{Pa(j)})\) is automatically determined by this matrix.  This implies that the oracle bandwidth matrix is of order $ O(N^{-2/(d+4)})$ (see Section~\ref{sec:2.1}). Consequently, condition (B3) is satisfied by the oracle bandwidth. Therefore, whenever conditions (B1) and (B2) also hold, the CKDE is a consistent estimator (Theorem~\ref{theorem1}).

In practice, the oracle bandwidth is unknown, which motivates the use of the proposed bandwidth selectors. For each selector \(\widehat{\mathbf{H}}\) targeting \(\mathbf{H}(X_j, \mathbf{X}_{Pa(j)})\), convergence to the oracle bandwidth is well established (see Section~\ref{sec:asymptotic_theory}). We demonstrate the consistency of the CKDE when these selectors are used, by extending results from multivariate KDE theory.

On one hand, Theorem~\ref{theorem1} establishes the consistency of CKDE based on the convergence in MISE of both the numerator and denominator KDEs. On the other hand, in standard KDE theory, consistency under a proposed bandwidth selector follows from two facts: (i) the MISE converges under the oracle bandwidth, and (ii) the selector itself converges to the oracle bandwidth. This conclusion holds because the MISE is a smooth functional operator (\cite{chacon2010multivariate}).

We adopt a similar line of reasoning to establish the consistency of the CKDE when using the proposed bandwidth selectors. Notably, the oracle bandwidth for the CKDE coincides with that of the numerator, and the convergence of the proposed selectors to this oracle bandwidth is well established in KDE theory. Given this convergence, the MISE consistency of both the numerator and denominator KDEs follows naturally, owing to the smoothness of the MISE functional.

In the proof of Theorem \ref{theorem1} (see Appendix \ref{apd:first}), we show that the consistency of the CKDE stems from the MISE-based consistency of the numerator and denominator estimators. Therefore, the CKDE is consistent when any of the proposed selectors is used.

In summary, our analysis provides a rigorous theoretical foundation that ensures the reliability and asymptotic consistency of CKDE estimation under the proposed bandwidth selection methods.

\section{Experiments on the impact of bandwidth selection on SPBNs}
\label{sec:experiments}
To study the impact of bandwidth selection on SPBNs, we have designed experiments involving both synthetic and real-world datasets. In the synthetic framework, we explored the parameter and structure learning paradigms for SPBNs. We demonstrate that the introduced bandwidth selection methods (UCV, SCV, PI) outperform the current normal rule method across both learning paradigms and various scenarios. However, it is important to note that while these methods often show superior performance, there are instances where the normal rule remains competitive, highlighting the context-dependent nature of their effectiveness. To evaluate the applicability of these findings in real-world scenarios, we conducted experiments on datasets from the UCI machine learning repository. By combining both synthetic and real-world analyses, we provide a comprehensive understanding of the advantages of the proposed bandwidth selection method in SPBNs.

For the implementation of all experiments, the \texttt{PyBNesian} package (\cite{atienza2022pybnesian}) has been extended. Specifically, UCV was already implemented in the package, though it was not available for the learning of SPBNs, so we modified it accordingly. Additionally, SCV and PI have been introduced, and, like UCV, they leverage GPU acceleration to alleviate the burden of their high computational cost. For the optimization task that each bandwidth selection poses given the dataset $\mathcal{D}$ (see Section \ref{sec:bandwidth_selection}), the Nelder-Mead algorithm (\cite{nelder1965simplex}) has been used for SCV and PI, following the implementation of UCV.

\subsection{Synthetic data experiments}
\label{sec:synthetic_experiments}
\subsubsection{Parameter learning paradigm}
\label{subsectionPL}
In this section, we analyze the introduction of the proposed state-of-the-art methods, demonstrating their ability to address the limitations of the normal rule for CPDs learning given a fixed structure $\mathcal{G}$. However, as discussed in Section \ref{sec:bandwidth_selection}, cross-validation and plug-in methods also exhibit certain limitations and properties, which will be reflected in the experiments.

To evaluate the performance of these methods, we designed three distinct SPBN structures varying the number of nodes, arcs, and node types (see Figure \ref{fig:all_networks}). For each SPBN size—five, ten, and fifteen nodes—we defined three different joint probability distributions, with each successive distribution being less smooth than the previous one. In the rest of the text, they will be referred to as smooth, medium, and rough density. Additionally, we considered four different sample sizes: $200$, $2000$, $10000$, and $20000$. Due to the high computational cost of SCV and PI methods, we limited their evaluation to scenarios with $200$ and $2000$ instances, as larger sample sizes were computationally infeasible, a key drawback of these methods. For each scenario, ten different training datasets were sampled according to their respective sample sizes, and a corresponding validation dataset of $1000$ instances was generated for performance evaluation.

The errors MISE and AMISE cannot be obtained due to the complexity of the joint probability distributions. Since the proposed bandwidth selection methods aim to minimize MISE or AMISE, the log-likelihood serves as an appropriate proxy measure for performance, as reducing these errors directly leads to a reduction in the log-likelihood error. Therefore, the log-likelihood has been proposed as a measure of the overall performance of the bandwidth selection methods in each scenario, as it is also well-suited for the context of learning probability distribution functions. To account for potential overfitting and underfitting, the absolute error with respect to the known ground-truth log-likelihood is computed for each repetition and used in the subsequent analysis.  

To assess differences in performance, a statistical analysis was conducted, addressing the multiple comparison problem inherent to the proposed design. The standard approach for this scenario often follows the methodology outlined by \cite{demvsar2006statistical} and \cite{garcia2008extension}. However, critiques of rank-based methods apply to our analysis as well. Consequently, we adopted the ideas of \cite{benavoli2016should}, which advocate for the use of correction methods such as Bonferroni or Bergmann-Hommel applied to pairwise tests like the Wilcoxon test for medians. To avoid assumptions about the underlying error distributions and due to non-symmetry in most of the error distributions, we employed a permutation test for medians (\cite{mangiafico2016summary}) combined with the Bergmann-Hommel correction procedure, as it is the least conservative method among those considered. The significance level chosen for all tests is $\alpha = 0.05$ as standard.

\subsubsection{Five nodes SPBN}
\label{sec:fixed_five_nodes}
\begin{figure}[h]
    \centering
    \begin{subfigure}[b]{0.12\linewidth}
        \includegraphics[width=\linewidth]{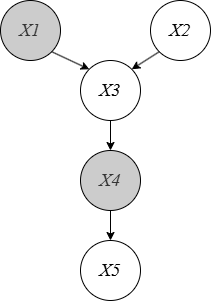}
        \caption{Five nodes}
        \label{fig:synthetic1}
    \end{subfigure}
    \hspace{7mm} 
    \begin{subfigure}[b]{0.28\linewidth}
        \includegraphics[width=\linewidth]{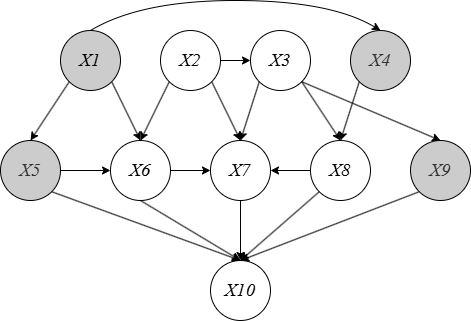}
        \caption{Ten nodes}
        \label{fig:synthetic2}
    \end{subfigure}
    \hspace{7mm} 
    \begin{subfigure}[b]{0.3\linewidth}
        \includegraphics[width=\linewidth]{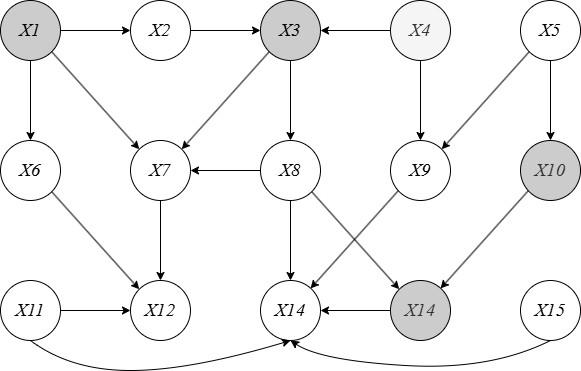}
        \caption{Fifteen nodes}
        \label{fig:synthetic3}
    \end{subfigure}
    \caption{Synthetic SPBNs. White nodes denote CKDE type, and gray shaded nodes correspond to the LG type.}
    \label{fig:all_networks}
\end{figure}

The structure for this case follows the original proposal in \cite{atienza2022semiparametric}, as shown in Figure \ref{fig:synthetic1}. The smooth density is defined as in the original work for SPBNs. Non-Gaussian conditional densities are defined through nonlinear relationships with Gaussian noise added. To achieve less smooth densities with more variability, including higher peaks and lower troughs, we introduce Gaussian mixtures. By adding components to these mixtures, we obtain denser distributions with greater variability. These distributions are then used in the conditional densities to generate the other two joint densities (see Appendix \ref{apd:second} for an example). 

Table \ref{tab:results_fixed_structure} shows a summary of the performance in terms of absolute error in the log-likelihood. The best results for each scenario (smooth, medium, rough) are highlighted in bold. Symbols $\dagger$ \footnote{$\dagger$ This symbol represents significant difference concerning the normal rule in terms of medians whenever the method has a lower median} and $\ddagger$\footnote{$\ddagger$ This symbol represents significant difference with the second best algorithm in terms of medians whenever the best is the normal rule} are used to complement with the statistical analysis.

As shown in Table \ref{tab:results_fixed_structure}, for the five nodes SPBN, NR outperforms the proposed methods in terms of medians only for $N=200$ and the smooth density. However, there is no statistical difference compared to the second-best method, UCV, which achieves the lowest result overall. In all other cases, the proposed methods generally outperform the normal rule and show a significant difference in medians. UCV achieves the best overall performance; however, the results from this method exhibit high variability compared to the normal rule, as indicated by the substantial difference between the median and the lowest value.

\begin{table}[h]
\centering
\caption{Synthetic data experiments absolute error summary for the log-likelihood across different node structures in learning parameters experiments}
\label{tab:results_fixed_structure}
\tiny 
\setlength{\tabcolsep}{4pt} 
\begin{tabular}{@{}c p{1.2cm} p{1cm} c c c c c c c c@{}}
\toprule
\textbf{Nodes} & \textbf{Density} & \textbf{Bandwidth Method} & \multicolumn{2}{c}{$\mathbf{N = 200}$} & \multicolumn{2}{c}{$\mathbf{N = 2000}$} & \multicolumn{2}{c}{$\mathbf{N = 10000}$} & \multicolumn{2}{c}{$\mathbf{N = 20000}$} \\
\midrule
& & & \textbf{Median} & \textbf{Lowest} & \textbf{Median} & \textbf{Lowest} & \textbf{Median} & \textbf{Lowest} & \textbf{Median} & \textbf{Lowest} \\
\midrule
\multirow{12}{*}{5}  & \multirow{4}{*}{Smooth}   & NR  & $\mathbf{\phantom{0}638.68}$ & $558.31$ & $347.06$ & $284.26$ & $217.85$ & $172.51$ & $166.54$ & $148.29$ \\
& & UCV & $\phantom{0}745.67$ & $\mathbf{473.18}$ & $\mathbf{\phantom{0}242.75}^{\dagger}$& $\mathbf{176.98}$ & $\mathbf{\phantom{0}152.95}^{\dagger}$ & $\mathbf{\phantom{0}69.82}$ & $\mathbf{\phantom{0}96.03}^{\dagger}$ & $\mathbf{\phantom{0}59.45}$ \\
& & SCV & $\phantom{0}823.71$ & $610.41$ & $833.27$ & $370.93$ & -- & -- & -- & -- \\
& & PI  & $1584.55$ & $696.09$ & $1883.31$ & $371.64$ & -- & -- & -- & -- \\
\cmidrule(l){2-11}
& \multirow{4}{*}{Medium} & NR  & $1019.88$ & $\phantom{0}982.21$ & $\phantom{0}577.65$ & $\phantom{0}553.16$ & $354.00$ & $326.95$ & $276.39$ & $251.28$ \\
& & UCV & $\mathbf{\phantom{00}355.87}^{\dagger}$ & $\mathbf{\phantom{0}299.29}$ & $\mathbf{\phantom{00}120.06}^{\dagger}$ & $\mathbf{\phantom{00}80.18}$ & $\mathbf{\phantom{00}65.71}^{\dagger}$ & $\mathbf{\phantom{0}38.84}$ & $\mathbf{\phantom{00}49.03}^{\dagger}$ & $\mathbf{\phantom{0}19.77}$ \\
& & SCV & $1078.58$ & $\phantom{0}983.70$ & $\phantom{0}368.55$ & $\phantom{0}251.12$ & -- & -- & -- & -- \\
& & PI  & $3678.83$ & $1757.17$ & $1945.60$ & $1904.73$ & -- & -- & -- & -- \\
\cmidrule(l){2-11}
& \multirow{4}{*}{Rough} & NR  & $1604.17$ & $1552.90$ & $1170.63$ & $1112.63$ & $859.36$ & $807.45$ & $728.23$ &  $683.12$ \\
& & UCV & $2112.65$ & $\mathbf{1204.36}$ & $\mathbf{\phantom{00}260.73}^{\dagger}$ & $\mathbf{\phantom{00}19.30}$ & $\mathbf{\phantom{0}118.32}^{\dagger}$ & $\mathbf{\phantom{0}62.31}$ & $\mathbf{\phantom{0}224.47}^{\dagger}$ & $\mathbf{111.03}$ \\
& & SCV & $\mathbf{\phantom{0}1473.73}^{\dagger}$ & $1350.33$ & $\phantom{0}900.92$ & $\phantom{0}565.90$ & -- & -- & -- & -- \\
& & PI  & $2881.83$ & $2298.17$ & $2615.32$ & $2240.65$ & -- & -- & -- & -- \\
\midrule

\multirow{12}{*}{10}  
    & \multirow{4}{*}{Smooth} 
    & NR & $\mathbf{\phantom{0}10130.03}$ & $\phantom{0}8146.28$ & $\mathbf{\phantom{0}7635.55}$ & $\phantom{0}6494.49$ & $7752.39$ & $5988.05$ & $7444.33$ & $5767.01$ \\
    & & UCV & $267441.14$ & $\mathbf{\phantom{0}5898.65}$ & $27999.40$ & $\mathbf{\phantom{0}3117.27}$ & $\mathbf{6516.09}$ & $\mathbf{1997.93}$ & $\mathbf{4395.64}$ & $\mathbf{1728.26}$ \\
    & & SCV & $\phantom{0}31592.43$ & $\phantom{0}9260.41$ & $\phantom{0}9419.72$ & $\phantom{0}7490.27$ & -- & -- & -- & -- \\
    & & PI  & $\phantom{0}16230.75$ & $12039.01$ & $12774.71$ & $11281.84$ & -- & -- & -- & -- \\
\cmidrule(l){2-11}
    & \multirow{4}{*}{Medium} 
    & NR & $\mathbf{\phantom{00}2104.24^{\ddagger}}$ & $\mathbf{2073.93}$ & $\phantom{0}1281.16$ & $\phantom{0}1250.87$ & $899.22$ & $869.84$ & $770.49$ & $743.66$ \\
    & & UCV & $\phantom{0}3906.32$ & $3015.29$ & $\mathbf{\phantom{00}1211.67^{\dagger}}$ & $\mathbf{\phantom{0}1012.23}$ & $\mathbf{\phantom{0}621.23^{\dagger}}$ & $\mathbf{565.73}$ & $\mathbf{\phantom{0}499.35^{\dagger}}$ & $\mathbf{457.89}$ \\
    & & SCV & $\phantom{0}4551.93$ & $3682.12$ & $\phantom{0}2756.79$ & $\phantom{0}2231.13$ & -- & -- & -- & -- \\
    & & PI  & $22033.32$ & $4852.40$ & $35084.64$ & $13162.63$ & -- & -- & -- & -- \\
\cmidrule(l){2-11}
    & \multirow{4}{*}{Rough} 
    & NR & $\phantom{000}4116.38$ & $\mathbf{\phantom{00}195.88}$ & $\phantom{0}3198.42$ & $\mathbf{\phantom{00}86.18}$ & $\mathbf{5087.19}$ & $1237.57$ & $\mathbf{5424.45}$ &  $1560.93$ \\
    & & UCV & $2824039.59$ & $84748.57$ & $35040.78$ & $2500.14$ & $7357.44$ & $\mathbf{\phantom{0}126.54}$ & $8574.05$ & $\mathbf{\phantom{0}985.84}$ \\
    & & SCV & $\phantom{00}14510.72$ & $\phantom{0}5416.94$ & $\mathbf{\phantom{0}2751.42}$ & $\phantom{0}282.96$ & -- & -- & -- & -- \\
    & & PI  & $\mathbf{\phantom{000}2128.21}$ & $\phantom{00}772.56$ & $\phantom{0}3100.01$ & $\phantom{00}95.48$ & -- & -- & -- & -- \\

\midrule

\multirow{12}{*}{15} 
    & \multirow{4}{*}{Smooth} 
    & NR & $\mathbf{\phantom{0}29082.58^{\ddagger}}$ & $\mathbf{28371.56}$ & $26210.88$ & $26077.55$ & $24572.82$ & $24448.32$ & $24081.15$ & $23954.17$ \\
    & & UCV & $67055.49$ & $39875.73$ & $\mathbf{\phantom{0}24518.47^{\dagger}}$ & $\mathbf{22649.97}$ & $\mathbf{\phantom{0}21109.66^{\dagger}}$ & $\mathbf{19710.95}$ & $\mathbf{\phantom{0}19478.53^{\dagger}}$ & $\mathbf{18887.07}$ \\
    & & SCV & $37080.12$ & $29765.59$ & $27104.46$ & $26566.11$ & -- & -- & -- & -- \\
    & & PI  & $33632.26$ & $32638.78$ & $31081.98$ & $28542.52$ & -- & -- & -- & -- \\
\cmidrule(l){2-11}
    & \multirow{4}{*}{Medium} 
    & NR & $\phantom{00}208876.75$ & $\phantom{00}42901.96$ & $\mathbf{\phantom{0}23664.89}$ & $21429.55$ & $\mathbf{22410.29}$ & $21023.31$ & $\mathbf{\phantom{0}21685.96^{\ddagger}}$ & $20734.40$ \\
    & & UCV & $20928676.18$ & $1567318.34$ & $272165.16$ & $\mathbf{17140.79}$ & $40994.63$ & $\mathbf{14060.48}$ & $41483.02$ & $\mathbf{13400.86}$ \\
    & & SCV & $\phantom{000}84825.02$ & $\mathbf{\phantom{00}29331.39}$ & $\phantom{0}26267.63$ & $22065.35$ & -- & -- & -- & -- \\
    & & PI  & $\mathbf{\phantom{000}73041.98}$ & $\phantom{00}32145.40$ & $\phantom{0}28305.97$ & $25510.72$ & -- & -- & -- & -- \\
\cmidrule(l){2-11}
    & \multirow{4}{*}{Rough} 
    & NR & $\mathbf{\phantom{00}32503.46}$ & $\mathbf{\phantom{0}29803.79}$ & $\mathbf{\phantom{00}28043.95^{\ddagger}}$ & $27228.12$ & $\mathbf{\phantom{0}26445.39^{\ddagger}}$ & $26103.48$ & $25983.32$ & $25679.12$ \\
    & & UCV & $1618180.29$ & $114395.64$ & $618036.72$ & $\mathbf{27055.33}$ & $42725.73$ & $\mathbf{20563.16}$ & $\mathbf{25601.22}$ & $\mathbf{19565.64}$ \\
    & & SCV & $\phantom{00}93643.87$ & $\phantom{0}31979.11$ & $\phantom{0}43962.01$ & $30841.39$ & -- & -- & -- & -- \\
    & & PI  & $\phantom{00}36620.47$ & $\phantom{0}33698.43$ & $\phantom{0}31632.86$ & $29849.46$ & -- & -- & -- & -- \\

\bottomrule
\end{tabular}
\end{table}

NR exhibits remarkable stability in terms of result variability, producing competitive outcomes across scenarios. In some cases, it even outperforms SCV and PI in terms of median values, as observed with the smooth density. More competitive results were expected for these two methods, given their design as bandwidth selectors. This may be due to both the optimization procedure and their tendency to overfit.

\begin{figure}[h]
\centering 

\includegraphics[scale=0.27]{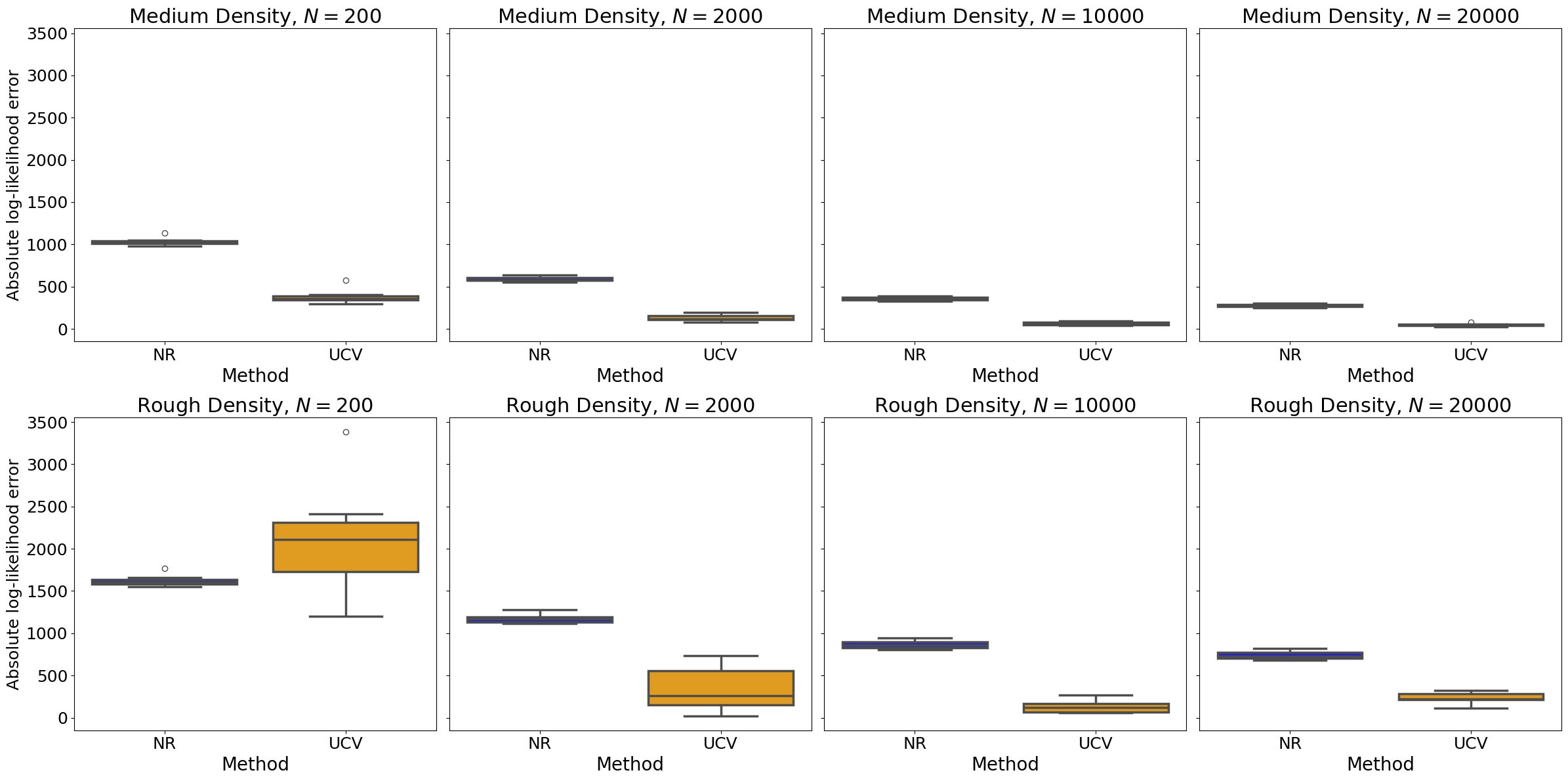}
\caption{Synthetic data experiments boxplots of the absolute log-likelihood error for the medium and rough densities for NR and UCV across different scenarios} 
\label{fig:boxplot_variability}
\end{figure}

As the conditional densities decrease in smoothness, the proposed methods tend to yield worse results due to the increased complexity. This also exacerbates the inherent issues of each method, such as the high variability of UCV. NR, on the other hand, demonstrates more robustness, maintaining competitive results with low variability due to its design. Nonetheless, this same design leads to oversmoothing and a bias toward minimizing the MISE, resulting in limited improvements compared to other methods as the available information (i.e., sample size) increases. Also, the increasing available information reduces the drawbacks of the proposed methods, particularly the variability of UCV (see Figure \ref{fig:boxplot_variability}).

\begin{figure}[h]
\centering 
\includegraphics[scale=0.27]{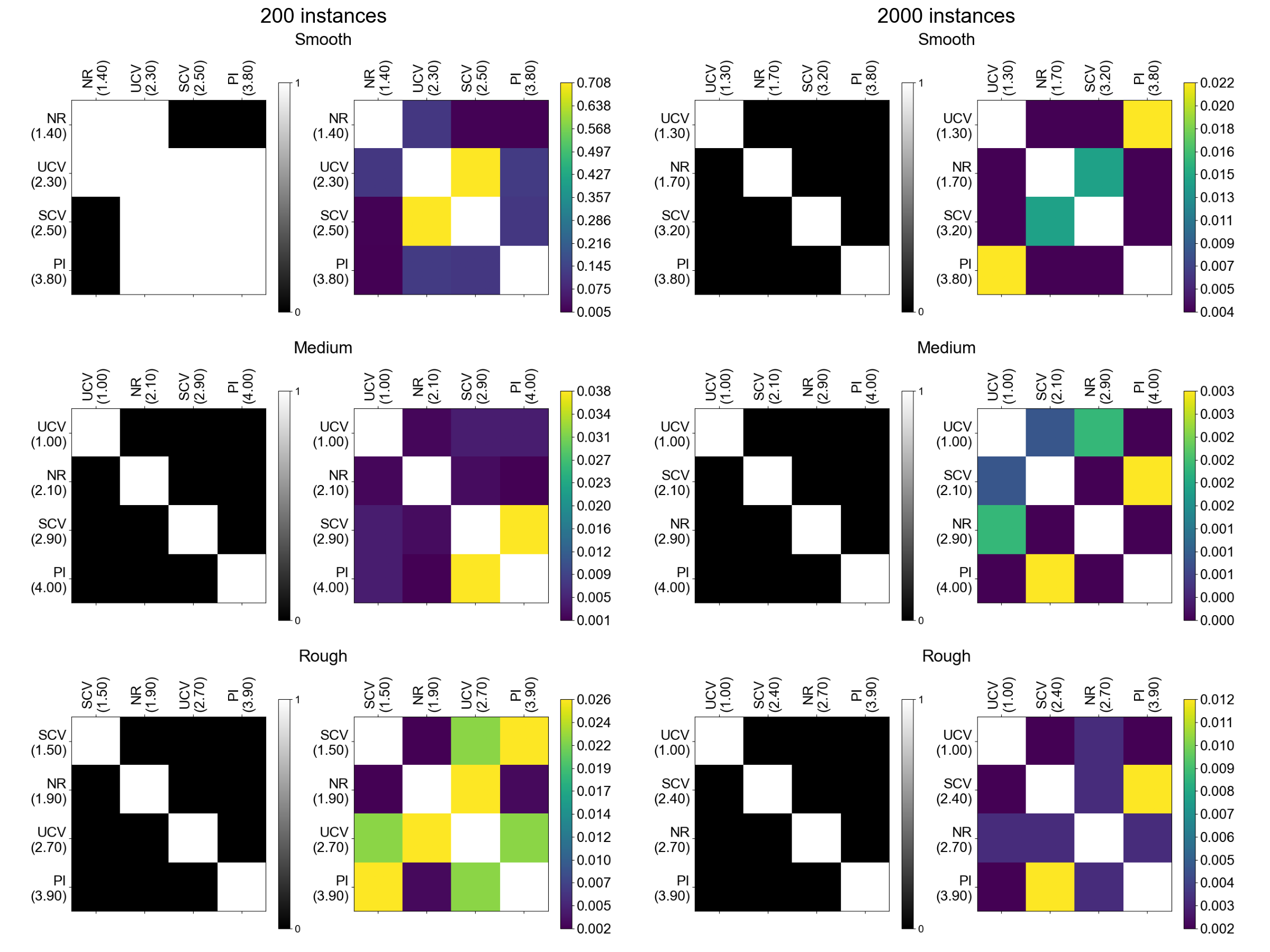}
\caption{Synthetic data experiments hypothesis acceptance (left) and p-value (right) matrices for the five nodes SPBN in learning parameters experiments} 
\label{fig:sa5}
\end{figure}

Figure \ref{fig:sa5} presents a detailed summary of the statistical analysis of differences in medians, displaying the acceptance and p-value matrices. The matrix axes represent the bandwidth selectors, ordered by their rankings and displaying their median rank. To understand the information conveyed in the figure, we focus on the smooth density case with $N = 2000$. In the hypothesis acceptance matrix, all methods exhibit statistically significant differences in medians, as all cells are black. The p-value matrix indicates that UCV is generally the best-performing method, with a median rank of 1.3. Moreover, p-values darken as they decrease, highlighting the most notable differences between the best and worst methods, specifically NR, UCV, versus SCV, PI. The statistical analysis is further complemented by Table \ref{tab:results_fixed_pvalues}, which presents p-values for the statistical tests comparing NR and UCV in high-sample-size scenarios. This structure is maintained throughout the remaining synthetic experiments.

\begin{table}[h] 
\centering
\caption{Synthetic data experiments p-values for the permutation test between NR and UCV across different network sizes in learning parameters experiments}
\label{tab:results_fixed_pvalues}
\footnotesize 
\setlength{\tabcolsep}{4pt} 
\begin{tabular}{@{}c p{1.5cm} cc@{}}
\toprule
\textbf{Nodes} & \textbf{Density} & \multicolumn{2}{c}{$\mathbf{N}$} \\
\cmidrule(l){3-4}
                     &                 & \textbf{10000} & \textbf{20000} \\
\midrule
\multirow{3}{*}{5}   & Smooth           & $0.0360$  & $0.0048$ \\
                     & Medium         & $0.0004$ & $0.0002$ \\
                     & Rough         & $0.0002$ & $0.0008$ \\
\midrule
\multirow{3}{*}{10}  & Smooth           & $0.6906$ & $0.1002$ \\
                     & Medium         & $0.0008$ & $0.0010$ \\
                     & Rough         & $0.3346$ & $0.1146$ \\
\midrule
\multirow{3}{*}{15}  & Smooth           & $0.0010$ & $0.0004$ \\
                     & Medium         & $0.0612$ & $0.0206$ \\
                     & Rough         & $0.0098$ & $0.0844$ \\
\bottomrule
\end{tabular}
\end{table}

Figure \ref{fig:sa5} and Table \ref{tab:results_fixed_pvalues} show that there are always pairwise differences between methods, except in the smooth density with a sample size of 200. In this case, no significant pairwise differences are found between the NR, UCV, and SCV. Since we are using permutation tests, the p-value can be interpreted as a measure of effect size. As we observe, p-values are very low in most of cases (darker colors). This indicates a substantial difference in the pairwise median comparisons.

\subsubsection{Ten nodes SPBN}
The structure for this case is illustrated in Figure \ref{fig:synthetic2}. It is designed to include the maximum possible combination of different types of nodes and varying numbers of parents (from $1$ to $5$). The definitions of the conditional densities follow the same approach as those in the five nodes network (Section \ref{sec:fixed_five_nodes}). 

At first glance, Table \ref{tab:results_fixed_structure} reveals worse results in terms of error due to the increased complexity. While the proposed bandwidth selectors still achieve the best results overall, NR outperforms them in more cases. For $N=200$, the best median results for the smooth and medium densities are achieved by NR. A significant difference with the second-best method is observed in only one case. For the rough density, PI achieves the best performance, but it does not show significant differences compared to NR. 

For the $2000$ sample size case, the roles are reversed as NR is outperformed by UCV and SCV for the less smooth densities. Nonetheless, only UCV achieves statistically significant results. For the highest sample sizes ($10000$ and $20000$), UCV outperforms NR for the smooth and medium densities, though significant differences are observed only in the latter case. For the rough density, NR achieves the best performance. However, the stagnation in improvement for NR becomes evident when compared to UCV, whose errors decrease as the sample size increases. The increased variability of UCV due to complexity is also apparent, particularly for the smooth density, where a large difference exists between the median and lowest values for the 200 and 2000 sample sizes. Notably, UCV achieves the lowest result for these cases. This variability must be considered for practical applications.

\begin{figure}[h]
\centering 

\includegraphics[scale=0.27]{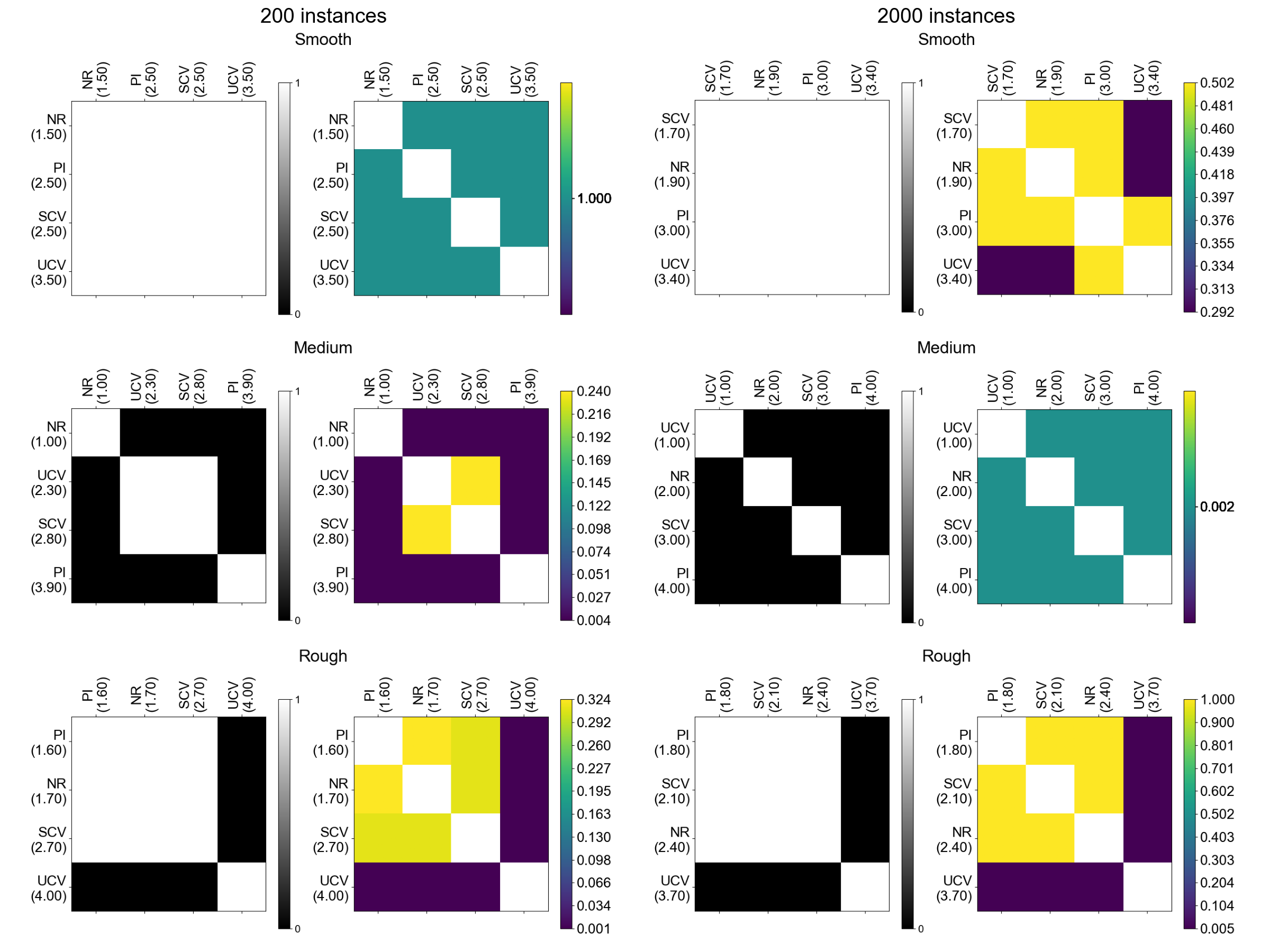}
\caption{Synthetic data experiments hypothesis acceptance (left) and p-value (right) matrices for the ten nodes SPBN in learning parameters experiments} 
\label{fig:sa10}
\end{figure}

In Figure \ref{fig:sa10} and Table \ref{tab:results_fixed_pvalues}, the smooth density shows no significant differences between methods, as all p-values are high. For the medium density, NR performs best at 200 samples, showing significant differences compared to the proposed methods. UCV and SCV show no significant difference from each other. From 2000 samples onward, UCV achieves the best performance for this density, with significant differences between all methods and very low p-values, indicating a larger effect size. In the rough density scenario, there is no significant difference between SCV, PI, and NR, although PI performs best at 200 and SCV at 2000 in terms of medians. Also in this case, for larger sample sizes, even though NR outperforms UCV there is no significant difference. The drastic reduction in p-values for UCV and NR in the smooth and rough densities shown in Table \ref{tab:results_fixed_pvalues} highlights UCV’s improved performance and reduced variability with increasing sample size, emphasizing its suitability for large-sample contexts.

To sum up, the increased complexity impacts the proposed methods. However, they still outperform NR in most cases, though significant differences are observed less frequently. NR remains robust and competitive, yielding better results in more scenarios than the five nodes case but with significant differences found in only one case. Notably, the proposed methods, particularly UCV, show pronounced performance improvements with increasing sample size, making UCV particularly suitable for large-sample contexts.

\subsubsection{Fifteen nodes SPBN}
The design of the fifteen nodes SPBN (Figure \ref{fig:synthetic3}) builds upon the ten nodes case, with increased complexity impacting the results presented in Table \ref{tab:results_fixed_structure}. This complexity manifests in higher lowest errors and medians across scenarios. We can also see how NR outperforms the proposed methods in half of the cases, showing significant differences in four scenarios. In addition, we find significant differences when the proposed methods outperform NR, specifically with UCV. Meanwhile, PI and SCV demonstrate consistent but less competitive results, with their performance standing out only in the $N=200$ scenario for the medium density.

The stagnation of NR compared to the proposed methods, particularly UCV, becomes more pronounced. For the smooth density, we observe a clear reduction in error as the sample size increases, leading to significant differences between UCV and NR. The same comparison in improvements applies to the rough density, where UCV only outperforms NR in the $N=20000$ case. To further validate this pattern, we conducted an experiment with  $N=40000$, yielding median values of $25522.09$ with NR and $23701.48$ with UCV. The permutation test resulted in a p-value of $0.01$, indicating a statistically significant difference at the $\alpha = 0.05$ level in favor of UCV. The same experiment was also performed for the medium density, where the median absolute error for UCV decreased from $41483.02$ to $24592.38$, and the variability was reduced, following the same trend. However, NR still outperforms UCV in this case.

\begin{figure}[h]
\centering 

\includegraphics[scale=0.27]{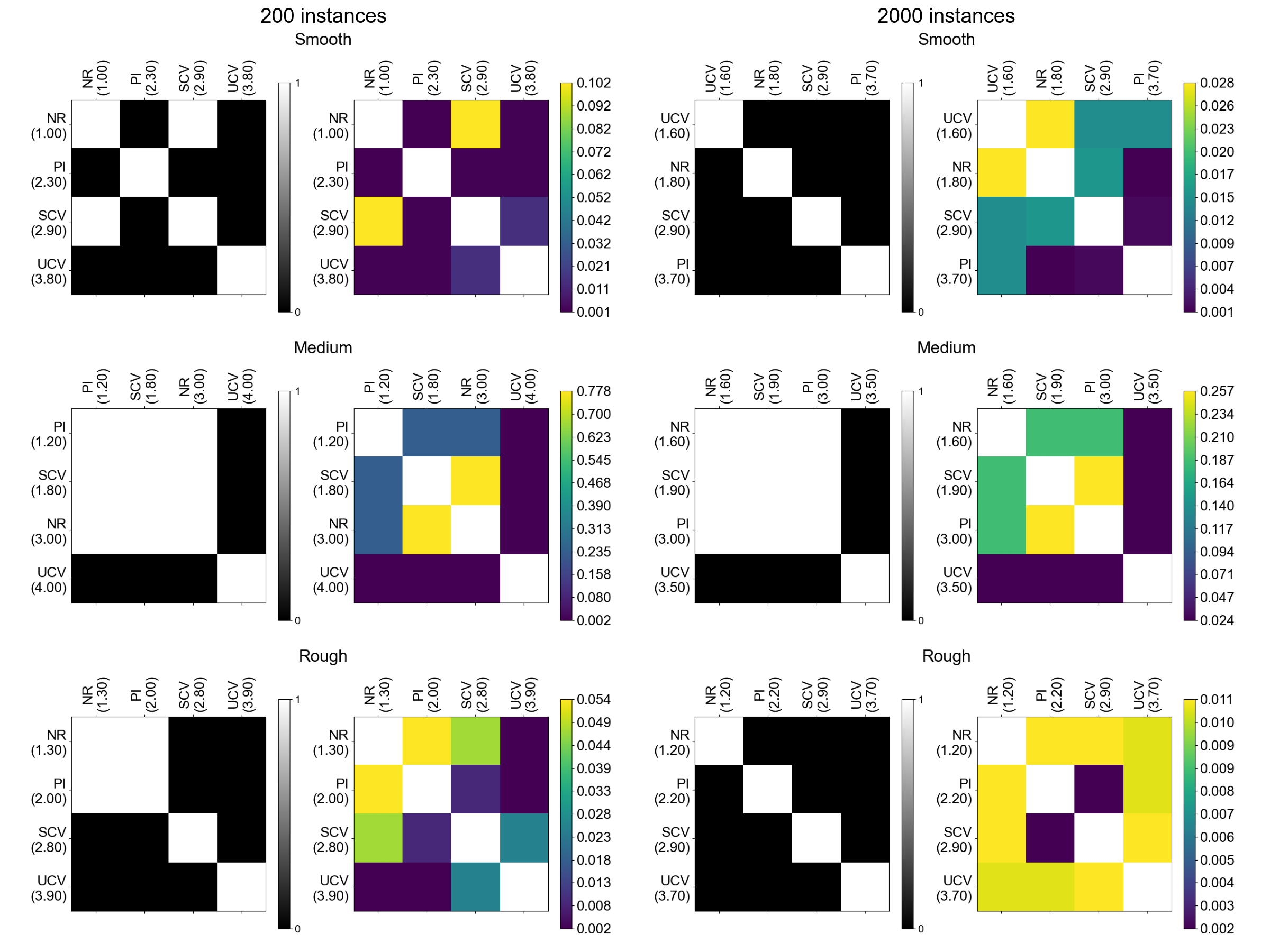}
\caption{Synthetic data experiments hypothesis acceptance (left) and p-value (right) matrices for the fifteen nodes SPBN in learning parameters experiments} 
\label{fig:sa15}
\end{figure}

Figure \ref{fig:sa15} and Table \ref{tab:results_fixed_pvalues} present a detailed statistical analysis. In Figure \ref{fig:sa15}, significant differences between all methods are observed in only two cases: the 2000 sample size for both the smooth and rough densities. Notably, in the smooth density with $N=200$, the normal rule is significantly different from all proposed methods but SCV. In other cases, no significant differences are generally observed, with the best-performing method varying. When significant differences are found, p-values are consistently low, indicating a large effect size, as seen in Table \ref{tab:results_fixed_structure}. 

Furthermore, Table \ref{tab:results_fixed_pvalues} shows overall lower p-values compared to the ten nodes case, suggesting a stronger effect size in this context. However, in two instances, p-values are slightly above the significance level ($\alpha=0.05$), suggesting weak but non-negligible evidence against the null hypothesis. This is coherent with the median results presented for those cases in Table \ref{tab:results_fixed_structure}. All results align with the general pattern of complexity influencing the proposed bandwidth selectors, as evident in the decreasing significance of differences between methods. Additionally, the p-values show the stagnation of NR compared to the improvements, particularly against UCV.

\subsection{Network structure learning}
\label{sec:synthetic_network}
Following the performance of the proposed bandwidth selection methods in parameter learning across various scenarios, this section examines their impact on structure learning. Specifically, we focus on the HC algorithm, where parameter inference is required at each iteration to compute the log-likelihood guiding the search. Thus, we investigate how bandwidth selection methods influence the search process, as they form an integral part of the parameter learning framework. 

The experiments in this section follow the structure design outlined in Section \ref{subsectionPL}. Specifically, we aim to learn three different network structures (see Figure \ref{fig:all_networks}) under their respective scenarios. In addition to evaluating absolute error in the log-likelihood, we assess the performance in recovering the original structure using the structural Hamming distance (SHD) (\cite{tsamardinos2006max}). SHD quantifies structural differences relative to the ground truth, accounting for equivalence within the BN class. A class of a BN refers to the set of all BNs that encode the same joint probability distribution. Formally, two BNs $\mathcal{B}_1$ and $ \mathcal{B}_2$ belong to the same class if they are Markov equivalent, meaning they impose the same conditional independence constraints on the distribution, which is taken into account by SHD.

Due to the high computational cost of SCV and PI in structure learning, we designed the following experiments. Structures are learned using only NR and UCV, but the final parameters are also learned with SCV and PI for the obtained structures. These approaches will be referred to as NR-SCV, NR-PI, UCV-SCV and UCV-PI. For each method, we carry out a Hill-Climbing (HC) search combined with 5-fold cross-validation. The search process stops when the improvement between iterations is less than the predefined tolerance $\epsilon = 0.01$.

To mitigate the impact of the initialization, we used two initial structures as starting points: one with all LG nodes and the other with all CKDE nodes, both initially empty (no arcs). For each dataset, the best model is selected between the two starting points based on the absolute error in log-likelihood. Finally, the statistical analysis of log-likelihood performance follows the structure proposed in Section \ref{subsectionPL}.

\subsubsection{Five nodes SPBN}
\begin{figure}[h]
\centering 
\includegraphics[scale=0.258]{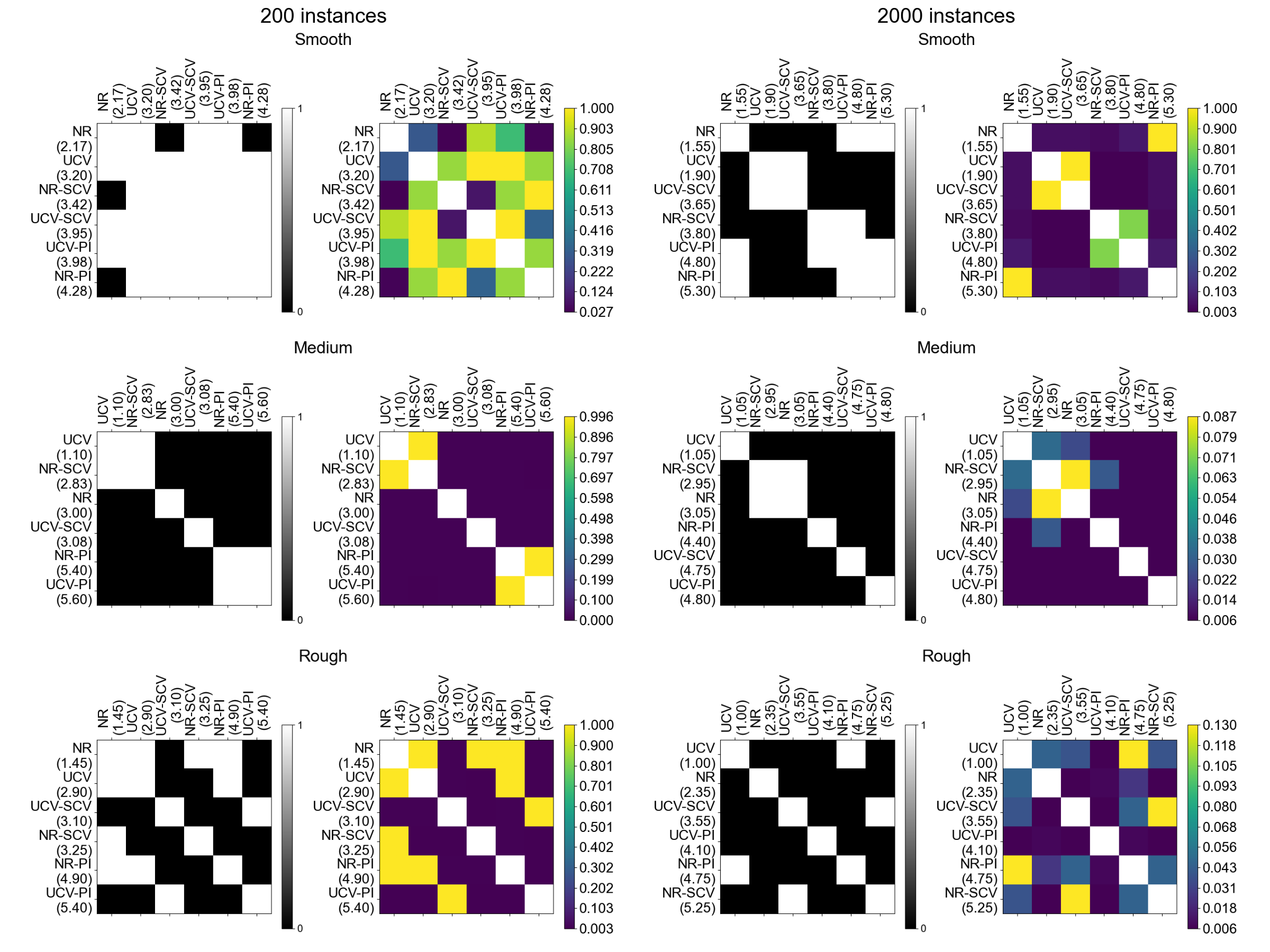}
\caption{Synthetic data experiments hypothesis acceptance (left) and p-value (right) matrices for the  five nodes SPBN in learning structures experiments} 
\label{fig:sa_nm_5}
\end{figure}

The performance results are shown in Table \ref{tab:results_5_nodes_network_mix}. Similar to the learning parameters framework, we see that NR only outperforms the proposed methods in the $N=200$ case with smooth and rough densities. Nonetheless, there are no statistical differences in terms of medians. UCV outperforms the rest of the methods for the rest of the scenarios. In all of them, there is a statistical difference between UCV and NR (see Figure \ref{fig:sa_nm_5} and Table \ref{tab:results_nm_pvalues}), which in some cases is not even the third best method in terms of medians or best result. Regardless of the bandwidth selector used for learning the structure (NR or UCV), SCV and PI exhibit moderate overall performance in terms of log-likelihood, as previously noted. However, for the medium density, they outperform NR while maintaining good structure results, even though the structure is optimized with this bandwidth selector.

Figure \ref{fig:sa_nm_5} and Table \ref{tab:results_nm_pvalues} summarize the statistical analysis of median comparisons. Overall, fewer significant differences are observed compared to the learning parameter experiments. This may stem from the increased number of comparisons and limited statistical power to detect true differences, a key consideration for this analysis.

\afterpage{\clearpage}
\begin{sidewaystable}[h]
\centering
\caption{Synthetic data experiments log-likelihood and SHD absolute error summary for the five nodes SPBN in learning structures experiments}
\label{tab:results_5_nodes_network_mix}
\scriptsize 
\setlength{\tabcolsep}{1.5pt} 
\begin{tabular}{@{}p{1.5cm}p{2cm}cc cc cc cc cc cc cc cc@{}}
\toprule
\textbf{Density} & \textbf{Bandwidth Method} & 
\multicolumn{4}{c}{$\mathbf{N = 200}$} & 
\multicolumn{4}{c}{$\mathbf{N = 2000}$} & 
\multicolumn{4}{c}{$\mathbf{N = 10000}$} & 
\multicolumn{4}{c}{$\mathbf{N = 20000}$} \\
\cmidrule(lr){3-6} \cmidrule(lr){7-10} \cmidrule(lr){11-14} \cmidrule(lr){15-18}
& & 
\multicolumn{2}{c}{\textbf{Log-likelihood}} & \multicolumn{2}{c}{\textbf{SHD}} & 
\multicolumn{2}{c}{\textbf{Log-likelihood}} & \multicolumn{2}{c}{\textbf{SHD}} & 
\multicolumn{2}{c}{\textbf{Log-likelihood}} & \multicolumn{2}{c}{\textbf{SHD}} & 
\multicolumn{2}{c}{\textbf{Log-likelihood}} & \multicolumn{2}{c}{\textbf{SHD}} \\
\cmidrule(lr){3-4} \cmidrule(lr){5-6} \cmidrule(lr){7-8} \cmidrule(lr){9-10} 
\cmidrule(lr){11-12} \cmidrule(lr){13-14} \cmidrule(lr){15-16} \cmidrule(lr){17-18}
& & 
\textbf{Median} & \textbf{Lowest} & \textbf{Median} & \textbf{Lowest} & 
\textbf{Median} & \textbf{Lowest} & \textbf{Median} & \textbf{Lowest} & 
\textbf{Median} & \textbf{Lowest} & \textbf{Median} & \textbf{Lowest} & 
\textbf{Median} & \textbf{Lowest} & \textbf{Median} & \textbf{Lowest} \\
\midrule
\multirow{6}{*}{Smooth} 
    & NR & $\phantom{0}\mathbf{718.03}$ & $\phantom{0}534.95$ & $4.5$ & $1$ & $\phantom{0}328.86$ & $268$ & $1\phantom{.0}$ & $1$ & $204.78$ & $175.46$ & $3\phantom{.0}$ & $0$ & $164.19$ & $143.23$ & $4\phantom{.0}$ & $1$ \\
    & UCV & $1138.32$ & $\mathbf{\phantom{0}424.75}$ & $4\phantom{.0}$ & $1$ & $\mathbf{\phantom{0}294.21\dagger}$ & $\mathbf{156.42}$ & $2.5$ & $1$ & $\mathbf{156.46\dagger}$ & $\mathbf{\phantom{0}78.03}$ & $2.5$ & $0$ & $\mathbf{\phantom{0}100.39\dagger}$ & $\mathbf{\phantom{0}52.21}$ & $1.5$ & $1$ \\
    & NR-SCV & $\phantom{0}971.58$ & $\phantom{0}534.95$ & $4\phantom{.0}$ & $1$ & $\phantom{0}686.22$ & $269.93$ & $6.5$ & $1$ & $--$ & $--$ & $--$ & $--$ & $--$ & $--$ & $--$ & $--$ \\
    & NR-PI & $1192.61$ & $\phantom{0}727.44$ & $4\phantom{.0}$ & $1$ & $1379.66$ & $954.94$ & $5.5$ & $1$ & $--$ & $--$ & $--$ & $--$ & $--$ & $--$ & $--$ & $--$ \\
    & UCV-SCV & $1127.51$ & $\phantom{0}615.46$ & $4\phantom{.0}$ & $1$ & $\phantom{0}612.85$ & $287.76$ & $6.5$ & $1$ & $--$ & $--$ & $--$ & $--$ & $--$ & $--$ & $--$ & $--$ \\
    & UCV-PI & $1192.61$ & $1138.39$ & $4\phantom{.0}$ & $4$ & $1479.30$ & $773.75$ & $4.5$ & $1$ & $--$ & $--$ & $--$ & $--$ & $--$ & $--$ & $--$ & $--$ \\
\midrule
\multirow{6}{*}{Medium} 
    & NR & $1022.7$ & $\phantom{0}950.68$ & $4\phantom{.0}$ & $1$ & $560.94$ & $523.15$ & $1$ & $1$ & $341.06$ & $314.13$ & $2$ & $2$ & $271.65$ & $246.6$ & $3$ & $2$ \\
    & UCV & $\mathbf{\phantom{00}337.01\dagger}$ & $\mathbf{\phantom{0}289.32}$ & $4.5$ & $4$ & $\mathbf{\phantom{0}170.92\dagger}$ & $\mathbf{132.02}$ & $4$ & $4$ & $\mathbf{\phantom{00}75.22\dagger}$ & $\mathbf{\phantom{0}57.08}$ & $5$ & $0$ & $\mathbf{\phantom{00}50.21\dagger}$ & $\mathbf{\phantom{0}19.77}$ & $5$ & $0$ \\
    & NR-SCV & $\phantom{0}542.44$ & $\phantom{0}362.36$ & $4\phantom{0.}$ & $1$ & $230.89$ & $185.70$ & $1$ & $1$ & $--$ & $--$ & $--$ & $--$ & $--$ & $--$ & $--$ & $--$ \\
    & NR-PI & $2633.19$ & $1990.62$ & $4\phantom{.0}$ & $1$ & $653.84$ & $582.72$ & $1$ & $1$ & $--$ & $--$ & $--$ & $--$ & $--$ & $--$ & $--$ & $--$ \\
    & UCV-SCV & $\phantom{0}532.57$ & $\phantom{0}393.99$ & $4\phantom{.0}$ & $1$ & $291.70$ & $244.93$ & $4$ & $4$ & $--$ & $--$ & $--$ & $--$ & $--$ & $--$ & $--$ & $--$ \\
    & UCV-PI & $2754.61$ & $1990.62$ & $4\phantom{.0}$ & $1$ & $764.89$ & $618.78$ & $4$ & $4$ & $--$ & $--$ & $--$ & $--$ & $--$ & $--$ & $--$ & $--$ \\
\midrule
\multirow{6}{*}{Rough} 
    & NR & $\mathbf{1569.44}$ & $1514.86$ & $5\phantom{.0}$ & $4$ & $1071.06$ & $1019.32$ & $4$ & $4$ & $745.67$ & $691.41$ & $4$ & $4$ & $608.84$ & $560.8$ & $4\phantom{.0}$ & $4$ \\
    & UCV & $1643.17$ & $\mathbf{1063.29}$ & $4\phantom{.0}$ & $4$ & $\mathbf{\phantom{00}331.39\dagger}$ & $\mathbf{\phantom{00}16.01}$ & $4$ & $4$ & $\mathbf{\phantom{0}100.96\dagger}$ & $\mathbf{\phantom{0}26.86}$ & $4$ & $4$ & $\mathbf{\phantom{0}324.49\dagger}$ & $\mathbf{\phantom{0}49.44}$ & $4.5$ & $4$ \\
    & NR-SCV & $1829.78$ & $1637.50$ & $4.5$ & $4$ & $2451.16$ & $\phantom{0}910.44$ & $4$ & $4$ & $--$ & $--$ & $--$ & $--$ & $--$ & $--$ & $--$ & $--$ \\
    & NR-PI & $2371.11$ & $1996.45$ & $5\phantom{.0}$ & $4$ & $2047.02$ & $1846.76$ & $4$ & $4$ & $--$ & $--$ & $--$ & $--$ & $--$ & $--$ & $--$ & $--$ \\
    & UCV-SCV & $1841.20$ & $1505.64$ & $4\phantom{.0}$ & $4$ & $1325.95$ & $\phantom{0}570.67$ & $4$ & $4$ & $--$ & $--$ & $--$ & $--$ & $--$ & $--$ & $--$ & $--$ \\
    & UCV-PI & $2470.74$ & $2250.44$ & $4\phantom{.0}$ & $4$ & $1979.20$ & $1773.84$ & $4$ & $4$ & $--$ & $--$ & $--$ & $--$ & $--$ & $--$ & $--$ & $--$ \\
\bottomrule
\end{tabular}

\end{sidewaystable}

For the rough density scenario with a sample size of 2000, UCV shows statistical differences with all methods except the second-worst, despite a moderate p-value of 0.130, which appears inconsistent. Notably, the p-values decrease as the sample size increases, highlighting growing differences between the proposed methods, particularly between UCV and NR. This trend reflects the stagnation of NR’s performance versus the improvement of the proposed methods.

The SHD results are summarized in Table \ref{tab:results_5_nodes_network_mix}. A key insight is that the best-performing method in terms of log-likelihood does not necessarily recover the best structures. For instance, in the 2000-sample case with the medium density, UCV achieves the best log-likelihood performance, yet all methods using NR nearly recover the original structure in median despite a poorer log-likelihood approximation. This is natural as it is one of the drawbacks of score-based methods for learning BNs.

Both NR and UCV produce comparable SHD errors overall, with NR often showing better median performance due to its stability, in contrast to the variability of UCV. However, UCV tends to achieve the best results, particularly in the 10000 and 20000-sample scenarios, where it occasionally recovers the original structure.

In summary, the results align with the patterns observed in parameter learning. Complexity negatively impacts the proposed methods, leading to worse structure recovery, as seen with parameter learning. NR remains a stable method, showing stagnation in both log-likelihood and structural errors with increased sample size. While UCV's variability affects median structure recovery, it achieves superior results in several scenarios for both log-likelihood and structural error.

Learning the structure solely with UCV or NR often diminishes the moderate performance of SCV and PI for the final parameters learning (UCV-SCV, UCV-PI, NR-UCV, NR-PI), typically resulting in worse outcomes compared to using UCV or NR alone. However, certain cases demonstrate improved log-likelihood performance while maintaining structural errors, suggesting their potential utility. Overall, the proposed methods outperform NR, but caution is advised in real-world applications, as UCV's variability and the occasional lack of alignment between log-likelihood performance and structure recovery can pose challenges.

\begin{table}[h!]
\centering
\caption{Synthetic data experiments p-values for the permutation test between NR and UCV across different network sizes in learning structures experiments}
\label{tab:results_nm_pvalues}
\footnotesize 
\setlength{\tabcolsep}{4pt} 
\begin{tabular}{@{}c p{1.5cm} cc@{}}
\toprule
\textbf{Nodes} & \textbf{Density} & \multicolumn{2}{c}{$\mathbf{N}$} \\
\cmidrule(l){3-4}
                     &                 & \textbf{10000} & \textbf{20000} \\
\midrule
\multirow{3}{*}{5}   & Smooth           & $0.0182$ & $0.0126$ \\
                     & Medium         & $0.0010$  & $0.0004$ \\
                     & Rough         & $0.0004$ & $0.0012$ \\
\midrule
\multirow{3}{*}{10}  & Smooth           & $0.1208$ & $0.7544$ \\
                     & Medium         & $0.0006$ & $0.0002$ \\
                     & Rough         & $0.0278$ & $0.0156$ \\
\midrule
\multirow{3}{*}{15}  & Smooth           & $0.0096$ & $0.0058$ \\
                     & Medium         & $0.0134$ & $0.1606$ \\
                     & Rough         & $0.0074$ & $0.0956$ \\
\bottomrule
\end{tabular}

\end{table}

\subsubsection{Ten nodes SPBN}
Table \ref{tab:results_10_nodes_network_mix} presents the log-likelihood absolute error results. As expected, increased complexity due to a larger number of variables impacts the proposed methods. However, they outperform NR in most scenarios, with significant median differences in nearly all cases.

NR performs best in all scenarios with the smooth density, albeit without statistical significance, reflecting its stagnation. In contrast, UCV systematically improves with increasing sample size across all scenarios. For the rough density and small sample sizes, the usage of SCV and PI for the final parameter learning outperforms both NR and UCV, despite the structures being optimized for log-likelihood under the latter methods. Specifically, for $N = 200$, UCV-PI and UCV-SCV outperform UCV, while NR-PI outperforms NR. For $N = 2000$, UCV-PI and UCV-SCV outperform UCV, while NR-PI and NR-SCV outperform NR.

\begin{figure}[h]
\centering 

\includegraphics[scale=0.27]{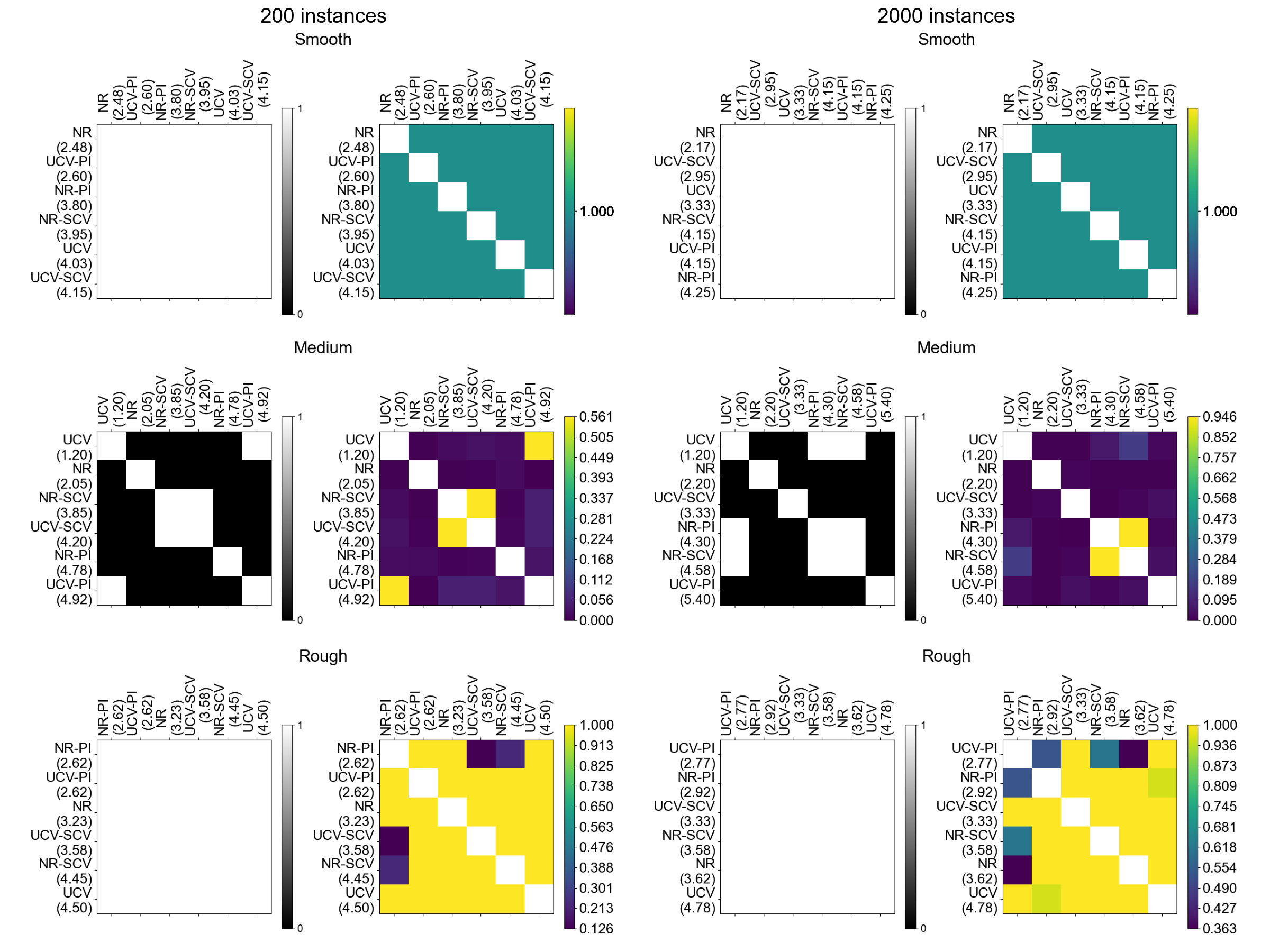}
\caption{Synthetic data experiments hypothesis acceptance (left) and p-value (right) matrices for the ten nodes SPBN in learning structures experiments} 
\label{fig:sa_nm_10}
\end{figure}

Figure \ref{fig:sa_nm_10} and Table \ref{tab:results_nm_pvalues} summarize the statistical analysis, showing minimal differences in the 200 and 2000 sample size scenarios due to increased complexity. For the medium density, significant differences are observed for UCV compared to almost all methods, except the second and third worst, a rare occurrence associated with the statistical power of the adjustment method. The p-values in Table \ref{tab:results_nm_pvalues} highlight the stagnation of NR and the improvement of UCV, with lower p-values indicating UCV's advantage.

\afterpage{\clearpage}
\begin{sidewaystable}[h]
\centering
\caption{Synthetic data experiments log-likelihood and SHD absolute error summary for the ten nodes SPBN in learning structures experiments}
\label{tab:results_10_nodes_network_mix}
\scriptsize 
\setlength{\tabcolsep}{1pt} 
\begin{tabular}{@{}p{1.5cm}p{2cm}cc cc cc cc cc cc cc cc@{}}
\toprule
\textbf{Density} & \textbf{Bandwidth Method} & 
\multicolumn{4}{c}{$\mathbf{N = 200}$} & 
\multicolumn{4}{c}{$\mathbf{N = 2000}$} & 
\multicolumn{4}{c}{$\mathbf{N = 10000}$} & 
\multicolumn{4}{c}{$\mathbf{N = 20000}$} \\
\cmidrule(lr){3-6} \cmidrule(lr){7-10} \cmidrule(lr){11-14} \cmidrule(lr){15-18}
& & 
\multicolumn{2}{c}{\textbf{Log-likelihood}} & \multicolumn{2}{c}{\textbf{SHD}} & 
\multicolumn{2}{c}{\textbf{Log-likelihood}} & \multicolumn{2}{c}{\textbf{SHD}} & 
\multicolumn{2}{c}{\textbf{Log-likelihood}} & \multicolumn{2}{c}{\textbf{SHD}} & 
\multicolumn{2}{c}{\textbf{Log-likelihood}} & \multicolumn{2}{c}{\textbf{SHD}} \\
\cmidrule(lr){3-4} \cmidrule(lr){5-6} \cmidrule(lr){7-8} \cmidrule(lr){9-10} 
\cmidrule(lr){11-12} \cmidrule(lr){13-14} \cmidrule(lr){15-16} \cmidrule(lr){17-18}
& & 
\textbf{Median} & \textbf{Lowest} & \textbf{Median} & \textbf{Lowest} & 
\textbf{Median} & \textbf{Lowest} & \textbf{Median} & \textbf{Lowest} & 
\textbf{Median} & \textbf{Lowest} & \textbf{Median} & \textbf{Lowest} & 
\textbf{Median} & \textbf{Lowest} & \textbf{Median} & \textbf{Lowest} \\
\midrule
\multirow{6}{*}{Smooth} 
    & NR & $\mathbf{12765.96}$ & $\mathbf{\phantom{0}8903.98}$ & $18\phantom{.0}$ & $14$ & $\mathbf{\phantom{0}8755.52}$ & $\mathbf{\phantom{0}6077.11}$ & $15.5$ & $12$ & $\mathbf{\phantom{0}9583.51}$ & $\mathbf{5438.71}$ & $16$ & $12$ & $\mathbf{9410.69}$ & $\mathbf{5261.49}$ & $16$ & $14$ \\
    & UCV & $14949.41$ & $11105.18$ & $18.5$ & $16$ & $12683.72$ & $\phantom{0}7050.23$ & $17\phantom{.0}$ & $13$ & $12724.84$ & $5993.12$ & $17$ & $10$ & $9764.21$ & $6162.01$ & $14$ & $12$ \\
    & NR-SCV & $12998.94$ & $\phantom{0}9341.81$ & $18\phantom{.0}$ & $16$ & $12146.24$ & $\phantom{0}8405.53$ & $16\phantom{.0}$ & $12$ & $--$ & $--$ & $--$ & $--$ & $--$ & $--$ & $--$ & $--$ \\
    & NR-PI & $13451.41$ & $11789.36$ & $18\phantom{.0}$ & $16$ & $11973.08$ & $\phantom{0}8961.19$ & $16.5$ & $14$ & $--$ & $--$ & $--$ & $--$ & $--$ & $--$ & $--$ & $--$ \\
    & UCV-SCV & $13130.46$ & $10637.82$ & $19\phantom{.0}$ & $16$ & $10482.09$ & $\phantom{0}8499.17$ & $16.5$ & $13$ & $--$ & $--$ & $--$ & $--$ & $--$ & $--$ & $--$ & $--$ \\
    & UCV-PI & $13130.46$ & $10637.82$ & $17.5$ & $16$ & $11713.41$ & $10120.64$ & $17\phantom{.0}$ & $13$ & $--$ & $--$ & $--$ & $--$ & $--$ & $--$ & $--$ & $--$ \\
\midrule
\multirow{6}{*}{Medium} 
    & NR & $1778.53$ & $1640.95$ & $17\phantom{.0}$ & $14$ & $\phantom{0}969.81$ & $\phantom{0}896.14$ & $19.5$ & $16$ & $602.58$ & $572.68$ & $23$ & $16$ & $481.86$ & $445.67$ & $25\phantom{.0}$ & $24$ \\
    & UCV & $\mathbf{\phantom{0}1139.87\dagger}$ & $\mathbf{\phantom{0}748.6}$ & $18\phantom{.0}$ & $14$ & $\mathbf{\phantom{00}521.08\dagger}$ & $\mathbf{\phantom{0}389.27}$ & $19.5$ & $15$ & $\mathbf{\phantom{0}267.62\dagger}$ & $\mathbf{242.67}$ & $20$ & $15$ & $\mathbf{\phantom{0}250.64\dagger}$ & $\mathbf{218.26}$ & $18.5$ & $17$ \\
    & NR-SCV & $1943.01$ & $1756.96$ & $18\phantom{.0}$ & $14$ & $3187.44$ & $\phantom{0}761.56$ & $17\phantom{.0}$ & $16$ & $--$ & $--$ & $--$ & $--$ & $--$ & $--$ & $--$ & $--$ \\
    & NR-PI & $2679.34$ & $2459.38$ & $18.5$ & $14$ & $2596.48$ & $\phantom{0}986.64$ & $17\phantom{.0}$ & $16$ & $--$ & $--$ & $--$ & $--$ & $--$ & $--$ & $--$ & $--$ \\
    & UCV-SCV & $2111.31$ & $1636.78$ & $18.5$ & $14$ & $1190.12$ & $\phantom{0}761.56$ & $18.5$ & $15$ & $--$ & $--$ & $--$ & $--$ & $--$ & $--$ & $--$ & $--$ \\
    & UCV-PI & $2852.47$ & $2475.32$ & $17\phantom{.0}$ & $14$ & $3320.94$ & $2301.99$ & $19.5$ & $15$ & $--$ & $--$ & $--$ & $--$ & $--$ & $--$ & $--$ & $--$ \\
\midrule
\multirow{6}{*}{Rough} 
    & NR & $3508.91$ & $917.44$ & $17.5$ & $15$ & $2683.28$ & $514.01$ & $18$ & $14$ & $5142.87$ & $188.78$ & $15$ & $12$ & $5595.95$ & $1568.92$ & $16$ & $11$ \\
    & UCV & $7012.13$ & $435.99$ & $19\phantom{.0}$ & $17$ & $3261.76$ & $578.98$ & $19$ & $17$ & $\mathbf{\phantom{0}1366.13\dagger}$ & $\mathbf{153.37}$ & $19$ & $17$ & $\mathbf{\phantom{0}1445.96\dagger}$ & $\mathbf{\phantom{0}132.12}$ & $18$ & $6$ \\
    & NR-SCV & $9903.42$ & $470.01$ & $17.5$ & $15$ & $1464.75$ & $300.96$ & $18$ & $14$ & $--$ & $--$ & $--$ & $--$ & $--$ & $--$ & $--$ & $--$ \\
    & NR-PI & $2524.9$ & $\mathbf{152.97}$ & $17.5$ & $16$ & $\mathbf{\phantom{0}917.16}$ & $235.51$ & $18$ & $14$ & $--$ & $--$ & $--$ & $--$ & $--$ & $--$ & $--$ & $--$ \\
    & UCV-SCV & $5757.81$ & $201.93$ & $17.5$ & $15$ & $\phantom{0}965.61$ & $\mathbf{218.04}$ & $19$ & $17$ & $--$ & $--$ & $--$ & $--$ & $--$ & $--$ & $--$ & $--$ \\
    & UCV-PI & $\mathbf{2057.61}$ & $435.99$ & $18\phantom{.0}$ & $15$ & $1290.17$ & $430.06$ & $19$ & $16$ & $--$ & $--$ & $--$ & $--$ & $--$ & $--$ & $--$ & $--$ \\
\bottomrule
\end{tabular}
\end{sidewaystable}

Regarding SHD performance (Table \ref{tab:results_10_nodes_network_mix}), similar patterns are observed as in the five-node case. The best log-likelihood algorithm does not always recover the best approximate structure with the lowest SHD, as seen in the 10000 and 20000 sample size cases for the smooth density. This is also reflected in some instances where SCV and PI are used for final parameter learning. See for example, the medium density with $N=2000$, NR-SCV and NR-PI outperform NR in median structure recovery despite their poorer log-likelihood results. However, in general, the best structural results are linked to the method with the best log-likelihood performance, particularly when using UCV. With UCV, slightly worse median SHD results are observed in some cases, reflecting the variability of the bandwidth selector.  

Overall, we observe the same patterns as in the five-node case, with worse results in both finding statistical differences and performance across both metrics. Notably, there are cases where the method that achieves the best log-likelihood does not necessarily retrieve the ground truth structure most effectively.

However, the proposed methods generally outperform NR in both log-likelihood and structural error. In particular, UCV excels in high-sample-size scenarios, encouraging its use in such contexts. Conversely, methods with SCV and PI for the final parameter learning (UCV-SCV, UCV-PI, NR-SCV, NR-PI) show superior performance in only two scenarios. Given their high computational cost, they are only worth considering for small-sample-size scenarios with sparse networks in terms of connections, where their unique properties provide an advantage. Finally, while NR is outperformed in several scenarios, it remains competitive, especially in structure learning. It demonstrates slightly better median performance, which can be attributed to its stable nature.

\subsubsection{Fifteen nodes SPBN}

Table \ref{tab:results_15_nodes_network_mix} presents patterns similar to those observed in previous sections, but with a more pronounced effect of complexity on the proposed methods, leading to poorer performance. In terms of log-likelihood error, NR consistently outperforms the proposed methods in almost every scenario with respect to the median error. Additionally, significant differences with the second-best method are found in some scenarios.

As observed in the parameter learning context, the stability and oversmoothing characteristics of this bandwidth method give NR an advantage in handling complexity, an effect that is further accentuated when combined with the challenges of learning structure. Conversely, methods using SCV and PI for the final parameter learning exhibit poor performance, achieving the best log-likelihood error only in the medium density with $N=200$ scenario for UCV-PI. UCV also outperforms NR only in this scenario.

Nonetheless, the nuanced pattern of stagnation versus improvement for NR and UCV, respectively, becomes evident as the sample size increases. This trend is also reflected in the p-values shown in Table \ref{tab:results_nm_pvalues}. As in Section \ref{subsectionPL}, an additional experiment with the rough density and a sample size of $40000$ demonstrated significant differences in median values, with $23032.93$ for NR and $20371.67$ for UCV, following the established trend.

\begin{figure}[h]
\centering 
\includegraphics[scale=0.27]{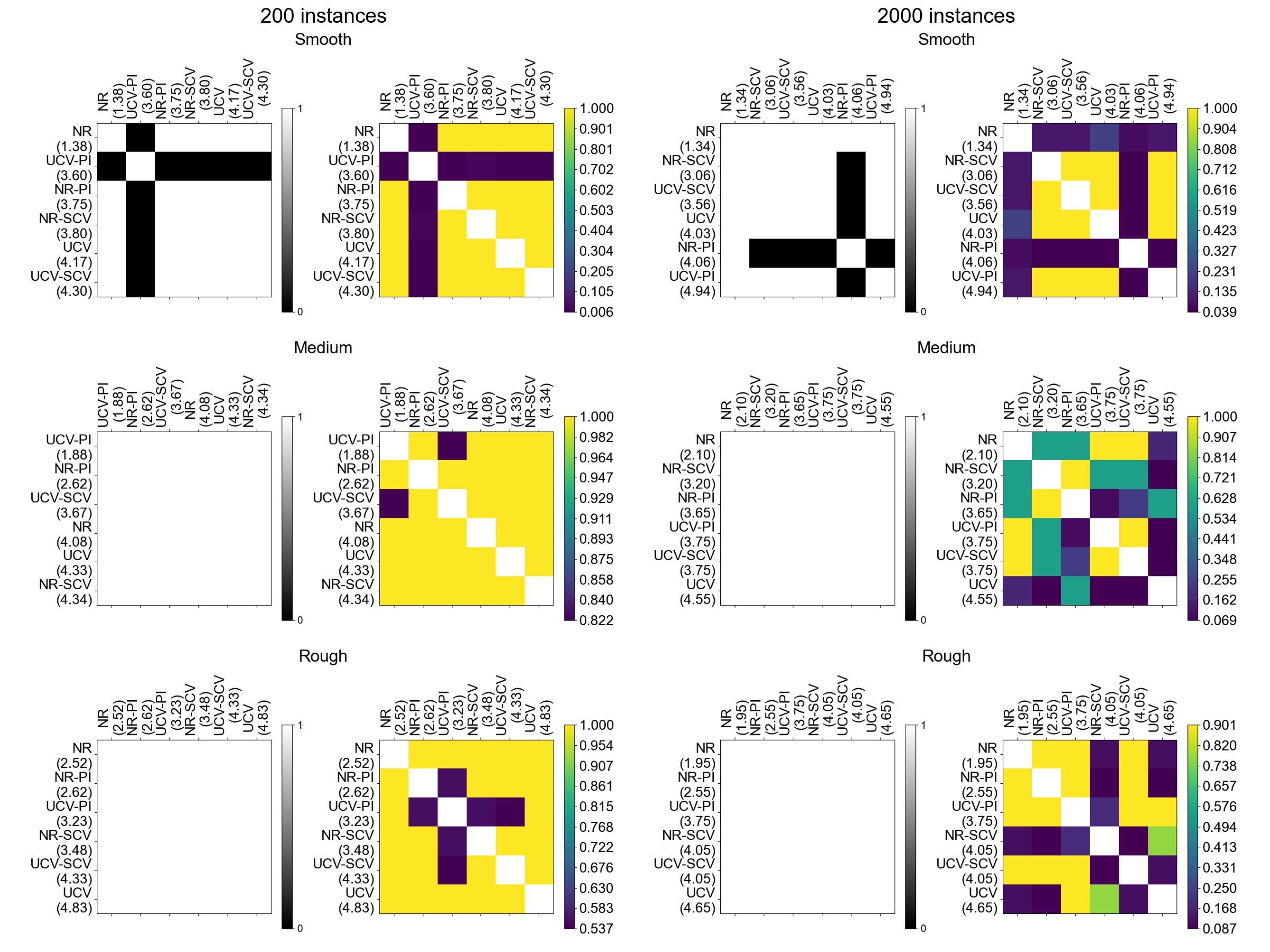}
\caption{Synthetic data experiments hypothesis acceptance (left) and p-value (right) matrices for the fifteen nodes SPBN in learning structures experiments} 
\label{fig:sa_nm_15}
\end{figure}

Figure \ref{fig:sa_nm_15} shows the detailed summary of the statistical comparison of medians. The main insight is that there is no significant difference between almost every method in almost every case. With respect to SHD in Table \ref{tab:results_15_nodes_network_mix}, we see that in general, the best log-likelihood results are associated with the best results in Hamming structural distance. Nonetheless, there are still cases where this does not happen, such as the high sample size scenarios with the smooth density, where UCV retrieves better structures. We also find slightly better results in the 2000 sample size scenario with the rough density with NR-SCV, even though it presented the worst log-likelihood errors.

\afterpage{\clearpage}
\begin{sidewaystable}[h]
\centering
\caption{Synthetic data experiments log-likelihood and SHD absolute error summary for the fifteen nodes SPBN in learning structures experiments}
\label{tab:results_15_nodes_network_mix}
\scriptsize 
\setlength{\tabcolsep}{0.75pt} 
\begin{tabular}{@{}p{1.5cm}p{2cm}cc cc cc cc cc cc cc cc@{}}
\toprule
\textbf{Density} & \textbf{Bandwidth Method} & 
\multicolumn{4}{c}{$\mathbf{N = 200}$} & 
\multicolumn{4}{c}{$\mathbf{N = 2000}$} & 
\multicolumn{4}{c}{$\mathbf{N = 10000}$} & 
\multicolumn{4}{c}{$\mathbf{N = 20000}$} \\
\cmidrule(lr){3-6} \cmidrule(lr){7-10} \cmidrule(lr){11-14} \cmidrule(lr){15-18}
& & 
\multicolumn{2}{c}{\textbf{Log-likelihood}} & \multicolumn{2}{c}{\textbf{SHD}} & 
\multicolumn{2}{c}{\textbf{Log-likelihood}} & \multicolumn{2}{c}{\textbf{SHD}} & 
\multicolumn{2}{c}{\textbf{Log-likelihood}} & \multicolumn{2}{c}{\textbf{SHD}} & 
\multicolumn{2}{c}{\textbf{Log-likelihood}} & \multicolumn{2}{c}{\textbf{SHD}} \\
\cmidrule(lr){3-4} \cmidrule(lr){5-6} \cmidrule(lr){7-8} \cmidrule(lr){9-10} 
\cmidrule(lr){11-12} \cmidrule(lr){13-14} \cmidrule(lr){15-16} \cmidrule(lr){17-18}
& & 
\textbf{Median} & \textbf{Lowest} & \textbf{Median} & \textbf{Lowest} & 
\textbf{Median} & \textbf{Lowest} & \textbf{Median} & \textbf{Lowest} & 
\textbf{Median} & \textbf{Lowest} & \textbf{Median} & \textbf{Lowest} & 
\textbf{Median} & \textbf{Lowest} & \textbf{Median} & \textbf{Lowest} \\
\midrule
\multirow{6}{*}{Smooth} 
    & NR & $\mathbf{\phantom{0}29765.24\ddagger}$ & $\mathbf{28011.28}$ & $22\phantom{.0}$ & $14$ & $\mathbf{25534.01}$ & $\mathbf{25073.91}$ & $19$ & $11$ & $\mathbf{\phantom{0}23764.95\ddagger}$ & $23676.34$ & $20$ & $17$ & $\mathbf{\phantom{0}23241.43\ddagger}$ & $23124.83$ & $22$ & $17$ \\
    & UCV & $35501.35$ & $29586.98$ & $26\phantom{.0}$ & $18$ & $34855.72$ & $28720.08$ & $22$ & $17$ & $33354.61$ & $\mathbf{20977.28}$ & $22$ & $14$ & $32313.97$ & $\mathbf{19353.62}$ & $22$ & $15$ \\
    & NR-SCV & $35951.88$ & $30678.38$ & $27\phantom{.0}$ & $16$ & $28219.22$ & $25375.38$ & $21$ & $16$ & $--$ & $--$ & $--$ & $--$ & $--$ & $--$ & $--$ & $--$ \\
    & NR-PI & $34734.89$ & $31686.67$ & $22.5$ & $14$ & $32683.28$ & $27481.73$ & $21$ & $18$ & $--$ & $--$ & $--$ & $--$ & $--$ & $--$ & $--$ & $--$ \\
    & UCV-SCV & $35974.13$ & $30695.59$ & $25\phantom{.0}$ & $18$ & $34600.82$ & $34019.36$ & $22$ & $19$ & $--$ & $--$ & $--$ & $--$ & $--$ & $--$ & $--$ & $--$ \\
    & UCV-PI & $34853.72$ & $33133.46$ & $26\phantom{.0}$ & $18$ & $37856.47$ & $34349.37$ & $22$ & $19$ & $--$ & $--$ & $--$ & $--$ & $--$ & $--$ & $--$ & $--$ \\
\midrule
\multirow{6}{*}{Medium} 
    & NR & $\phantom{0}87034.07$ & $38830.35$ & $24$ & $22$ & $\mathbf{24292.75}$ & $\mathbf{21408.23}$ & $20.5$ & $16$ & $\mathbf{\phantom{0}24033.56\ddagger}$ & $\mathbf{19174.96}$ & $20.5$ & $16$ & $\mathbf{22434.55}$ & $\mathbf{18939.98}$ & $20$ & $15$ \\
    & UCV & $\phantom{0}77961.46$ & $35502.82$ & $24$ & $22$ & $34751.05$ & $24390.22$ & $22.5$ & $22$ & $31494.48$ & $27064.01$ & $23\phantom{.0}$ & $18$ & $24801.67$ & $22113.22$ & $24$ & $20$ \\
    & NR-SCV & $187492.08$ & $\mathbf{33042.24}$ & $24$ & $22$ & $30514.55$ & $25819.01$ & $22\phantom{.0}$ & $16$ & $--$ & $--$ & $--$ & $--$ & $--$ & $--$ & $--$ & $--$ \\
    & NR-PI & $103483.15$ & $36550.98$ & $24$ & $21$ & $28123.97$ & $24432.19$ & $22\phantom{.0}$ & $16$ & $--$ & $--$ & $--$ & $--$ & $--$ & $--$ & $--$ & $--$ \\
    & UCV-SCV & $\phantom{0}86175.02$ & $35502.82$ & $24$ & $22$ & $34751.05$ & $29627.89$ & $22.5$ & $22$ & $--$ & $--$ & $--$ & $--$ & $--$ & $--$ & $--$ & $--$ \\
    & UCV-PI & $\mathbf{\phantom{0}64997.33}$ & $35502.82$ & $24$ & $22$ & $34031.09$ & $27465.52$ & $23.5$ & $22$ & $--$ & $--$ & $--$ & $--$ & $--$ & $--$ & $--$ & $--$ \\
\midrule
\multirow{6}{*}{Rough} 
    & NR & $\mathbf{33258.16}$ & $\mathbf{29274.67}$ & $22.5$ & $16$ & $\mathbf{27409.55}$ & $25566.53$ & $23.5$ & $10$ & $\mathbf{\phantom{0}25087.53\ddagger}$ & $23882.15$ & $19$ & $12$ & $\mathbf{23814.59}$ & $23353.02$ & $14\phantom{.0}$ & $11$ \\
    & UCV & $38531.22$ & $32303.41$ & $22.5$ & $20$ & $37907.78$ & $28809.53$ & $22\phantom{.0}$ & $20$ & $38230.86$ & $\mathbf{19166.11}$ & $24$ & $20$ & $26644.58$ & $\mathbf{18954.92}$ & $26.5$ & $22$ \\
    & NR-SCV & $35305.68$ & $30145.51$ & $22.5$ & $16$ & $38845.62$ & $\mathbf{\phantom{00}755.42}$ & $22.5$ & $10$ & $--$ & $--$ & $--$ & $--$ & $--$ & $--$ & $--$ & $--$ \\
    & NR-PI & $35961.68$ & $32906.28$ & $23\phantom{.0}$ & $16$ & $30870.22$ & $28134.17$ & $24\phantom{.0}$ & $10$ & $--$ & $--$ & $--$ & $--$ & $--$ & $--$ & $--$ & $--$ \\
    & UCV-SCV & $37975.95$ & $33008.49$ & $22.5$ & $20$ & $36728.44$ & $31608.18$ & $23\phantom{.0}$ & $22$ & $--$ & $--$ & $--$ & $--$ & $--$ & $--$ & $--$ & $--$ \\
    & UCV-PI & $36937.88$ & $34624.46$ & $22.5$ & $20$ & $33743.43$ & $29821.1$ & $24.5$ & $22$ & $--$ & $--$ & $--$ & $--$ & $--$ & $--$ & $--$ & $--$ \\
\bottomrule
\end{tabular}
\end{sidewaystable}

Overall, the analysis highlights the impact of complexity on the proposed methods compared to the stability of NR in both structural and log-likelihood errors. However, UCV demonstrates better structural errors in some high-sample-size scenarios, accompanied by improvements in log-likelihood performance, suggesting its potential in such settings. Methods involving SCV and PI for the final parameter learning exhibit trends consistent with previous analyses, showing relevance primarily in small sample size scenarios, where they occasionally retrieve better structures or log-likelihoods. See, for example, NR-SCV with the rough density and $N = 2000$ outperforming NR in median structure recovery, or UCV-PI with the medium density and $N = 200$ achieving the best median log-likelihood error.

One of the key insights from this analysis is the applicability of UCV in high sample size scenarios, showcasing the usefulness of the proposed methods. We also have to take into account the significance of NR in complex scenarios, where its stability due to the oversmoothing proves advantageous for learning structures in multiple scenarios.

\subsection{Real data experiments}
In this section, we conduct two different experiments using real-world datasets to validate the findings from the synthetic data experiments in Section \ref{sec:synthetic_experiments}. First, we learn the structure and parameters of various datasets from the UCI repository, exploring different configurations of data sample sizes and the number of variables. Second, we use the MetroPT-3 dataset (\cite{metropt-3}) to examine the behavior of NR and UCV as the sample size increases. 

\subsubsection{First experiment}
\label{sec:first_experiment}
The datasets used in the experiments are listed in Table \ref{uci_tables}. These datasets have been selected to align with the synthetic experiments, representing the various scenarios studied there with respect to the number of variables and sample size.

For each dataset, we perform a 10-fold split into training and validation sets to evaluate the behavior of different methods, following the same procedure as in the synthetic experiments. The log-likelihood is computed for each validation set. Unlike the synthetic scenario, where the true density is known, we do not have access to the true density in this case. Consequently, the absolute error in log-likelihood cannot be used as a comparison metric. Instead, we adopt the maximum log-likelihood as the primary criterion for evaluation and comparison.

This criterion does not explicitly penalize oversmoothing (underfitting), which can result in higher log-likelihood values yet suboptimal density approximations. However, in the case of CKDEs approximating the underlying density, oversmoothing rarely leads to log-likelihood values that surpass those of the true density, reinforcing the validity of this criterion for selecting and comparing SPBNs. Furthermore, it aligns with the network learning process, where log-likelihood is maximized while incorporating mechanisms to mitigate overfitting. Additionally, BNs offer interpretability, making them particularly valuable in real-world applications where expert knowledge can be leveraged for posterior corrections and refinement of the learned models.

Building on the learning network experiments in Section \ref{sec:synthetic_network}, we use the UCV and NR bandwidth selectors for both structure and parameter learning. For low-sample scenarios, we incorporate SCV and PI for parameter learning, resulting in four configurations: NR-PI, NR-SCV, UCV-PI, and UCV-SCV. To mitigate the influence of the starting point, we apply the same scheme to two initial empty structures, one consisting entirely of LG nodes and the other of CKDE nodes.

\begin{figure}[h]
\centering 
\includegraphics[scale = 0.37]{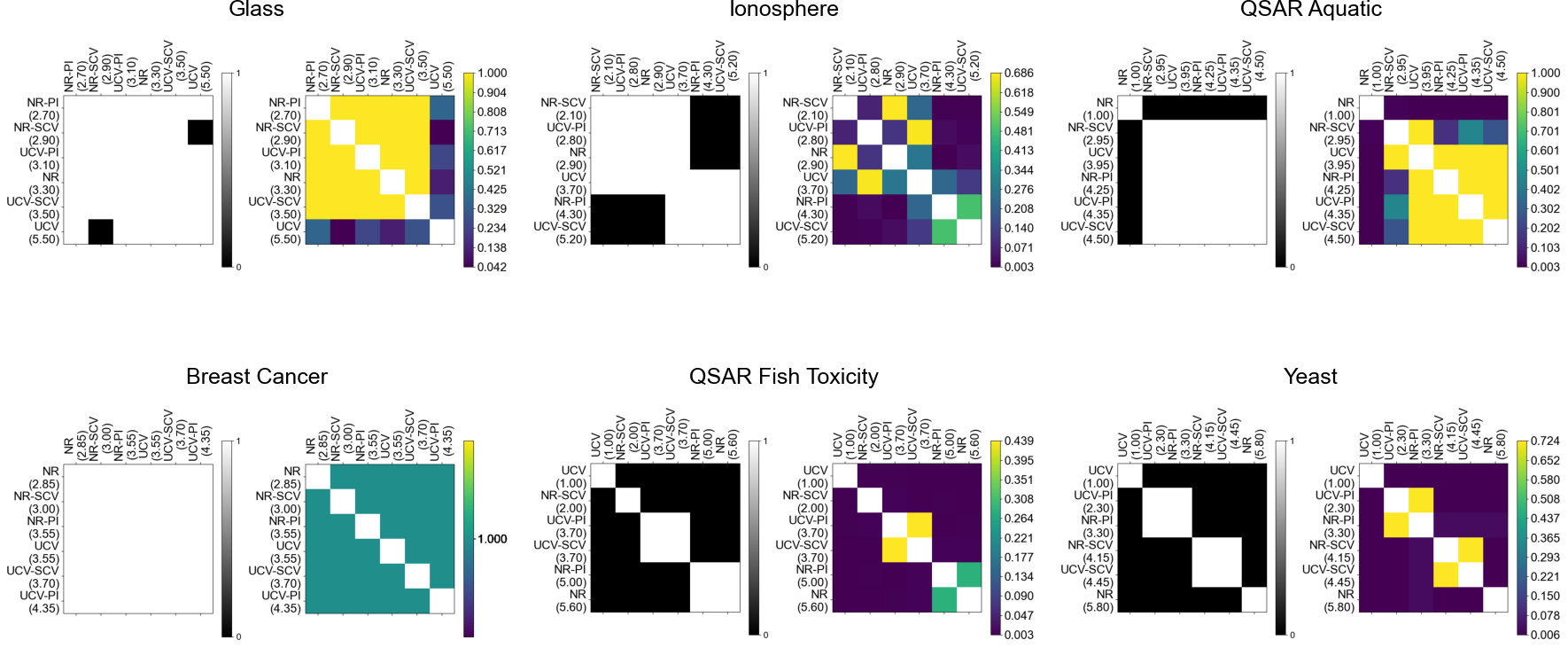}
\caption{Hypothesis acceptance (left) and p-value (right) matrices for the first experiment on low-sample size real data scenarios} 
\label{fig:sa_real_data_1}
\end{figure}

\begin{figure}[h]
\centering 
\includegraphics[scale = 0.35]{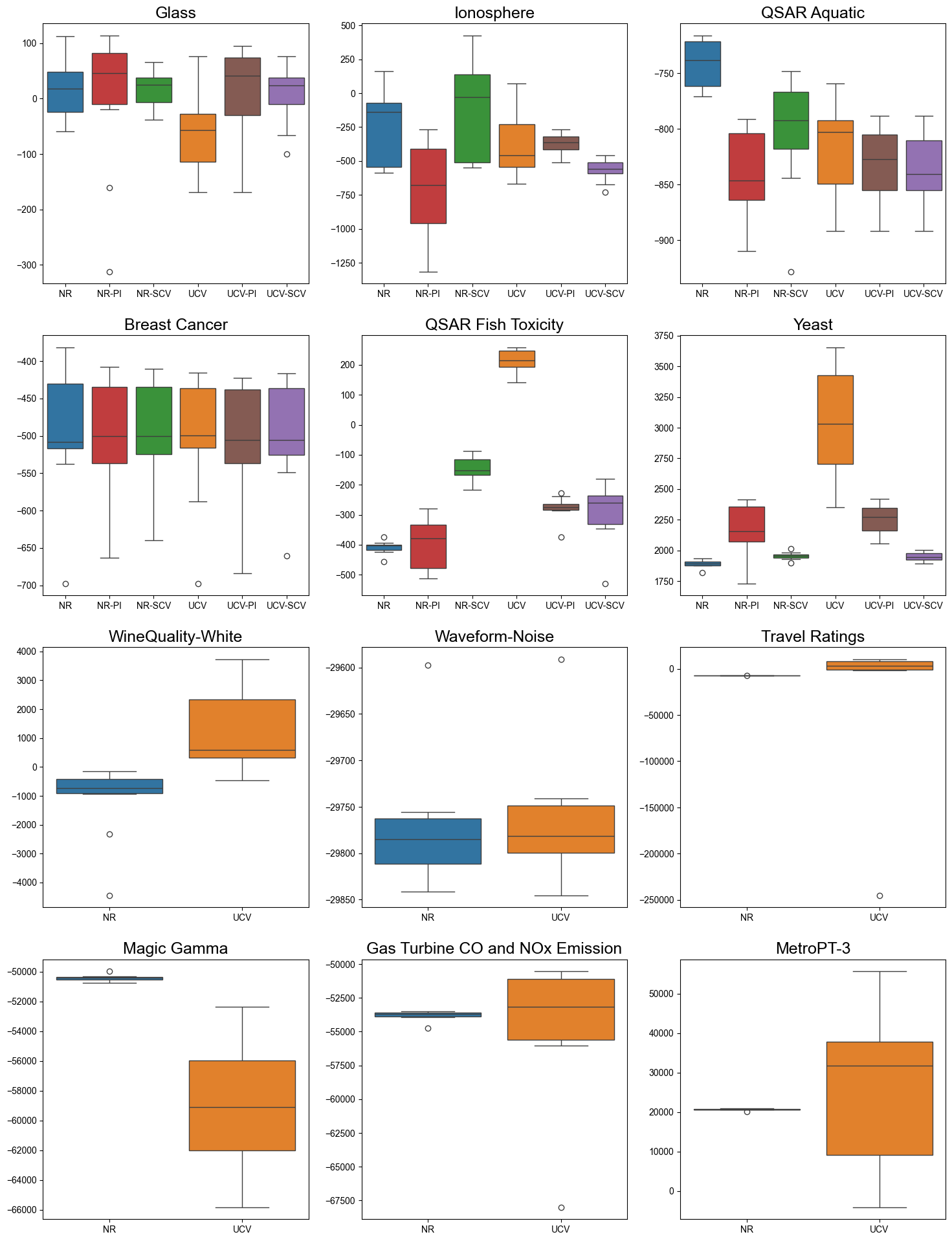}
\caption{Boxplots of the log-likelihood results for the first experiment on real data
} 
\label{fig:boxplots_real_data_1}
\end{figure}

The results are displayed in Table \ref{tab:real_data_pvalues},  Figure \ref{fig:sa_real_data_1} and Figure \ref{fig:boxplots_real_data_1}. Table \ref{tab:real_data_pvalues} reports the p-values in the case of a single comparison test of medians between NR and UCV. Figure \ref{fig:boxplots_real_data_1} shows the boxplot of the results for each dataset. Similarly to the synthetic experiments in Section \ref{subsectionPL}, Figure \ref{fig:sa_real_data_1} presents a summary of the statistical multiple comparisons. The statistical analysis follows the same premises as in Section \ref{sec:synthetic_experiments}. 

\begin{table}[h]
\centering
\caption{Datasets from the UCI repository with low-sample size scenarios (left) and high-sample size scenarios (right)}
\footnotesize
\label{uci_tables}
\renewcommand{\arraystretch}{1.2}
\setlength{\tabcolsep}{2pt}

\begin{tabular}{cc}
    \begin{subtable}[t]{0.48\textwidth}
        \centering
        \begin{tabular}{p{2.8cm} c c}
            \toprule
            \textbf{Dataset} & $\mathbf{N}$ & \textbf{\# Variables} \\
            \midrule
            Glass             & $214$   & $9$  \\
            Ionosphere        & $351$   & $31$ \\
            QSAR Aquatic      & $546$ & $9$ \\
            Breast Cancer      & $683$   & $9$  \\
            QSAR Fish Toxicity & $908$ & $7$ \\
            Yeast             & $1,484$  & $8$  \\
            \bottomrule
        \end{tabular}
    \end{subtable}
    &
    \begin{subtable}[t]{0.48\textwidth}
        \centering
        \begin{tabular}{p{5cm} c c}
            \toprule
            \textbf{Dataset} & $\mathbf{N}$ & \textbf{\# Variables} \\
            \midrule
            WineQuality-White  & $4,898$  & $12$ \\
            Waveform-Noise     & $5,000$ & $40$ \\
            Travel Ratings     & $5,546$ & $24$ \\
            Magic Gamma & $19,020$ & $10$ \\
            Gas Turbine CO and NOx Emission & $29,349$ & $11$ \\
            MetroPT-3          & $50,000$ & $7$ \\
            \bottomrule
        \end{tabular}
    \end{subtable}
\end{tabular}
\end{table}

\begin{table}[h]
\centering
\footnotesize
\caption{The p-values of the median comparison between NR and UCV in the high-sample size scenarios of the first experiment on real data}
\label{tab:real_data_pvalues}
\renewcommand{\arraystretch}{1.2} 
\begin{tabular}{l c}
\toprule
\textbf{Dataset} & \textbf{p-values} \\
\midrule
WineQuality-White    & $6e-04$ \\
Waveform-Noise     & $0.8704$ \\
Travel-Ratings    & $0.0040$ \\
Magic Gamma  & $8e-04$ \\
Gas Turbine CO and NOx Emission  & $0.0584$ \\
MetroPT-3    & $0.0090$ \\

\bottomrule
\end{tabular}
\end{table}

The Glass dataset presents a moderate number of variables and a relatively low sample size. The observed patterns in this dataset align with those found in the synthetic data experiments, where the proposed methods demonstrated moderate performance in this type of scenario.

Among the evaluated approaches, NR achieves a robust median value. However, when SCV and PI is utilized for learning the parameters (NR-SCV, NR-PI), both outperform NR, highlighting the effectiveness of these methods in this type of setting. This result is consistent with findings from the synthetic dataset experiments. Nevertheless, the observed performance differences in terms of medians do not exhibit statistical significance, see Figure \ref{fig:sa_real_data_1}. The same can be observed for UCV-SCV and UCV-PI.

The Ionosphere dataset presents a markedly different scenario, characterized by a very low sample size combined with a high number of variables. Under these conditions, NR proves to be robust and performs well, as seen in the synthetic data. Nonetheless, SCV once again demonstrates its usefulness by achieving a superior median performance compared to NR. However, this difference is not statistically significant. Moreover, UCV does not yield competitive results, while SCV and PI can also underperform in certain cases.

Furthermore, the QSAR Aquatic dataset presents a moderate number of variables, as in the Glass dataset, with a relatively increased  sample size. NR outperforms all proposed methods, with statistically significant differences which is a possible results in this type of settings. 

The Breast Cancer and Glass datasets share similar characteristics, such as a moderate number of variables and a relatively small sample size. Additionally, the same analysis and results presented for the Glass dataset hold for the Breast Cancer dataset.

In the QSAR Fish Toxicity dataset, the number of variables decreases while the sample size increases. This shift leads to a noticeable change in UCV’s performance, aligning with the patterns observed in the synthetic datasets. Again, in low-sample size conditions, SCV and PI prove to be effective, as they outperform NR when the structure is learned using this method and the parameters with the formers. However, both SCV and PI exhibit weaker results when UCV is used for learning, reinforcing the recommendation that UCV should be employed independently and especially in high-sample size scenarios. In Figure \ref{fig:sa_real_data_1}, UCV demonstrates statistically significant differences compared to all other methods.

\afterpage{\clearpage}

The results for the Yeast dataset also align with the findings from the synthetic experiments. In the Yeast dataset, the number of variables is moderate while the sample size increases, leading to improved performance of UCV. Similar to the QSAR Fish Toxicity scenario, SCV and PI demonstrate their effectiveness in this relatively low sample size setting by outperforming NR (NR-SCV, NR-PI). However, as observed previously, their performance declines when the structure is learned using UCV. Statistically, UCV exhibits significant differences compared to all other methods. Additionally, NR also shows statistically significant differences in terms of median performance, ranking as the worst-performing method in this dataset.

For higher-sample-size scenarios, the WineQuality-White dataset results prove the superior performance of UCV in higher-sample-size settings with a moderate number of variables. The p-values in Table \ref{tab:real_data_pvalues} indicate a high effect size in favor of UCV’s median performance against NR.

In the Waveform-Noise dataset, the increased complexity, due to the presence of 40 variables, impacts UCV's performance. Although UCV slightly outperforms NR in this case, the difference is not statistically significant in terms of medians. 

Moreover, in the Travel Ratings dataset, we observe results similar to those of the WineQuality-White dataset, with UCV outperforming NR. The low p-value in the median comparison indicates a substantial difference, consistent with the findings from the synthetic data experiments and further highlighting the effectiveness of UCV. However, UCV produces the worst result in one of the ten iterations due to its variability (note the outlier). This contrast between UCV’s variability and NR’s robustness is evident across all datasets.

The results in Figure \ref{fig:boxplots_real_data_1} highlight the Magic Gamma dataset. Despite the dataset having a moderate number of variables, UCV is significantly outperformed by NR, as indicated by the p-value in Table \ref{tab:real_data_pvalues}, with a substantial effect size. This finding deviates from the general pattern observed, suggesting that NR cannot always be disregarded.

Finally, in the highest sample size scenarios with a moderate number of variables (Gas Turbine CO and NOx Emission and MetroPT-3), UCV outperforms NR, following the synthetic patterns. Although no statistically significant difference in medians is observed for Gas Turbine CO and NOx Emission, the p-value is close to 0.05, suggesting that the difference may not be negligible. The variability of UCV is evident in the results, contrasting with the stability of NR.

\subsubsection{Second experiment}
\label{sec:second_experiment}
The synthetic experiments highlight the advantage of the proposed bandwidth selectors, particularly UCV, over NR. While UCV improves as more data becomes available, NR stagnates due to the bias introduced by its underlying assumption. For checking this, we use the MetroPT-3 dataset, which is well-suited for the task due to its large volume. To compare UCV and NR in an increasing data scenario, we generate samples ranging from 10,000 to 50,000 in increments of 10,000.

We perform a 10-fold split into training and validation sets for the five different sampled datasets. The experiment and results analysis follows the same scheme as in Section \ref{sec:first_experiment}.

\begin{figure}[h]
\centering 
\includegraphics[scale = 0.32]{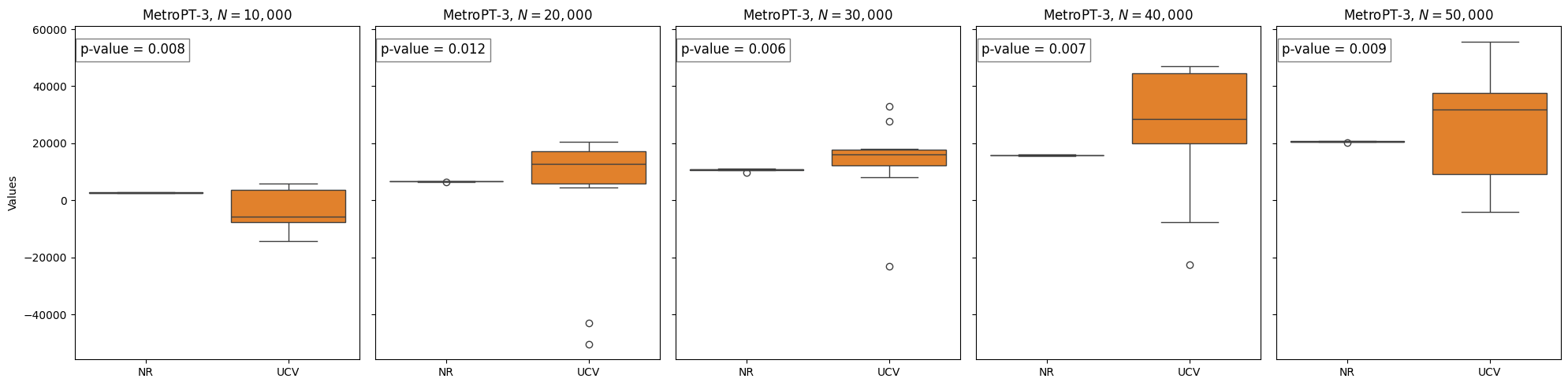}
\caption{Boxplots of the results from the second experiment on real data using the MetroPT-3 dataset. The p-value for each comparison is shown in the upper-left corner of each boxplot.
} 
\label{fig:boxplots_real_data_2}
\end{figure}

Figure \ref{fig:boxplots_real_data_2} shows that for the 10,000-sample case, NR statistically outperforms UCV in terms of medians. However, as the sample size increases, the previously observed pattern of NR stagnation versus UCV improvement is reaffirmed, with UCV consistently outperforming NR with statistically significant differences. While the p-value for the 20,000-sample case is not as low as for other sizes, the p-values remain consistently low overall, indicating a strong effect size.

\subsection{Discussion}
The results presented in Section \ref{sec:experiments} demonstrate the effectiveness of the proposed plug-in and cross-validation bandwidth selectors across different scenarios. SCV, PI, and especially UCV emerge as viable alternatives to NR, consistently leading to better density estimates for each CKDE and, consequently, for the joint PDF.

In Section \ref{subsectionPL}, we analyze synthetic data, exploring a diverse range of scenarios that vary in the number of instances, variables, structures, and densities. These scenarios differ in complexity, particularly in terms of smoothness. We compare the proposed bandwidth selectors against the conventional NR for CKDEs, as well as against one another. From this analysis, we derive several key conclusions.

Given the known fixed structures of the different SPBNs, the bandwidth selection methods we propose achieve the best results for learning the conditional distributions. In terms of absolute error in log-likelihood, our approach generally outperforms NR in terms of medians, with statistically significant differences. Additionally, we identify two main patterns: as the complexity of the underlying density increases, the performance declines for the proposed bandwidth selection methods, whereas NR remains robust due to oversmoothing, yielding competitive results. However, with larger sample sizes—i.e., more information—the performance of the proposed bandwidth selection improves, reinforcing its effectiveness and outperforming NR. 

\begin{table}[h]
\centering
\caption{Summary of execution times (s) for the synthetic data experiments}
\label{tab:results_executiontimes_fixed_structure}
\scriptsize 
\setlength{\tabcolsep}{4pt} 
\begin{tabular}{@{}p{1cm}p{1.2cm}p{1cm}cc cc cc cc@{}}
\toprule
\textbf{Nodes} & \textbf{Density} & \textbf{Bandwidth Method} & \multicolumn{2}{c}{$\mathbf{N = 200}$} & \multicolumn{2}{c}{$\mathbf{N = 2000}$} & \multicolumn{2}{c}{$\mathbf{N = 10000}$} & \multicolumn{2}{c}{$\mathbf{N = 20000}$} \\
\midrule
& & & \textbf{Median} & \textbf{Highest} & \textbf{Median} & \textbf{Highest} & \textbf{Median} & \textbf{Highest} & \textbf{Median} & \textbf{Highest} \\
\midrule
\multirow{12}{*}{5} 
    & \multirow{4}{*}{Smooth} 
        & NR &  $\phantom{0}0\phantom{.10}$ & $\phantom{0}1.71$ & $\phantom{0}0\phantom{.00}$ & $\phantom{000}1.61$ & $0\phantom{.00}$ & $0.1$ & $\phantom{0}0\phantom{.00}$ & $\phantom{00}0.08$ \\
    & & UCV & $\phantom{0}0.11$ & $\phantom{0}0.2\phantom{0}$ & $\phantom{000}0.7\phantom{.0}$ & $\phantom{000}2.23$ & $26.63$ & $27.98$ & $96.34$ & $115.17$ \\
    & & SCV & $46.56$ & $63.72$ & $1432.46$ & $6518.55$ & -- & -- & -- & -- \\
    & & PI  & $16.03$ & $16.69$ & $1383.07$ & $4565.61$ & -- & -- & -- & -- \\
    \cmidrule(lr){2-11}
    & \multirow{4}{*}{Medium} 
        & NR & $\phantom{0}0\phantom{.00}$ & $\phantom{0}0.91$ & $\phantom{00}0.55$ & $\phantom{00}1.34$ & $\phantom{0}0\phantom{0.0}$ & $\phantom{0}0.1\phantom{0}$ & $\phantom{00}0\phantom{.00}$ & $\phantom{00}0.1\phantom{0}$ \\
    & & UCV & $\phantom{0}0.11$ & $\phantom{0}0.17$ & $\phantom{00}2.95$ & $\phantom{00}3.38$ & $56.45$ & $75.12$ & $304.54$ & $345.41$ \\
    & & SCV & $57.22$ & $68.28$ & $788.75$ & $806.61$ & -- & -- & -- & -- \\
    & & PI  & $15.88$ & $16.02$ & $742.46$ & $745.33$ & -- & -- & -- & -- \\
    \cmidrule(lr){2-11}
    & \multirow{4}{*}{Rough} 
        & NR & $\phantom{0}0\phantom{.00}$ & $\phantom{0}1.49$ & $0$ & $\phantom{000}0.93$ & $\phantom{000}0\phantom{.00}$ & $\phantom{000}0.07$ & $\phantom{00}0\phantom{.00}$ & $\phantom{00}0.07$ \\
    & & UCV & $\phantom{0}0.14$ & $\phantom{0}0.18$ & $\phantom{000}1.21$ & $\phantom{000}1.53$ & $64.24$ & $99.34$ & $205.25$ & $234.72$ \\
    & & SCV & $50.78$ & $82.17$ & $1426.92$ & $1445.96$ & -- & -- & -- & -- \\
    & & PI  & $16.07$ & $16.14$ & $1383.06$ & $1383.11$ & -- & -- & -- & -- \\
\midrule
\multirow{12}{*}{10} 
    & \multirow{4}{*}{Smooth} 
        & NR &  $\phantom{000}0\phantom{.00}$ & $\phantom{0000}0.38$ & $\phantom{0000}0\phantom{.00}$ & $\phantom{0000}0.13$ & $\phantom{000}0.01$ & $\phantom{000}0.11$ & $\phantom{000}0.01$ & $\phantom{000}0.12$ \\
    & & UCV & $\phantom{000}0.47$ & $\phantom{0000}0.61$ & $\phantom{000}35.30$ & $\phantom{000}58.1$ & $1123.47$ & $1618.78$ & $5459.44$ & $7602.21$ \\
    & & SCV & $13251.3$ & $16927.34$ & $75926.74$ & $93350.15$ & -- & -- & -- & -- \\
    & & PI  & $8397.73$ & $8400.24$ & $52479.61$ & $52711.56$ & -- & -- & -- & -- \\
    \cmidrule(lr){2-11}
    & \multirow{4}{*}{Medium} 
        & NR & $\phantom{000}0.77$ & $\phantom{000}1.64$ & $\phantom{0000}1.03$ & $\phantom{0000}1.19$ & $\phantom{00}0.01$ & $\phantom{00}0.11$ & $\phantom{000}0.01$ & $\phantom{000}0.1$ \\
    & & UCV & $\phantom{000}1.08$ & $\phantom{000}1.37$ & $\phantom{000}32.79$ & $\phantom{000}41.6$ & $620.79$ & $655.14$ & $2224.89$ & $2559.46$ \\
    & & SCV & $5370.21$ & $7565.37$ & $66401.85$ & $77521.11$ & -- & -- & -- & -- \\
    & & PI  &  $1393.94$ & $1513.37$ & $54207.02$ & $54301.28$ & -- & -- & -- & -- \\
    \cmidrule(lr){2-11}
    & \multirow{4}{*}{Rough} 
        & NR & $\phantom{0000}0\phantom{.00}$ & $\phantom{0000}0.41$ & $\phantom{0000}0.01$ & $\phantom{00000}0.21$ & $\phantom{000}0.01$ & $\phantom{000}0.1$ & $\phantom{000}0.01$ &  $\phantom{000}0.1$ \\
    & & UCV & $\phantom{0000}0.46$ & $\phantom{0000}0.58$ & $\phantom{00000}9.5$ & $\phantom{0000}12.44$ & $1656.51$ & $1974.67$ & $6186.02$ & $7339.68$ \\
    & & SCV & $16346.19$  & $18720.03$ & $97090.56$ & $114177.22$ & -- & -- & -- & -- \\
    & & PI  & $8397.2$ & $8398.34$  & $78858.54$ & $79451.69$ & -- & -- & -- & -- \\
\midrule
\multirow{12}{*}{15} 
    & \multirow{4}{*}{Smooth} 
        & NR &  $\phantom{000}0\phantom{.00}$ & $\phantom{0000}0.23$ & $\phantom{0000}0.01$ & $\phantom{0000}2.68$ & $\phantom{000}0\phantom{.00}$ & $\phantom{000}0.11$ & $\phantom{000}0.01$ & $\phantom{000}0.23$ \\
    & & UCV & $\phantom{000}1.44$ & $\phantom{0000}1.73$ &  $\phantom{000}48.72$ & $\phantom{000}72.99$ & $1646.41$ & $1930.03$ & $6045.36$ & $7622.78$ \\
    & & SCV & $1132.87$ & $31590.35$ & $53604.08$ & $55754.14$ & -- & -- & -- & -- \\
    & & PI  & $922.54$ & $924.56$ & $52760.53$ & $52912.44$ & -- & -- & -- & -- \\
    \cmidrule(lr){2-11}
    & \multirow{4}{*}{Medium} 
        & NR & $\phantom{000}1.38$ & $\phantom{000}1.63$ &  $\phantom{0000}0.05$ & $\phantom{0000}0.19$ & $\phantom{000}0\phantom{.00}$ & $\phantom{000}0.11$ & $\phantom{000}0.01$ & $\phantom{000}0.1\phantom{0}$ \\
    & & UCV & $\phantom{000}0.79$ & $\phantom{00}10.18$ & $\phantom{0000}9.69$ & $\phantom{000}12.01$ & $1610.67$ & $2036.25$ & $6731.10$ & $7534.14$ \\
    & & SCV & $6841.22$ & $8332.88$ & $83599.71$ & $84442.41$ & -- & -- & -- & -- \\
    & & PI  & $1391.18$ & $1402.28$ & $78963.09$ & $79288.33$ & -- & -- & -- & -- \\
    \cmidrule(lr){2-11}
    & \multirow{4}{*}{Rough} 
        & NR & $\phantom{0000}0\phantom{.00}$ & $\phantom{00000}0.14$ & $\phantom{0000}0\phantom{.00}$ & $\phantom{0000}0.17$ & $\phantom{000}1.04$ & $\phantom{000}1.19$ & $\phantom{000}1.04$ &  $\phantom{000}1.05$ \\
    & & UCV & $\phantom{000}1.48$ & $\phantom{00000}1.75$ & $\phantom{000}54.53$ & $\phantom{000}72.39$ & $1263.65$ & $1363.48$ & $5161.13$ & $5433.39$ \\
    & & SCV & $9244.81$ & $114395.64$ & $60968.74$ & $82962.52$ & -- & -- & -- & -- \\
    & & PI  & $925.51$ & $946.87$ & $52743.86$ & $79147.42$ & -- & -- & -- & -- \\
\bottomrule
\end{tabular}
\end{table}

Overall, UCV performs best. Although SCV and PI are theoretically expected to outperform UCV, our results show that they only yield the best performance in a limited number of scenarios. This discrepancy may stem from the optimization algorithm. As discussed in Section \ref{sec:bandwidth_selection}, while SCV and PI have strong theoretical foundations, their reliance on an optimization procedure can significantly impact their practical performance, highlighting a potential future research direction—exploring alternative optimization algorithms.

Nevertheless, based on our results, we recommend using SCV or PI in scenarios with small sample sizes. This conclusion is further supported by computational efficiency considerations. As shown in Table \ref{tab:results_executiontimes_fixed_structure}, the introduced bandwidth selectors are computationally expensive, with SCV and PI being the most costly. This suggests that their application should be prioritized for small sample sizes.

Among all methods, NR is the most computationally efficient due to its simplicity, yielding highly competitive computational performance compared to the other proposed bandwidth selectors. Finally, UCV, while not as computationally expensive as SCV and PI, delivers strong results, making it particularly suitable for high-sample-size scenarios in real-world applications—a trend that is also evident in our real experiments.

Consistent with the results observed in the parameter learning experiments, the structural learning experiments using the proposed bandwidth selector reveal the same two key patterns, albeit in a more pronounced manner. In these experiments, NR, which ensures robustness in structure fitting, delivers highly competitive results. In certain scenarios, it achieves the lowest log-likelihood error and the best structural recovery performance. However, it still exhibits stagnation as the sample size increases.

UCV remains a strong competitor, particularly in high-sample-size settings, where it also achieves the best log-likelihood error and structural recovery performance. It is also important to note that, in some scenarios, both algorithms performed worse in terms of log-likelihood but better in structural recovery compared to one another. This outcome is a consequence of the design of score-based structure learning methods.

Moreover, the same pattern of stagnation vs improvement is present in the second real experiment (Section \ref{sec:second_experiment}), reinforcing the conclusions. Also, these findings suggest a potential complementary use of both methods: NR for structural learning and UCV for the final parameter learning. This conclusion aligns with the results observed in both structural and parameter learning and computational costs. 

Finally, SCV and PI exhibit moderate performance, likely due to limitations in the optimization algorithm. Nevertheless, in real experiments with small sample sizes, they cannot be entirely dismissed, as they achieve the best results in some cases, albeit without statistically significant differences.

Overall, the proposed bandwidth selectors effectively fulfill their purpose, as they play a crucial role in defining the smoothing parameter $\mathbf{H}$ in CKDEs, addressing the trade-off between bias and variance in the approximations. As stated in \cite{chacon2018multivariate}, NR and straightforward bandwidth selectors should be used for exploratory purposes, whereas the proposed bandwidth selectors are more appropriate for density estimation. This aligns with the findings of this paper, demonstrating that SPBNs can leverage the extensive literature on bandwidth selection.

\section{Conclusions}
\label{sec:conclusions}
In this paper, we have introduced a comprehensive theoretical framework for incorporating state-of-the-art bandwidth selectors into SPBNs. Through extensive experimentation, we have demonstrated their effectiveness in enhancing density estimation, particularly for conditional distributions, ultimately leading to improved joint density estimates.

Our findings underscore the distinct strengths of various bandwidth selection methods—UCV, PI, and SCV—across different learning scenarios. These advanced selectors have proven instrumental in enhancing the learning capabilities of SPBNs through the improvement of CKDEs estimations, refining their adaptability, and elevating their overall performance in parameter and structure learning. 

For parameter learning, UCV consistently achieves the best overall performance, making it a strong choice, particularly in high-sample-size settings. In contrast, for structural learning, NR is highly competitive, especially in small-sample-size scenarios. Notably, UCV also performs well in structural learning. These results indicate that using UCV, especially in high-sample-size scenarios, enhances the learning capabilities of SPBNs in both parameter and structural learning. Furthermore, the findings suggest that a hybrid approach combining NR and UCV could further optimize SPBNs' estimation performance.

Despite these promising results, our study reveals that SCV and PI do not fully exhibit their theoretical advantages, opening avenues for future research. Investigating alternative optimization algorithms while balancing accuracy and computational cost could further refine these methods. Also, although UCV remains computationally expensive, its strong performance justifies its use, and exploring optimizations in this direction would be valuable.

Moreover, expanding this research to include other bandwidth selection strategies could offer new insights, particularly for online learning applications, where computational efficiency is crucial. Future studies should explore these possibilities to refine and extend the applicability of SPBNs in real-world scenarios.

\section*{Research data statement}
The datasets used in this work include publicly available UCI datasets, which can be accessed via the UCI Machine Learning Repository. Additionally, synthetic datasets developed by the authors, as well as the modified PyBNesian package and the associated experiments code, are available in the first author's GitHub repository at https://github.com/Victor-Alejandre/BandwidthSelection.git. All data and code used in this study are publicly accessible for reproducibility.

\section*{Acknowledgments}
The authors gratefully acknowledge the Universidad Politécnica de Madrid (www.upm.es) for providing computing resources on Magerit Supercomputer.

\section*{Funding sources}
This work was partially supported by the Ministry of Science, Innovation and Universities under Project AEI/10.13039/501100011033-PID2022-139977NB-I00, Project TED2021-131310B-I00, and \newline
Project PLEC2023-010252/MIG-20232016. Also, by the Autonomous Region of Madrid under Project ELLIS Unit Madrid and TEC-2024/COM-89. Finally, this work has also been funded through the R+D activity program with reference TEC-2024/COM-89 and acronym IDEA-CM, granted by the Community of Madrid through the Dirección General de Investigación e Innovación Tecnológica under Orden2402/2024, dated May 31.

\section*{Appendices}
\appendix

\section{Proofs}\label{apd:first}

\paragraph{Proof of Theorem \ref{theorem1}}\mbox{}\\
\textbf{\textit{Theorem}.}
Given a SPBN's CKDE, if conditions (B1)-(B3) hold, then the CKDE (Definition \ref{definition3:1}) is a consistent estimator of $f(x_j\mid \mathbf{x}_{Pa(j)})$:
\[\hat{f}(x_j\mid \mathbf{x}_{Pa(j)};\mathbf{H})\xrightarrow{P}f(x_j\mid \mathbf{x}_{Pa(j)})\]
\begin{pf}
To prove the consistency of CKDE (Equation \ref{eq:CKDE}), we first establish the consistency of its numerator and denominator KDEs, $\hat{f}(x_j,\mathbf{x}_{Pa(j)};\mathbf{H}(X_j, \mathbf{X}_{Pa(j)}))$ and $\hat{f}(\mathbf{x}_{Pa(j)};\mathbf{H}( \mathbf{X}_{Pa(j)}))$ respectively. Since CKDE is defined as their ratio, its consistency will be proven upon the consistency of these individual estimators. Without loss of generality, we assume that $\hat{f}(\mathbf{x}_{j};\mathbf{H}(\mathbf{X}_{Pa(j)})) > 0$, as conditioning on a point with zero mass leads to an undefined expression, which is typically avoided in density estimation.

MISE convergence implies consistency of the KDE estimator. We first establish this convergence for the numerator, and then for the denominator. First notice that condition (A2) (Section \ref{item:conditionA1}) is fulfilled by Gaussian kernels. Therefore, given conditions (B1) and (B3) (Section \ref{item:conditionB1}) we have that when $N\rightarrow \infty$ (\cite{chacon2018multivariate}):
\[\text{MISE}\left\{\hat{f}(x_j,\mathbf{x}_{Pa(j)};\mathbf{H}(X_j, \mathbf{X}_{Pa(j)}))\right\}\rightarrow 0\implies\hat{f}(x_j,\mathbf{x}_{Pa(j)};\mathbf{H}(X_j, \mathbf{X}_{Pa(j)}))\xrightarrow{P}f(x_j,\mathbf{x}_{Pa(j)}).\]

Once MISE convergence is established for the numerator, we proceed similarly for the denominator. Thus, we demonstrate that the sequence $\mathbf{H}(\mathbf{X}{Pa(j)}) = \mathbf{H}(\mathbf{X}{Pa(j)})_N$ consists of symmetric, positive definite matrices satisfying the following properties:
$\operatorname{vec}\mathbf{H}(\mathbf{X}{Pa(j)}) \rightarrow \mathbf{0}_{(d-1)^2}$ and $N^{-1}|\mathbf{H}(\mathbf{X}_{Pa(j)})|^{-1/2} \rightarrow 0$ as $N \rightarrow \infty$. To establish these results, we employ Cauchy's interlacing theorem (Theorem \ref{theorem3}) (\cite{horn2012matrix}) and condition (B3). \\
    
Since $\mathbf{H}(\mathbf{X}_{Pa(j)}) = \mathbf{M}(\mathbf{H}(X_j,\mathbf{X}_{Pa(j)}))$ (see Equation \ref{eq:bandwidth_relationship}) it is easy to prove that $vec\mathbf{H}( \mathbf{X}_{Pa(j)})\rightarrow  \mathbf{0}_{(d-1)^2} $ as $N\rightarrow
\infty$ given condition (B3) which states that $vec\mathbf{H}( X_j,\mathbf{X}_{Pa(j)})\rightarrow  \mathbf{0}_{d^2} $. In order to prove $N^{-1}|\mathbf{H}(\mathbf{X}_{Pa(j)})|^{-1/2}\rightarrow 0$ we consider $|\mathbf{H}(\mathbf{X}_{Pa(j)})| = \mu_1\mu_2\cdots\mu_{d-1}$ and $|\mathbf{H}(X_j, \mathbf{X}_{Pa(j)})| = \lambda_1\lambda_2\cdots\lambda_{d}$. Then:
\[\lambda_d^{-1/2} N^{-1}|\mathbf{H}(\mathbf{X}_{Pa(j)})|^{-1/2} = \lambda_d^{-1/2} N^{-1}(\mu_1\mu_2\cdots\mu_{d-1})^{-1/2} \]
\[\stackrel{CIT}{\leq} \lambda_d^{-1/2} N^{-1}(\lambda_1\lambda_2\cdots\lambda_{d-1})^{-1/2} = N^{-1}|\mathbf{H}(X_j, \mathbf{X}_{Pa(j)})|^{-1/2}\]

Therefore, $N^{-1}|\mathbf{H}(\mathbf{X}_{Pa(j)})|^{-1/2}\leq \lambda_d^{1/2}N^{-1}|\mathbf{H}(X_j, \mathbf{X}_{Pa(j)})|^{-1/2}\rightarrow 0\hspace{2mm}\text{when }N\rightarrow\infty,$ since  $\lambda_d\rightarrow 0$ as $|\mathbf{H}(X_j, \mathbf{X}_{Pa(j)})|\rightarrow 0$, due to $vec\mathbf{H}(X_j, \mathbf{X}_{Pa(j)})\rightarrow  \mathbf{0}_{d^2} $, and condition (B3) implies that $$N^{-1}|\mathbf{H}(X_j, \mathbf{X}_{Pa(j)})|^{-1/2}\rightarrow 0\hspace{2mm}\text{when }N\rightarrow\infty.$$

Now, since conditions (A2) and (A3) hold for $\hat{f}(\mathbf{x}_{Pa(j)};\mathbf{H}(\mathbf{X}_{Pa(j)})$ as proven, it follows that, under condition (B2), $\hat{f}(\mathbf{x}_{Pa(j)};\mathbf{H}(\mathbf{X}_{Pa(j)}))$ converges in MISE (\cite{bosq1987th}). Consequently, this ensures that $\hat{f}(\mathbf{x}_{Pa(j)};\mathbf{H}( \mathbf{X}_{Pa(j)}))$ converges in probability.

For ease of notation, we define \(\widehat{f}(\mathbf{x}_{Pa(j)};\mathbf{H}(\mathbf{X}_{Pa(j)}))\) as \(\widehat{f}(\mathbf{x}_{Pa(j)})\) and  
\(\widehat{f}(x_j, \mathbf{x}_{Pa(j)};\mathbf{H}(X_j, \mathbf{X}_{Pa(j)}))\) as \(\widehat{f}(x_j, \mathbf{x}_{Pa(j)})\). We now conclude the proof of the theorem by establishing the convergence in probability of  
\(\widehat{f}(x_j \mid \mathbf{x}_{Pa(j)}; \mathbf{H})\), given that  
\(\widehat{f}(x_j, \mathbf{x}_{Pa(j)}) \xrightarrow{P} f(x_j, \mathbf{x}_{Pa(j)})\)  
and  
\(\widehat{f}(\mathbf{x}_{Pa(j)}) \xrightarrow{P} f(\mathbf{x}_{Pa(j)})\).  

From the definition of convergence in probability and the given assumptions, we have:
\begin{itemize}
    \item $\hat{f}(x_j,\mathbf{x}_{Pa(j)}) \xrightarrow{P} f(x_j,\mathbf{x}_{Pa(j)})$, i.e., for any $\epsilon > 0$, $ \lim_{N \to \infty} P(|\hat{f}(x_j,\mathbf{x}_{Pa(j)}) - f(x_j,\mathbf{x}_{Pa(j)})| \geq \epsilon) = 0.
    $
    \item $\hat{f}(\mathbf{x}_{Pa(j)}) \xrightarrow{P} f(\mathbf{x}_{Pa(j)})$, i.e., for any $\delta > 0$, $
    \lim_{N \to \infty} P(|\hat{f}(\mathbf{x}_{Pa(j)}) - f(\mathbf{x}_{Pa(j)})| \geq \delta) = 0.
    $
\end{itemize}

Again, for ease of notation, we consider the limit values as $a = f(x_j,\mathbf{x}_{Pa(j)})$ and $b = f(\mathbf{x}_{Pa(j)})$. Consequently, we want to show that:
\[
\frac{\hat{f}(x_j, \mathbf{x}_{Pa(j)})}{\hat{f}(\mathbf{x}_{Pa(j)})} \xrightarrow{P} \frac{a}{b}.
\]

By definition, we need to show that for any $\rho > 0$,
\[
\lim_{N \to \infty} P\left( \left| \frac{\hat{f}(x_j, \mathbf{x}_{Pa(j)})}{\hat{f}(\mathbf{x}_{Pa(j)})} - \frac{a}{b} \right| \geq \rho \right) = 0.
\]

Consider the difference:
\[
\left| \frac{\hat{f}(x_j, \mathbf{x}_{Pa(j)})}{\hat{f}(\mathbf{x}_{Pa(j)})} - \frac{a}{b} \right| = \left| \frac{\hat{f}(x_j, \mathbf{x}_{Pa(j)})b - \hat{f}(\mathbf{x}_{Pa(j)})a}{\hat{f}(\mathbf{x}_{Pa(j)})b} \right| = \left| \frac{b(\hat{f}(x_j, \mathbf{x}_{Pa(j)}) - a) + a(b - \hat{f}(\mathbf{x}_{Pa(j)}))}{\hat{f}(\mathbf{x}_{Pa(j)})b} \right|.
\]

Using the triangle inequality:
\begin{equation}
\begin{aligned}[b]
\left| \frac{b(\hat{f}(x_j, \mathbf{x}_{Pa(j)}) - a) + a(b - \hat{f}(\mathbf{x}_{Pa(j)}))}{\hat{f}(\mathbf{x}_{Pa(j)})b} \right| 
&\leq \left| \frac{b(\hat{f}(x_j, \mathbf{x}_{Pa(j)}) - a)}{\hat{f}(\mathbf{x}_{Pa(j)})b} \right| + \left| \frac{a(b - \hat{f}(\mathbf{x}_{Pa(j)}))}{\hat{f}(\mathbf{x}_{Pa(j)})b} \right| \\
&= \frac{|\hat{f}(x_j, \mathbf{x}_{Pa(j)}) - a|}{|\hat{f}(\mathbf{x}_{Pa(j)})|} + \frac{|a||b - \hat{f}(\mathbf{x}_{Pa(j)})|}{|b||\hat{f}(\mathbf{x}_{Pa(j)})|}.
\end{aligned}
\tag{1}
\end{equation}

Now we choose $\eta > 0$ such that $|b| - \eta > 0$ and sufficiently small such that:
\begin{equation}
\frac{|a|\, \eta}{|b| (|b| - \eta)} < \frac{\rho}{2}
\tag{3}
\end{equation}
which exists as $|b| > 0$. Consider the events:
\[
A = \left\{ |\hat{f}(x_j, \mathbf{x}_{Pa(j)}) - a| \geq \frac{\rho}{2} (|b| - \eta) \right\},
\]
\[
B = \left\{ |\hat{f}(\mathbf{x}_{Pa(j)}) - b| \geq \eta \right\}.
\]

From the definitions of convergence in probability:
\[
\lim_{N \to \infty} P(A) = 0 \quad \text{and} \quad \lim_{N \to \infty} P(B) = 0,
\]

the event $C= A^c \cap B^c$ exists, where both $A$ and $B$ do not occur. This means that $\exists N^{'}$ such that for $N\geq N{'}$:
\begin{equation}
|\hat{f}(x_j, \mathbf{x}_{Pa(j)}) - a| < \frac{\rho}{2} (|b| - \eta)
\hspace{3mm} \text{and} \hspace{3mm}
|\hat{f}(\mathbf{x}_{Pa(j)}) - b| < \eta.
\tag{2}
\end{equation}

On the event $C$, we have $|\hat{f}(\mathbf{x}_{Pa(j)}) - b| < \eta$, which implies $|\hat{f}(\mathbf{x}_{Pa(j)})| > |b| - \eta$. Therefore:
\[
\left| \frac{\hat{f}(x_j, \mathbf{x}_{Pa(j)})}{\hat{f}(\mathbf{x}_{Pa(j)})} - \frac{a}{b} \right| 
\overset{(1)}{\leq}
\frac{|\hat{f}(x_j, \mathbf{x}_{Pa(j)}) - a|}{|\hat{f}(\mathbf{x}_{Pa(j)})|} + \frac{|a||b - \hat{f}(\mathbf{x}_{Pa(j)})|}{|b||\hat{f}(\mathbf{x}_{Pa(j)})|} 
\overset{(2)}{<}
\frac{\frac{\rho}{2} (|b| - \eta)}{|b| - \eta} + \frac{|a| \eta}{|b| (|b| - \eta)}
\overset{(3)}{<}
\frac{\rho}{2}+\frac{\rho}{2}.
\]

Hence:
\[
\left| \frac{\hat{f}(x_j, \mathbf{x}_{Pa(j)})}{\hat{f}(\mathbf{x}_{Pa(j)})} - \frac{a}{b} \right| < \rho.
\]

It follows that:
\[
\lim_{N \to \infty} P\left( \left| \frac{\hat{f}(x_j, \mathbf{x}_{Pa(j)})}{\hat{f}(\mathbf{x}_{Pa(j)})} - \frac{a}{b} \right| \geq \rho \right) = 0.
\]

Therefore, we have shown that $$\frac{\hat{f}(x_j, \mathbf{x}_{Pa(j)})}{\hat{f}(\mathbf{x}_{Pa(j)})} \xrightarrow{P} \frac{a}{b}$$.
\end{pf}

\section{Synthetic Densities}\label{apd:second}
\textbf{Smooth} \\ 
\[
\footnotesize
\begin{aligned}
    &\bullet \quad f(x_1) = \mathcal{N}(0, 1) \\
    &\bullet \quad f(x_2) = 0.5 \mathcal{N}(-2, 2^2) + 0.5 \mathcal{N}(2, 2^2) \\
    &\bullet \quad f(x_3 \mid x_1, x_2) = \mathcal{N}(x_1 x_2, 1) \\
    &\bullet \quad f(x_4 \mid x_3) = \mathcal{N}(10 + 0.8 x_3, 0.5^2) \\
    &\bullet \quad f(x_5 \mid x_4) = \mathcal{N}\left(\frac{1}{1 + e^{-x_4}}, 0.5^2\right)
\end{aligned} \\
\]

\noindent \textbf{Medium}  \\
\[
\footnotesize
\begin{aligned}
    &\bullet \quad f(x_1) = \mathcal{N}(0, 3^2) \\
    &\bullet \quad f(x_2) = 0.25 \mathcal{N}(-8, 1^2) + 0.25 \mathcal{N}(8, 1^2) + 0.25 \mathcal{N}(-4, 0.5^2) + 0.25 \mathcal{N}(4, 0.5^2)\\
    &\bullet \quad f(x_3 \mid x_1, x_2) = 
    w_1 \mathcal{N}(-0.5 x_1, 1) + 
    w_2 \mathcal{N}(0.5 x_2, 1) + 
    w_3 \mathcal{N}(0, 1), \\ 
    &\begin{aligned}
        \hspace{0.7cm}w_1 &= \frac{x_1^2}{x_1^2 + x_2^2 + 2|x_1 x_2|}, 
        \hspace{0.7cm}w_2 &= \frac{x_2^2}{x_1^2 + x_2^2 + 2|x_1 x_2|},
        \hspace{0.7cm}w_3 &= \frac{2|x_1 x_2|}{x_1^2 + x_2^2 + 2|x_1 x_2|}
    \end{aligned}\\
    &\bullet \quad f(x_4 \mid x_3) = \mathcal{N}(0.5 x_3, 0.5^2) \\
    &\bullet \quad f(x_5 \mid x_4) = \lambda \mathcal{N}(-2, 0.5^2) + (1 - \lambda) \mathcal{N}(2, 0.5^2), \quad
    \lambda = \frac{1}{1 + \exp\left(-x_4/3\right)}
\end{aligned} \\
\]

\noindent \textbf{Rough} \\
\[
\footnotesize
\begin{aligned}
    &\bullet \quad f(x_1) = \mathcal{N}(0, 3^2) \\
    &\bullet \quad f(x_2) = 0.2 \mathcal{N}(-8, 1^2) + 0.2 \mathcal{N}(8, 1^2) + 0.2 \mathcal{N}(-4, 0.5^2) + 0.2 \mathcal{N}(0, 0.5^2) + 0.2 \mathcal{N}(4, 0.5^2)\\
    &\bullet \quad f(x_3 \mid x_1, x_2) = 
    w_1 \mathcal{N}(-0.5 x_1^2, 1) + 
    w_2 \mathcal{N}(0.5 x_2^2, 1) + 
    w_3 \mathcal{N}(0.5 x_1, 1) + 
    w_4 \mathcal{N}(-0.5 x_2, 1), \\
    &\begin{aligned}
        \hspace{0.7cm}w_1 &= \frac{x_1^2}{x_1^2 + x_2^2 + |x_1| + |x_2|},
        \hspace{0.4cm}w_2 &= \frac{x_2^2}{x_1^2 + x_2^2 + |x_1| + |x_2|},
        \hspace{0.4cm}w_3 &= \frac{|x_1|}{x_1^2 + x_2^2 + |x_1| + |x_2|},
        \hspace{0.4cm}w_4 &= \frac{|x_2|}{x_1^2 + x_2^2 + |x_1| + |x_2|}
    \end{aligned} \\
    &\bullet \quad f(x_4 \mid x_3) = \mathcal{N}(0.5 x_3, 0.5^2) \\
    &\bullet \quad f(x_5 \mid x_4) = 
    w_1 \mathcal{N}(-2, 0.5^2) + w_2 \mathcal{N}(0, 0.5^2) + w_3 \mathcal{N}(2, 0.5^2), \\
    &\begin{aligned}
        \hspace{0.7cm}w_1 &= \frac{1}{1 + \exp\left(-x_4/3\right)/2},
        \hspace{0.7cm}w_2 &= \frac{1}{1 + \exp\left(-x_4/3\right)},
        \hspace{0.7cm}w_3 &= 1 - w_2
    \end{aligned}
\end{aligned}
\]

\section{Support Theory}\label{apd:third}
\noindent \textbf{Cauchy's interlacing theorem (CIT):}
\begin{thm}
Let $\mathbf{A}$ be a $d\times d$ real symmetric matrix with eigenvalues ordered as: $
\lambda_1 \leq \lambda_2 \leq \dots \leq \lambda_d.$
Let $\mathbf{B}$ be a $(d-1) \times (d-1)$ principal submatrix of $\mathbf{A}$ with eigenvalues: $
\mu_1 \leq \mu_2 \leq \dots \leq \mu_{d-1}.$
Then, the eigenvalues of $\mathbf{B}$ interlace those of $\mathbf{A}$, i.e., $
\lambda_1 \leq \mu_1 \leq \lambda_2 \leq \mu_2 \leq \dots \leq \lambda_{d-1} \leq \mu_{d-1} \leq \lambda_d.
$
\label{theorem3}
\end{thm}

\bibliographystyle{apalike}
\bibliography{references}  

\begin{thebibliography}{}

\bibitem[Atienza et~al., 2022a]{atienza2022pybnesian}
Atienza, D., Bielza, C., and Larra{\~n}aga, P. (2022a).
\newblock Py{BN}esian: An extensible {P}ython package for {B}ayesian networks.
\newblock {\em Neurocomputing}, 504:204--209.

\bibitem[Atienza et~al., 2022b]{atienza2022semiparametric}
Atienza, D., Bielza, C., and Larra{\~n}aga, P. (2022b).
\newblock Semiparametric {B}ayesian networks.
\newblock {\em Information Sciences}, 584:564--582.

\bibitem[Benavoli et~al., 2016]{benavoli2016should}
Benavoli, A., Corani, G., and Mangili, F. (2016).
\newblock Should we really use post-hoc tests based on mean-ranks?
\newblock {\em Journal of Machine Learning Research}, 17(1):152--161.

\bibitem[Bishop, 2006]{bishop2006pattern}
Bishop, C.~M. (2006).
\newblock {\em Pattern Recognition and Machine Learning}.
\newblock Springer.

\bibitem[Bosq and Lecoutre, 1987]{bosq1987th}
Bosq, D. and Lecoutre, J.-P. (1987).
\newblock {\em Théorie de l’{E}stimation {F}onctionnelle}.
\newblock Économica.

\bibitem[Chac{\'o}n and Duong, 2010]{chacon2010multivariate}
Chac{\'o}n, J.~E. and Duong, T. (2010).
\newblock Multivariate plug-in bandwidth selection with unconstrained pilot bandwidth matrices.
\newblock {\em Test}, 19:375--398.

\bibitem[Chac{\'o}n and Duong, 2011]{chacon2011unconstrained}
Chac{\'o}n, J.~E. and Duong, T. (2011).
\newblock Unconstrained pilot selectors for smoothed cross-validation.
\newblock {\em Australian \& New Zealand Journal of Statistics}, 53(3):331--351.

\bibitem[Chac{\'o}n and Duong, 2018]{chacon2018multivariate}
Chac{\'o}n, J.~E. and Duong, T. (2018).
\newblock {\em Multivariate {K}ernel {S}moothing and its {A}pplications}.
\newblock Chapman and Hall/CRC.

\bibitem[Chac{\'o}n and Monfort, 2013]{chacon2013comparison}
Chac{\'o}n, J.~E. and Monfort, P. (2013).
\newblock A comparison of bandwidth selectors for mean shift clustering.
\newblock {\em arXiv preprint arXiv:1310.7855}.

\bibitem[Chickering, 2002]{chickering1996learning}
Chickering, D.~M. (2002).
\newblock Learning equivalence classes of {B}ayesian network structures.
\newblock {\em Journal of Machine Learning Research}, 2:445--498.

\bibitem[Cooper and Herskovits, 1992]{cooper1992bayesian}
Cooper, G.~F. and Herskovits, E. (1992).
\newblock A {B}ayesian method for the induction of probabilistic networks from data.
\newblock {\em Machine Learning}, 9:309--347.

\bibitem[Davari et~al., 2021]{metropt-3}
Davari, N., Veloso, B., Ribeiro, R., and Gama, J. (2021).
\newblock {MetroPT-3 Dataset}.
\newblock UCI Machine Learning Repository.
\newblock {DOI}: https://doi.org/10.24432/C5VW3R.

\bibitem[De~Gooijer and Zerom, 2003]{gooijer2003conditional}
De~Gooijer, J.~G. and Zerom, D. (2003).
\newblock On conditional density estimation.
\newblock {\em Statistica Neerlandica}, 57(2):159--176.

\bibitem[Dem{\v{s}}ar, 2006]{demvsar2006statistical}
Dem{\v{s}}ar, J. (2006).
\newblock Statistical comparisons of classifiers over multiple data sets.
\newblock {\em Journal of Machine Learning Research}, 7:1--30.

\bibitem[Duong and Hazelton, 2005]{duong2005cross}
Duong, T. and Hazelton, M.~L. (2005).
\newblock Cross-validation bandwidth matrices for multivariate kernel density estimation.
\newblock {\em Scandinavian Journal of Statistics}, 32(3):485--506.

\bibitem[Fan et~al., 1996]{fan1996estimation}
Fan, J., Yao, Q., and Tong, H. (1996).
\newblock Estimation of conditional densities and sensitivity measures in nonlinear dynamical systems.
\newblock {\em Biometrika}, 83(1):189--206.

\bibitem[Friedman and Nachman, 2000]{Friedman2000GaussianPN}
Friedman, N. and Nachman, I. (2000).
\newblock Gaussian process networks.
\newblock In {\em Proceedings of the 16th Conference on Uncertainty in Artificial Intelligence}, UAI '00, page 211–219. Morgan Kaufmann Publishers Inc.

\bibitem[Garc{{\'i}}a and Herrera, 2008]{garcia2008extension}
Garc{{\'i}}a, S. and Herrera, F. (2008).
\newblock An extension on ``statistical comparisons of classifiers over multiple data sets'' for all pairwise comparisons.
\newblock {\em Journal of Machine Learning Research}, 9(89):2677--2694.

\bibitem[Geiger and Heckerman, 1994]{geiger1994learning}
Geiger, D. and Heckerman, D. (1994).
\newblock Learning {G}aussian networks.
\newblock In {\em Proceedings of the Tenth International Conference on Uncertainty in Artificial Intelligence}, page 235–243. Morgan Kaufmann Publishers Inc.

\bibitem[Gelman et~al., 2020]{Gelman_Hill_Vehtari_2020}
Gelman, A., Hill, J., and Vehtari, A. (2020).
\newblock {\em Regression and {O}ther Stories}.
\newblock Cambridge University Press.

\bibitem[Heidenreich et~al., 2013]{heidenreich2013bandwidth}
Heidenreich, N.-B., Schindler, A., and Sperlich, S. (2013).
\newblock Bandwidth selection for kernel density estimation: {A} review of fully automatic selectors.
\newblock {\em Advances in Statistical Analysis}, 97:403--433.

\bibitem[Henderson and Searle, 1979]{henderson1979vec}
Henderson, H.~V. and Searle, S.~R. (1979).
\newblock Vec and vech operators for matrices, with some uses in jacobians and multivariate statistics.
\newblock {\em Canadian Journal of Statistics}, 7(1):65--81.

\bibitem[Hofmann and Tresp, 1995]{hofmann1995discovering}
Hofmann, R. and Tresp, V. (1995).
\newblock Discovering structure in continuous variables using {B}ayesian networks.
\newblock {\em Advances in Neural Information Processing Systems}, 8:500--506.

\bibitem[Horn and Johnson, 2012]{horn2012matrix}
Horn, R.~A. and Johnson, C.~R. (2012).
\newblock {\em Matrix {A}nalysis}.
\newblock Cambridge University Press.

\bibitem[Hyndman et~al., 1996]{hyndman1996estimating}
Hyndman, R.~J., Bashtannyk, D.~M., and Grunwald, G.~K. (1996).
\newblock Estimating and visualizing conditional densities.
\newblock {\em Journal of Computational and Graphical Statistics}, 5(4):315--336.

\bibitem[Ickstadt et~al., 2010]{ickstadt2010nonparametric}
Ickstadt, K., Bornkamp, B., Grzegorczyk, M., Wieczorek, J., Sheriff, M.~R., Grecco, H.~E., and Zamir, E. (2010).
\newblock Nonparametric {B}ayesian networks.
\newblock {\em Bayesian Statistics}, 9:283--316.

\bibitem[Koller and Friedman, 2009]{koller2009probabilistic}
Koller, D. and Friedman, N. (2009).
\newblock {\em Probabilistic {G}raphical {M}odels: {P}rinciples and {T}echniques}.
\newblock The MIT Press.

\bibitem[Langseth et~al., 2012]{langseth2012mixtures}
Langseth, H., Nielsen, T.~D., Rum{\'{i}}, R., and Salmer{\'o}n, A. (2012).
\newblock Mixtures of truncated basis functions.
\newblock {\em International Journal of Approximate Reasoning}, 53(2):212--227.

\bibitem[Larra{\~n}aga et~al., 1996]{larranaga1996learning}
Larra{\~n}aga, P., Kuijpers, C., Murga, R., and Yurramendi, Y. (1996).
\newblock Learning {B}ayesian network structures by searching for the best ordering with genetic algorithms.
\newblock {\em IEEE Transactions on Systems, Man, and Cybernetics - Part A: Systems and Humans}, 26(4):487--493.

\bibitem[Leiva and Art{\'e}s, 2012]{leiva2012algorithms}
Leiva, J.~M. and Art{\'e}s, A. (2012).
\newblock Algorithms for maximum-likelihood bandwidth selection in kernel density estimators.
\newblock {\em Pattern Recognition Letters}, 33(13):1717--1724.

\bibitem[Mangiafico, 2016]{mangiafico2016summary}
Mangiafico, S. (2016).
\newblock Summary and {A}nalysis of {E}xtension {E}ducation {P}rogram {E}valuation in {R}.
\newblock {\em Rutgers Cooperative Extension}.

\bibitem[Masmoudi and Masmoudi, 2019]{masmoudi2019new}
Masmoudi, K. and Masmoudi, A. (2019).
\newblock A new class of continuous {B}ayesian networks.
\newblock {\em International Journal of Approximate Reasoning}, 109:125--138.

\bibitem[Moral et~al., 2001]{moral2001mixtures}
Moral, S., Rum{\'\i}, R., and Salmer{\'o}n, A. (2001).
\newblock Mixtures of truncated exponentials in hybrid {B}ayesian networks.
\newblock In {\em Proceedings of the 6th European Conference, Symbolic and Quantitative Approaches to Reasoning with Uncertainty}, pages 156--167. Springer.

\bibitem[Nelder and Mead, 1965]{nelder1965simplex}
Nelder, J.~A. and Mead, R. (1965).
\newblock A simplex method for function minimization.
\newblock {\em The Computer Journal}, 7(4):308--313.

\bibitem[Pearl, 1988]{pearl1988probabilistic}
Pearl, J. (1988).
\newblock {\em Probabilistic {R}easoning in {I}ntelligent {S}ystems: {N}etworks of {P}lausible {I}nference}.
\newblock Morgan Kaufmann.

\bibitem[Rosenblatt, 1969]{rosenblatt1969conditional}
Rosenblatt, M. (1969).
\newblock Conditional probability density and regression estimators.
\newblock In {\em Multivariate Analysis, {II}}, pages 25--31. Academic Press.

\bibitem[Sain et~al., 1994]{sain1994cross}
Sain, S.~R., Baggerly, K.~A., and Scott, D.~W. (1994).
\newblock Cross-validation of multivariate densities.
\newblock {\em Journal of the American Statistical Association}, 89(427):807--817.

\bibitem[Scott, 2015]{scott2015multivariate}
Scott, D.~W. (2015).
\newblock {\em Multivariate {D}ensity {E}stimation: {T}heory, {P}ractice, and {V}isualization}.
\newblock John Wiley \& Sons.

\bibitem[Scott and Terrell, 1987]{scott1987biased}
Scott, D.~W. and Terrell, G.~R. (1987).
\newblock Biased and unbiased cross-validation in density estimation.
\newblock {\em Journal of the American Statistical Association}, 82(400):1131--1146.

\bibitem[Shachter and Kenley, 1989]{gaussiannetworks}
Shachter, R.~D. and Kenley, C.~R. (1989).
\newblock Gaussian influence diagrams.
\newblock {\em Management Science}, 35(5):527–550.

\bibitem[Shenoy and West, 2011]{shenoy2011inference}
Shenoy, P.~P. and West, J.~C. (2011).
\newblock Inference in hybrid {B}ayesian networks using mixtures of polynomials.
\newblock {\em International Journal of Approximate Reasoning}, 52(5):641--657.

\bibitem[Silverman, 2018]{silverman2018density}
Silverman, B.~W. (2018).
\newblock {\em Density {E}stimation for {S}tatistics and {D}ata {A}nalysis}.
\newblock Routledge.

\bibitem[Spirtes et~al., 2000]{spirtes2000causation}
Spirtes, P., Glymour, C.~N., and Scheines, R. (2000).
\newblock {\em Causation, {P}rediction, and {S}earch}.
\newblock The MIT Press.

\bibitem[Tsamardinos et~al., 2006]{tsamardinos2006max}
Tsamardinos, I., Brown, L.~E., and Aliferis, C.~F. (2006).
\newblock The max-min hill-climbing {B}ayesian network structure learning algorithm.
\newblock {\em Machine Learning}, 65:31--78.

\bibitem[Wand, 1992]{wand1992error}
Wand, M.~P. (1992).
\newblock Error analysis for general multtvariate kernel estimators.
\newblock {\em Journal of Nonparametric Statistics}, 2(1):1--15.

\bibitem[Wand and Jones, 1994]{wand1994multivariate}
Wand, M.~P. and Jones, M.~C. (1994).
\newblock Multivariate plug-in bandwidth selection.
\newblock {\em Computational Statistics}, 9(2):97--116.

\bibitem[Zhang et~al., 2006]{zhang2006bayesian}
Zhang, X., King, M.~L., and Hyndman, R.~J. (2006).
\newblock A {B}ayesian approach to bandwidth selection for multivariate kernel density estimation.
\newblock {\em Computational Statistics \& Data Analysis}, 50(11):3009--3031.

\end{thebibliography}






\end{document}